\documentclass[12pt]{article}
\PassOptionsToPackage{unicode}{hyperref}
\PassOptionsToPackage{hyphens}{url}
\PassOptionsToPackage{dvipsnames,svgnames,x11names}{xcolor}

\usepackage{amsmath,amssymb}
\usepackage{iftex}
\ifPDFTeX
  \usepackage[T1]{fontenc}
  \usepackage[utf8]{inputenc}
  \usepackage{textcomp} 
\else 
  \usepackage{unicode-math}
  \defaultfontfeatures{Scale=MatchLowercase}
  \defaultfontfeatures[\rmfamily]{Ligatures=TeX,Scale=1}
\fi
\usepackage{lmodern}
\ifPDFTeX\else  
\fi
\IfFileExists{upquote.sty}{\usepackage{upquote}}{}
\IfFileExists{microtype.sty}{
  \usepackage[]{microtype}
  \UseMicrotypeSet[protrusion]{basicmath} 
}{}
\makeatletter
\@ifundefined{KOMAClassName}{
  \IfFileExists{parskip.sty}{%
    \usepackage{parskip}
  }{
    \setlength{\parindent}{0pt}
    \setlength{\parskip}{6pt plus 2pt minus 1pt}}
}{
  \KOMAoptions{parskip=half}}
\makeatother
\usepackage{xcolor}
\makeatletter
\ifx\paragraph\undefined\else
  \let\oldparagraph\paragraph
  \renewcommand{\paragraph}{
    \@ifstar
      \xxxParagraphStar
      \xxxParagraphNoStar
  }
  \newcommand{\xxxParagraphStar}[1]{\oldparagraph*{#1}\mbox{}}
  \newcommand{\xxxParagraphNoStar}[1]{\oldparagraph{#1}\mbox{}}
\fi
\ifx\subparagraph\undefined\else
  \let\oldsubparagraph\subparagraph
  \renewcommand{\subparagraph}{
    \@ifstar
      \xxxSubParagraphStar
      \xxxSubParagraphNoStar
  }
  \newcommand{\xxxSubParagraphStar}[1]{\oldsubparagraph*{#1}\mbox{}}
  \newcommand{\xxxSubParagraphNoStar}[1]{\oldsubparagraph{#1}\mbox{}}
\fi
\makeatother

\providecommand{\tightlist}{%
  \setlength{\itemsep}{0pt}\setlength{\parskip}{0pt}}\usepackage{longtable,booktabs,array}
\usepackage{calc} 
\usepackage{etoolbox}
\makeatletter
\patchcmd\longtable{\par}{\if@noskipsec\mbox{}\fi\par}{}{}
\makeatother
\IfFileExists{footnotehyper.sty}{\usepackage{footnotehyper}}{\usepackage{footnote}}
\makesavenoteenv{longtable}
\usepackage{graphicx}
\makeatletter
\def\maxwidth{\ifdim\Gin@nat@width>\linewidth\linewidth\else\Gin@nat@width\fi}
\def\maxheight{\ifdim\Gin@nat@height>\textheight\textheight\else\Gin@nat@height\fi}
\makeatother
\setkeys{Gin}{width=\maxwidth,height=\maxheight,keepaspectratio}
\makeatletter
\def\fps@figure{htbp}
\makeatother

\makeatletter
\@ifpackageloaded{caption}{}{\usepackage{caption}}
\AtBeginDocument{%
\ifdefined\contentsname
  \renewcommand*\contentsname{Table of contents}
\else
  \newcommand\contentsname{Table of contents}
\fi
\ifdefined\listfigurename
  \renewcommand*\listfigurename{List of Figures}
\else
  \newcommand\listfigurename{List of Figures}
\fi
\ifdefined\listtablename
  \renewcommand*\listtablename{List of Tables}
\else
  \newcommand\listtablename{List of Tables}
\fi
\ifdefined\figurename
  \renewcommand*\figurename{Figure}
\else
  \newcommand\figurename{Figure}
\fi
\ifdefined\tablename
  \renewcommand*\tablename{Table}
\else
  \newcommand\tablename{Table}
\fi
}
\@ifpackageloaded{float}{}{\usepackage{float}}
\floatstyle{ruled}
\@ifundefined{c@chapter}{\newfloat{codelisting}{h}{lop}}{\newfloat{codelisting}{h}{lop}[chapter]}
\floatname{codelisting}{Listing}

\makeatother

\ifLuaTeX
  \usepackage{selnolig}  
\fi
\usepackage[]{natbib}
\usepackage{bookmark}

\IfFileExists{xurl.sty}{\usepackage{xurl}}{} 
\hypersetup{
  pdftitle={Generalized Linear Markov Decision Process},
  pdfauthor={Sinian Zhang; Kaicheng Zhang; Ziping Xu; Tianxi Cai; Doudou Zhou},
  pdfkeywords={Structured MDPs, Bellman Completeness, Generalized Linear Models, Offline Reinforcement Learning, Sample Efficiency},
  colorlinks=true,
  linkcolor={blue},
  filecolor={Maroon},
  citecolor={Blue},
  urlcolor={Blue},
  pdfcreator={LaTeX via pandoc}}

\usepackage{mathtools}
\usepackage{amsfonts}
\usepackage{algorithm}      
\usepackage[noend]{algpseudocode} 
\usepackage{wrapfig}
\usepackage{graphicx}
\usepackage{caption}
\usepackage{xcolor}         
\usepackage[normalem]{ulem} 
\usepackage{amsmath}
\usepackage{enumerate}
\usepackage{amsthm}
\usepackage{booktabs}
\usepackage[capitalize,noabbrev]{cleveref}
\usepackage{booktabs}
\usepackage{tabularx}
\usepackage{array}
\usepackage{multirow}
\usepackage{tikz}
\usepackage{pgfplots}
\usepackage{subcaption}
\pgfplotsset{compat=1.18}
\usetikzlibrary{arrows.meta,positioning}

\newtheorem{theorem}{Theorem}
\newtheorem{definition}{Definition}
\newtheorem{assumption}{Assumption}
\newtheorem{lemma}{Lemma} 
\newtheorem{proposition}{Proposition} 
\newtheorem{remark}{Remark}
\newtheorem{corollary}{Corollary}

\usepackage{comment}
\usepackage{xcolor}
\newcount\comments  
\newcommand{\genComment}[2]{\ifnum\comments=1{\textcolor{#1}{\textsf{\footnotesize #2}}}\fi}

\newcommand{\sn}[1]{{\color{blue}#1}}

\def\Vsc{\mathcal{V}}

\def\Dsc{\mathcal{D}}

\def \be{\begin{equs}}
\def \ee{\end{equs}}

\def\bI{\mathbf{I}}

\def\Asc{\mathcal{A}}
\def\Lsc{\mathcal{L}}

\def\Dsc{\mathcal{D}}
\def\Ssc{\mathcal{S}}
\def\Esc{\mathcal{E}}
\def\Fsc{\mathcal{F}}

\def\bone{\mathbf{1}}

\def\Isc{\mathcal{I}}

\def\P{\mathbb{P}}

\def\E{\mathbb{E}}
\def\R{\mathbb{R}}
\def\B{\mathbb{B}}

\def\bM{\mathbf{M}}

\def\argmindum{\mathop{\mbox{argmin}}}
\def\argmin#1{\argmindum_{#1}}

\def\Nsc{\mathcal{N}}

\usepackage{xr}
\makeatletter
\newcommand*{\addFileDependency}[1]{
  \typeout{(#1)}
  \@addtofilelist{#1}
  \IfFileExists{#1}{}{\typeout{No file #1.}}
}
\makeatother

\newcommand{\anon}{1}

\begin{document}

\def\spacingset#1{\renewcommand{\baselinestretch}%
{#1}\small\normalsize} \spacingset{1}


\if1\anon
{
  \title{\bf Generalized Linear Markov Decision Process}
    \author{
    Sinian Zhang\textsuperscript{1}\thanks{Equal contribution.},
    Kaicheng Zhang\textsuperscript{2}\footnotemark[1], 
    Ziping Xu\textsuperscript{3},  \\
    Zongqi Xia\textsuperscript{4},
    Jue Hou\textsuperscript{1},
    Tianxi Cai\textsuperscript{5}\footnotemark[2], 
    Doudou Zhou\textsuperscript{6}\thanks{Corresponding author.} \bigskip \\
    \small 
    \textsuperscript{1} Division of Biostatistics and Health Data Science, University of Minnesota, USA  \\
    \small 
    \textsuperscript{2} Department of Biostatistics, University of North Carolina at Chapel Hill, USA \\
    \small 
    \textsuperscript{3} Department of Statistics, Harvard University, USA\\
    \small 
    \textsuperscript{4} Department of Neurology, University of Pittsburgh, USA\\
    \small
    \textsuperscript{5} Department of Biostatistics, Harvard T.H. Chan School of Public Health, USA \\
    \small 
    \textsuperscript{6} Department of Statistics and Data Science, National University of Singapore, Singapore 
    \\
    \small
    \texttt{tcai@hsph.harvard.edu}, \texttt{ddzhou@nus.edu.sg} 
    }
    \date{}
  \maketitle
} \fi

\if0\anon
{
  \bigskip
  \bigskip
  \bigskip
  \begin{center}
    {\LARGE\bf Title}
\end{center}
} \fi

\smallskip

\begin{abstract}
Offline reinforcement learning for longitudinal studies often faces two linked challenges: rewards may be binary or bounded, and reward observations may be available only for a subset of trajectories or time points even when the corresponding state-action-next-state histories are available. Linear Markov decision process methods are tractable because Bellman backups remain linear, but they require linear rewards and do not indicate how transition-only observations should be used. We introduce {\bf GRASP-MDP}, Generalized Reward And Semi-supervised Pessimism for Markov Decision Processes, a reward-transition separated framework addressing both issues through one Bellman decomposition. It preserves linear transition dynamics while modeling reward means through generalized linear models. Although this breaks the usual linear Bellman form, the backup remains explicit as a nonlinear reward component plus a linear continuation component, yielding a nonlinear-plus-linear Bellman-complete class. In the resulting recursion, observed rewards estimate the generalized reward component, while all available transitions estimate the continuation component. The resulting pessimistic value iteration controls the two estimation errors separately without reward imputation. Finite-sample guarantees show that transition-only observations reduce transition-estimation error while reward uncertainty remains governed by observed rewards. Simulations and a multiple sclerosis electronic health record application illustrate the empirical benefit of retaining transition-only observations.
\end{abstract}

\vfill

\newpage
\spacingset{1.8} 

\section{Introduction}
\label{sec:intro}

Many longitudinal decision problems now arise in settings where the process of care or user interaction is recorded in detail, but the outcome that should guide decision making is observed much less completely. In healthcare, for example, electronic health records (EHRs) contain longitudinal information on diagnoses, treatments, and laboratory measurements, while the clinically meaningful outcomes may require specialist assessment, registry linkage, manual validation, or a longer follow-up window. Similar gaps arise in digital intervention and recommendation settings, where user actions and subsequent states are routinely logged but the reward of scientific interest may be delayed, observed intermittently, or expensive to measure. These settings make offline reinforcement learning (RL) appealing, because policies must often be learned from existing observational data rather than through prospective experimentation.

Our motivating application, learning disease-modifying treatment (DMT) strategies for patients with multiple sclerosis (MS), provides one such setting. MS is a chronic autoimmune disease that can lead to progressive neurological disability, making treatment decisions inherently longitudinal. Clinicians repeatedly decide whether and how to modify therapy based on evolving disease activity, prior treatment history, and functional status. DMTs vary substantially in their effects on relapse and inflammatory disease activity \citep{liu2021disease,grand2018sequencing}; for this analysis, we operationally group them into high-efficacy and standard-efficacy classes following prior real-world MS treatment classifications \citep{gan2025knowledge}. Although relapse is often used as a short-term marker of disease activity, long-term functional status, as measured by the Expanded Disability Status Scale (EDSS), remains widely used in treatment evaluation and clinical trials despite well-recognized limitations in its sensitivity to individual change \citep{kurtzke1983rating,chard2017resolving}. In EHRs, treatment histories, diagnosis codes, laboratory measurements, and clinical-note features may be recorded repeatedly over time, but EDSS scores, which lie on a bounded disability scale, are typically available only at selected clinical or registry visits. Learning DMT strategies from such data requires a framework that can use abundant state-action-next-state information while modeling the EDSS-based reward on its natural bounded scale. Restricting the analysis to EDSS-observed periods discards state-action-next-state transitions from periods without observed EDSS rewards, even though those transitions still inform continuation values; reward imputation brings them back only by adding assumptions about unobserved rewards. This motivates a Bellman recursion in which reward and transition components remain separated and are estimated from the data sources that directly identify them.

Linear MDPs \citep{jin2020provably} provide a natural starting point for statistically tractable RL because they combine structured transition dynamics, computationally feasible value iteration, and finite-sample guarantees. Their tractability relies on a strong linear structure: both the expected reward and the transition kernel are linear in a feature representation, so each Bellman backup remains a single linear object. This is restrictive for many longitudinal outcomes, such as binary or bounded continuous rewards, whose conditional means are more naturally represented by generalized or related mean models \citep[e.g.][]{mccullagh2019generalized,ferrari2004beta}. In multi-step decision making, a nonlinear reward mean also breaks the usual linear Bellman recursion: the conditional reward mean plus the continuation value no longer belongs to the standard linear function class. Thus, extending linear MDPs to generalized rewards requires a new Bellman structure rather than a simple replacement of the reward regression. Flexible nonlinear value-function approximators can accommodate nonlinear reward relationships \citep{mnih2015human,levine2020offline}, but they do not by themselves preserve the explicit Bellman structure of linear MDPs or characterize how reward-observed and transition-only trajectories contribute separately to policy learning. We instead seek a structured extension that relaxes the linear reward specification while retaining tractable value iteration and finite-sample analysis.

We address this gap by introducing Generalized Reward And Semi-supervised Pessimism for Markov Decision Processes (GRASP-MDP).  GRASP-MDP is built around a reward-transition separation. The transition kernel remains linear in a feature map, preserving the tractability of linear MDPs, while the conditional mean reward is modeled through a generalized linear model (GLM). Specifically, GRASP-MDP assumes
\begin{equation*}
    \mathbb{E}[ r_h(x_h,a_h) \mid x_h=x,a_h=a]
    = g( \langle \phi_r(x,a),\theta_h^* \rangle),
\end{equation*}
\begin{equation*}
    \P_{h}(x_{h+1} \mid x_h, a_h)
    = \langle \phi_p(x_h, a_h), \mu_{h}(x_{h+1})\rangle,
\end{equation*}
where $g$ is a known mean function and $\phi_r$ and $\phi_p$ are the reward and transition feature maps; formal definitions are given in Section~\ref{sec:problem}. When $g$ is the identity map and the reward and transition representations coincide, this reduces to the standard linear-MDP model.

The Bellman backup induced by this model belongs to a nonlinear-plus-linear function class: the reward component is transformed through the generalized linear mean map, whereas the continuation-value component remains linear in the transition feature. The resulting Bellman-complete class is not the usual linear class, but a class tailored to the two components of the model. The decomposition also determines how reward availability enters policy learning: observed rewards are needed to estimate the nonlinear reward component, but observations with state-action-next-state information and no observed reward still identify the transition component and therefore the continuation value. We consider trajectory-level partial reward observation as the primary setting, complete reward observation as its specialization, and step-level partial reward observation as an extension. GRASP-MDP uses transition-only observations to estimate the transition component, with pessimism guarding separately against reward- and transition-estimation errors.

The proposed framework is related to several literatures. Linear-MDP methods provide templates for tractable planning and online learning \citep[e.g.][]{yang2019sample,jin2020provably}; related finite-sample results address batch off-policy evaluation \citep{duan2020minimax} and offline policy learning \citep{jin2020pessimism}. These methods retain a linear reward specification. Under complete reward observation, when $g$ is the identity map and the reward and transition representations coincide, GRASP-MDP recovers the same suboptimality order as pessimistic value iteration (PEVI) \citep{jin2020pessimism}. Generalized linear bandits handle nonlinear reward means in one-step decision problems \citep{filippi2010parametric,li2017provably,faury2020improved,sawarni2024generalized}, but they do not involve transitions or Bellman recursion. Broader Bellman-complete or realizable-function-class frameworks impose structure directly on value or Bellman classes \citep{wang2019optimism,Wang2020Reinforcement,xie2021bellman,zanette2021provable}. GRASP-MDP instead specifies separate data-generating reward and transition components, yielding an explicit nonlinear-plus-linear Bellman class with GLM and ridge-regression updates matched to the two components. Existing semi-supervised offline RL approaches exploit incompletely labeled trajectories through reward learning, inverse-dynamics-based action imputation, or reward-free data sharing \citep{konyushkova2020semisupervisedrewardlearningoffline,zheng2023semisupervisedofflinereinforcementlearning,hu2023provable}. GRASP-MDP instead uses transition-only observations exclusively for transition estimation; the semi-supervised gain thus comes from the reward-transition Bellman decomposition itself rather than from an additional reward-observation model.

The paper makes four main contributions. First, we establish Bellman completeness for the nonlinear-plus-linear class induced by the reward-transition model. Second, we develop componentwise pessimistic value iteration and finite-sample guarantees that control reward uncertainty and transition uncertainty separately; the proof handles the data-dependent continuation values through uniform transition concentration. Third, the analysis covers representative trajectory-level reward labeling, its complete-reward specialization, and step-level reward observation under missing at random and positivity. Under partial reward observation, transition-only observations improve the transition term without being used to estimate the reward model. Fourth, simulations and an MS EHR application illustrate the method for bounded, intermittently observed outcomes with nonlinear conditional means.

\section{GRASP-MDP Framework}
\label{sec:problem}
In this section, we formalize the GRASP-MDP framework and derive the Bellman-complete function class induced by its reward-transition separation. We consider an episodic MDP with state space \(\Ssc\), action space \(\Asc\), horizon $H$, stage-wise rewards $\{r_h\}_{h=1}^H$, and transition kernels $\{P_h\}_{h=1}^H$. For any positive integer $m$, write \([m]=\{1,\ldots,m\}\), and let \(r_h=r_h(x_h,a_h)\) denote the random reward received at step $h$. At each step $h\in[H]$, GRASP-MDP assumes
\begin{equation}
\begin{aligned}
\mathbb{E}[r_h(x_h,a_h)\mid x_h=x,a_h=a]
&=g\bigl(\langle\phi_r(x,a),\theta_h^*\rangle\bigr),\\
\P_h(x_{h+1}\mid x_h,a_h)
&=\langle\phi_p(x_h,a_h),\mu_h(x_{h+1})\rangle,
\end{aligned}
\label{eq:grasp-model} 
\end{equation}
where $\phi_r:\Ssc\times\Asc\to\mathbb{R}^{d_r}$ and $\phi_p:\Ssc\times\Asc\to\mathbb{R}^{d_p}$ are known reward and transition feature maps, $g$ is a known mean function, $\theta_h^*\in\mathbb{R}^{d_r}$ is an unknown reward parameter, and $\mu_h=(\mu_{h,1},\ldots,\mu_{h,d_p})$ is an unknown vector of measures over $\Ssc$. The reward and transition feature maps $\phi_r$ and $\phi_p$ need not coincide, so reward-relevant and transition-relevant information may play different statistical roles.

For a policy $\pi=\{\pi_h\}_{h=1}^H$, let $\E_{\pi}$ denote expectation under the trajectory distribution induced by $\pi$, and define
\[
V_h^\pi(x)=\mathbb{E}_\pi\left[\sum_{t=h}^H r_t\mid x_h=x\right], \qquad Q_h^\pi(x,a)=\mathbb{E}_\pi\left[\sum_{t=h}^H r_t\mid x_h=x,a_h=a\right].
\]
We set $V_{H+1}^\pi\equiv V_{H+1}^*\equiv0$ and define
\[
V_h^*(x)=\sup_\pi V_h^\pi(x), \qquad Q_h^*(x,a)=\mathbb{E}\left[r_h+V_{h+1}^*(x_{h+1})\mid x_h=x,a_h=a\right].
\]
When an optimal policy exists, it satisfies
\[
\pi_h^*(\cdot\mid x)\in\arg\max_{\nu\in\Delta(\Asc)}\langle Q_h^*(x,\cdot),\nu\rangle_{\Asc}, \qquad V_h^*(x)=\langle Q_h^*(x,\cdot),\pi_h^*(\cdot\mid x)\rangle_{\Asc}.
\]
Here $\Delta(\Asc)$ denotes the set of probability distributions over the action space $\Asc$, and $\langle\cdot,\cdot\rangle_{\Asc}$ denotes summation over a discrete action space or integration with respect to the action distribution for a continuous action space. Our objective is to learn a policy from offline data whose value is close to the optimal value. For an initial state \(x\), we measure the performance loss of a policy \(\pi\) by
\begin{equation}
\operatorname{SubOpt}(\pi;x)=V_1^*(x)-V_1^\pi(x).
\label{eq:subopt}
\end{equation}
For any bounded measurable function \(V:\Ssc\to\mathbb{R}\), define the Bellman operator at step \(h\) by 
$$(\B_h V)(x,a)=\mathbb{E}[r_h+V(x_{h+1})\mid x_h=x,a_h=a].
$$
The value and Q-functions satisfy 
\(V_h^\pi(x)=\big\langle Q_h^\pi(x,\cdot),\pi_h(\cdot\mid x)\big\rangle_{\Asc}\) and \(Q_h^\pi(x,a)=(\B_h V_{h+1}^\pi)(x,a)\).

Under GRASP-MDP, the Bellman operator admits the decomposition
\begin{equation}
    (\B_h V)(x,a)
    =
    g\bigl(\langle \phi_r(x,a),\theta_h^* \rangle\bigr)
    +
    \left\langle
        \phi_p(x,a),
        \int_{\Ssc} V(x')\,\mu_h(dx')
    \right\rangle .
    \label{eq:grasp-bellman-decomposition}
\end{equation}
Replacing the linear reward model by a generalized mean model does not preserve the usual linear Bellman class; nevertheless, because the transition kernel remains linear, the backup in \eqref{eq:grasp-bellman-decomposition} still splits into a nonlinear reward term and a linear continuation term.
This motivates the nonlinear-plus-linear function class
\begin{equation}
\mathcal{F}=\left\{f_{\theta,\beta}(x,a)=g\bigl(\langle\phi_r(x,a),\theta\rangle\bigr)+\langle\phi_p(x,a),\beta\rangle:\theta\in\mathbb{R}^{d_r},\ \beta\in\mathbb{R}^{d_p}\right\}.
\label{eq:grasp-function-class}
\end{equation}
Thus, GRASP-MDP does not impose a generic realizability assumption directly on the Q-functions. Instead, the nonlinear-plus-linear class in \eqref{eq:grasp-function-class} is induced explicitly by the underlying reward and transition models.

\begin{proposition}[Bellman completeness under GRASP-MDP]
\label{prop:bellman_completeness}
For any bounded measurable function $V:\Ssc\to\mathbb{R}$ and any $h\in[H]$, the Bellman backup $\B_h V$ belongs to the function class $\mathcal{F}$. Specifically, with \(\beta_h(V)=\int_{\Ssc} V(x')\,\mu_h(dx')\),
\begin{equation*}
    (\B_h V)(x,a)
    =
    g\bigl(\langle \phi_r(x,a),\theta_h^*\rangle\bigr)
    +
    \langle \phi_p(x,a),\beta_h(V)\rangle.
\end{equation*}
Consequently, whenever the corresponding continuation-value functions are bounded, $Q_h^\pi\in\mathcal{F}$ for any policy $\pi$, and in particular $Q_h^*\in\mathcal{F}$ with \(Q_h^*(x,a)=g\bigl(\langle \phi_r(x,a),\theta_h^*\rangle\bigr)+\langle \phi_p(x,a),\beta_h^*\rangle\) and \(\beta_h^*=\int_{\Ssc} V_{h+1}^*(x')\,\mu_h(dx')\).
\end{proposition}
Thus, under the GRASP-MDP model, the nonlinear reward mean introduces no Bellman approximation error relative to \(\mathcal F\).

This structural separation also determines how the two Bellman components should be estimated. Let
\[
m_h(x,a)=g\bigl(\langle\phi_r(x,a),\theta_h^*\rangle\bigr).
\]
For an estimated reward mean $\widehat m_h$ and continuation parameter $\widehat\beta_h(V)$, the Bellman-estimation error decomposes as
\begin{equation}
(\widehat{\B}_hV)(x,a)-(\B_hV)(x,a)=\underbrace{\widehat m_h(x,a)-m_h(x,a)}_{\text{reward-estimation error}}+\underbrace{\langle\phi_p(x,a),\widehat\beta_h(V)-\beta_h(V)\rangle}_{\text{transition-estimation error}}.
\label{eq:bellman-error-decomposition}
\end{equation}
Suppose \(\Gamma_{r,h}\) and \(\Gamma_{p,h}\) bound the absolute reward- and transition-estimation errors, respectively, for the continuation function under consideration. Then the Bellman uncertainty can be controlled componentwise:
\begin{equation}
\left|(\widehat{\B}_hV)(x,a)-(\B_hV)(x,a)\right|\leq\Gamma_{r,h}(x,a)+\Gamma_{p,h}(x,a).
\label{eq:componentwise-uncertainty}
\end{equation}
The reward component is estimated from observed rewards, whereas the continuation component uses all available state$\to$action$\to$next-state transitions. This motivates the componentwise pessimistic value-iteration procedure in Section~\ref{sec:algorithms}, with separate reward and transition uncertainties governed by their respective sample sizes.

\section{Optimization Procedure under GRASP-MDP}
\label{sec:algorithms}

We now translate the componentwise Bellman representation in Section~\ref{sec:problem} into an offline pessimistic value-iteration procedure.

\subsection{Observed-data structure and estimation principle}

Consider the observed offline dataset
\[
\Dsc=\left\{\left(x_h^\tau,a_h^\tau,O_h^\tau r_h^\tau,O_h^\tau\right):\tau\in[n+N],\ h\in[H]\right\},
\]
where the state--action--next-state triple is observed for every trajectory and step, and \(O_h^\tau\in\{0,1\}\) indicates whether the reward \(r_h^\tau\) is observed; by convention, \(O_h^\tau r_h^\tau=0\) when \(O_h^\tau=0\). Define \(\Isc_h^r=\{\tau\in[n+N]:O_h^\tau=1\}\) and \(n_h^r=|\Isc_h^r|\). This representation accommodates three reward-observation settings: (i) trajectory-level partial reward observation, where \(\Isc_h^r=[n]\) for all \(h\); (ii) complete reward observation, where \(N=0\) and \(\Isc_h^r=[n]\); and (iii) step-level partial reward observation, where \(\Isc_h^r\) may vary with \(h\).

For trajectory-level partial reward observation, the \(n\) reward-labeled trajectories and \(N\) transition-only trajectories are assumed to be i.i.d. under a common behavior policy \(\pi^b\).

Following Proposition~\ref{prop:bellman_completeness}, GRASP implements a decomposed pessimistic value iteration: at each step, the reward parameter $\theta_h^*$ is estimated from $\Isc_h^r$ through a generalized linear model, while the continuation parameter $\beta_h(V)$ is estimated from all $n+N$ trajectories through linear regression. The uncertainty penalties are calibrated using the same split, so trajectories without an observed reward contribute to transition estimation without requiring reward imputation. For simplicity, we assume that rewards are almost surely bounded in $[0,1]$; the unbounded case is discussed in Supplementary~\ref{sec:finite_mean_range}.

\subsection{The GRASP algorithms}
\label{subsec:grasp_reward_observation}

We present a single backward GRASP recursion in terms of $\Isc_h^r$. Starting from $\widehat V_{H+1}\equiv0$, the following updates are performed for $h=H,\ldots,1$. First, the reward parameter is estimated from $\Isc_h^r$ by
\begin{equation}
\label{est:theta}
\widetilde\theta_h=\arg\min_{\theta\in\mathbb{R}^{d_r}}\Lsc_h(\theta), \
\Lsc_h(\theta)=(n_h^r)^{-1}\sum_{\tau\in\Isc_h^r}\left\{-r_h^\tau\langle\phi_r(x_h^\tau,a_h^\tau),\theta\rangle+G(\langle\phi_r(x_h^\tau,a_h^\tau),\theta\rangle)\right\},
\end{equation}
where $\Lsc_h$ is the canonical generalized linear loss with mean function $g$ and cumulant function $G(a)=\int_0^a g(u)\,{\rm d}u$ \citep{mccullagh2019generalized}.

Write $\phi_{r,h}^\tau=\phi_r(x_h^\tau,a_h^\tau)$ and define the local curvature matrix
\[
\widetilde\Sigma_h(\widetilde\theta_h)=\sum_{\tau\in\Isc_h^r}w_{h,\tau}\phi_{r,h}^\tau(\phi_{r,h}^\tau)^\top, \quad \mbox{where}\quad
w_{h,\tau}=\dot g\{\langle\phi_{r,h}^\tau,\widetilde\theta_h\rangle\}  \ \mbox{and} \ \dot{g}(x)=dg(x)/dx.
\]
With $\|v\|_A=(v^\top Av)^{1/2}$, the reward uncertainty radius is
\begin{equation}
\widetilde\Gamma_{r,h}(x,a)=\alpha_r\left\|\dot g\{\langle\phi_r(x,a),\widetilde\theta_h\rangle\}\phi_r(x,a)\right\|_{\widetilde\Sigma_h(\widetilde\theta_h)^{-1}},
\label{eq:reward_radius}
\end{equation}
where $\alpha_r>0$ is a tuning constant. This radius measures local uncertainty in the estimated generalized reward mean. The displayed form assumes that $\widetilde\theta_h$ exists and $\widetilde\Sigma_h(\widetilde\theta_h)$ is nonsingular. Section~\ref{sec:theory} establishes the main results under a population-curvature condition, and Supplementary~\ref{sec:relax} gives a regularized alternative.

Next, define the transition design matrix
\[
\widehat\Lambda_h=\sum_{\tau=1}^{n+N}\phi_p(x_h^\tau,a_h^\tau)\phi_p(x_h^\tau,a_h^\tau)^\top.
\]
With ridge parameter $\lambda>0$, the continuation parameter is estimated by
\begin{equation}
\widehat\beta_h=(\widehat\Lambda_h+\lambda\bI_{d_p})^{-1}\sum_{\tau=1}^{n+N}\phi_p(x_h^\tau,a_h^\tau)\widehat V_{h+1}(x_{h+1}^\tau),
\label{est:beta_reward_observation}
\end{equation}
where $\bI_{d_p}$ is the $d_p\times d_p$ identity matrix. The resulting empirical Bellman backup is
\[
(\widehat\B_h\widehat V_{h+1})(x,a)=g\bigl(\phi_r(x,a)^\top\widetilde\theta_h\bigr)+\phi_p(x,a)^\top\widehat\beta_h.
\]

Following pessimistic value iteration \citep{jin2020pessimism}, we use the following uncertainty-quantifier definition. Let \(\P_{\Dsc}\) denote probability over the sampling distribution of the offline dataset \(\Dsc\).
\begin{definition}[\texorpdfstring{\(\xi\)-uncertainty quantifier}{xi-uncertainty quantifier}]
For \(\xi\in(0,1)\), a collection of nonnegative functions \(\{\Gamma_h:\Ssc\times\Asc\to\mathbb R\}_{h=1}^H\) is a \(\xi\)-uncertainty quantifier for \(\{\widehat{\B}_h\widehat V_{h+1}\}_{h=1}^H\) if
\[
\P_{\Dsc}\!\left(
\left| (\widehat{\B}_h\widehat V_{h+1})(x,a)-(\B_h\widehat V_{h+1})(x,a) \right|
\leq \Gamma_h(x,a)
\text{ for all }h\in[H],\ (x,a)\in\Ssc\times\Asc
\right)
\geq 1-\xi.
\]
\end{definition}
Unlike linear-MDP PEVI, which uses a single radius for a jointly estimated linear Bellman object, GRASP-MDP decomposes uncertainty across its separately estimated reward and continuation components:
\begin{equation}
    \widehat\Gamma_h(x,a) =
    \underbrace{\widetilde\Gamma_{r,h}(x,a)}_{\text{reward uncertainty}}+
    \underbrace{\widehat\Gamma_{p,h}(x,a)}_{\text{transition uncertainty}}, \ \widehat\Gamma_{p,h}(x,a)=\alpha_p\|\phi_p(x,a)\|_{(\widehat\Lambda_h+\lambda\bI_{d_p})^{-1}},
\label{def:gamma_reward_observation}
\end{equation}
where $\alpha_p>0$ is a tuning constant. Theorem~\ref{thm:partial_reward_observation_suboptimality} verifies this bound for trajectory-level partial reward observation. Corollaries~\ref{cor:complete_reward_observation_rate} and~\ref{cor:step_level_reward_observation_rate} give the corresponding complete- and step-level specializations, respectively.

For $z\in\mathbb R$, write $z^+=\max\{z,0\}$. The pessimistic \(Q\)-update is
\begin{equation}
\widehat Q_h(x,a)
=\min\left\{(\widehat\B_h\widehat V_{h+1})(x,a)-
\widehat\Gamma_h(x,a),\; H-h+1\right\}^+.
\label{eq:q_hat}
\end{equation}
The corresponding policy and value updates are
\begin{equation}
  \widehat\pi_h(\cdot\mid x)
\in
\arg\max_{\nu\in\Delta(\Asc)}
\big\langle
\widehat Q_h(x,\cdot),
\nu
\big\rangle_{\Asc},
\qquad
\widehat V_h(x)
=
\big\langle
\widehat Q_h(x,\cdot),
\widehat\pi_h(\cdot\mid x)
\big\rangle_{\Asc}.  
\label{eq:pi_hat}
\end{equation}
Algorithm~\ref{alg:grasp_reward_observation} summarizes the GRASP recursion for trajectory-level partial and complete reward observation. The step-level version is provided in Supplementary~\ref{sec:offline_algorithms_appendix}.

\begin{algorithm}[htbp]
\caption{GRASP for offline data with trajectory-level reward observation}
\label{alg:grasp_reward_observation}
\begin{algorithmic}[1]
\State Input: offline dataset \(\Dsc\); hyperparameters $\lambda$, $\alpha_r$, $\alpha_p$.
\State Convention: trajectory-level partial observation uses \(\Isc_h^r=[n]\); complete reward observation sets \(N=0\) and uses \(\Isc_h^r=[n]\).
\State Initialization: set $\widehat V_{H+1}(x)\leftarrow 0$.
\For{step $h=H,H-1,\ldots,1$}
    \State Estimate $\widetilde\theta_h$ from the reward-observation set \(\Isc_h^r\) using \eqref{est:theta}.
    \State Estimate $\widehat\beta_h$ from all \(n+N\) trajectories using \eqref{est:beta_reward_observation}.
    \State Construct $\widehat\Gamma_h=\widetilde\Gamma_{r,h}+\widehat\Gamma_{p,h}$ by \eqref{def:gamma_reward_observation}.
    \State Form the pessimistic Q-function $\widehat Q_h$ and greedy policy $\widehat\pi_h$ using \eqref{eq:q_hat} and \eqref{eq:pi_hat}.
    \State Set $\widehat V_h(x)=\langle \widehat Q_h(x,\cdot),\widehat\pi_h(\cdot\mid x)\rangle_{\Asc}$.
\EndFor
\State Output: $\widehat\pi=\{\widehat\pi_h\}_{h=1}^H$.
\end{algorithmic}
\end{algorithm}

\section{Theoretical Analysis}
\label{sec:theory}

We establish finite-sample guarantees for trajectory-level partial, complete, and step-level reward observation. The bounds separate reward uncertainty, governed by observed rewards, from transition uncertainty, governed by all available state--action--next-state triples. Throughout, $O(\cdot)$ hides constants independent of $n,N,H,d_r,d_p,\xi$, and $\widetilde O(\cdot)$ additionally hides logarithmic factors.

For a data-dependent integrand, $\E_{\pi^*}$ is understood as expectation over an independent trajectory generated by $\pi^*$ in the true MDP, conditional on the offline data.

\begin{assumption}[Mean function]
\label{assump:link}
The mean function $g(\cdot)$ is twice continuously differentiable, with derivatives denoted by $\dot g$ and $\ddot g$. There exists a constant $L>0$ such that $0<\dot g(u)\leq L$ and $|\ddot g(u)|\leq L$ for every $u\in\mathbb R$; consequently, $|\dot g(u)-\dot g(v)|\leq L|u-v|$ for all $u,v\in\mathbb R$.
\end{assumption}




\begin{assumption}[Reward curvature]
\label{assump:sigma}
For the population reward curvature matrix
\[
\Sigma_{h}(\theta)
=\E_{\pi^b}\bigl[\dot{g}\{\langle \phi_r(x_h,a_h),\theta\rangle\}
\phi_r(x_h,a_h)\phi_r(x_h,a_h)^\top\bigr],
\]
there exists a constant $\rho>0$ such that, for every $h\in[H]$, $\lambda_{\min}\{\Sigma_h(\theta_h^*)\}\geq \rho$.
\end{assumption}

For technical simplicity, we also assume that $\max \{ \|\phi_r(x, a)\|_2^2, \|\phi_p(x, a)\|_2^2 \}\leq 1$ for all $(x, a) \in \Ssc \times \Asc$, and that $\big \|\mu_{h}(\Ssc)\big\|\leq \sqrt{d_p}$, where $\big\|\mu_{h}(\Ssc)\big\|\coloneqq\int_{\Ssc}\big\|\mu_{h}(x)\big\|_2 \mathrm{d} x$. These normalization conditions are standard in the linear MDP literature and can be achieved by suitable rescaling.
\begin{remark}[The role of the curvature condition]
Assumption~\ref{assump:sigma} is a convenient sufficient condition for localizing the unregularized reward estimator. Supplementary~\ref{sec:relax} shows that, under complete reward observation, regularization removes this lower-eigenvalue condition and yields a data-dependent reward radius, although the radius need not vanish in reward-feature directions without sufficient coverage.
\end{remark}
Under these conditions, we first give the main guarantee for trajectory-level partial reward observation, before specializing it to complete and step-level reward observation.

\begin{theorem}[\texorpdfstring{GRASP under partial reward observation}{GRASP  under partial reward observation}]
\label{thm:partial_reward_observation_suboptimality}
Suppose trajectory-level partial reward observation holds, so \(\Isc_h^r=[n]\) for every \(h\), and the \(n+N\) trajectories are i.i.d. under the behavior policy \(\pi^b\). Under Assumptions~\ref{assump:link}--\ref{assump:sigma}, set $\lambda=1$,
\[
\alpha_r=c_r \sqrt{d_r+\log (H / \xi)},
\qquad
\alpha_p=c_p(d_p+d_r) H \sqrt{\zeta_+},
\qquad
\zeta_+=\log \{2(d_r+d_p)H(n+N)/\xi\},
\]
where $c_r,c_p>0$ are sufficiently large constants and $\xi \in (0,1)$. Then $\widehat \Gamma_h$ in \eqref{def:gamma_reward_observation} is a $\xi$-uncertainty quantifier for $\widehat{\B}_h \widehat{V}_{h+1}$. For any initial state $x \in \Ssc$ and \(n\) large enough, the policy $\widehat \pi = \{ \widehat \pi_h \}_{h=1}^H$ returned by Algorithm~\ref{alg:grasp_reward_observation} satisfies
\[
    \operatorname{SubOpt}(\widehat{\pi} ; x)
    \leq
    2\sum_{h=1}^H
    \E_{\pi^*}\left[
    \widehat{\Gamma}_h(x_h, a_h)
    \mid x_1=x
    \right]
\]
with probability at least $1-\xi$.
\end{theorem}

Theorem~\ref{thm:partial_reward_observation_suboptimality} gives the main offline guarantee in terms of the estimated reward and transition uncertainty radii. The reward radius is governed by \(|\Isc_h^r|=n\), whereas the transition radius is governed by \(n+N\). To express this separation as an explicit sample-size rate, define the population transition covariance matrix under the behavior policy by
\[
\Lambda_{h}
=\E_{\pi^b}\bigl[\phi_p(x_h,a_h)\phi_p(x_h,a_h)^\top\bigr].
\]
When this matrix is uniformly well conditioned, the uncertainty-radius bound in Theorem~\ref{thm:partial_reward_observation_suboptimality} yields the following rate.

\begin{corollary}[\texorpdfstring{Rate for GRASP under partial reward observation}{Rate for GRASP under partial reward observation}]
\label{cor:partial_reward_observation_rate}
Under the assumptions of Theorem~\ref{thm:partial_reward_observation_suboptimality}, if there exists a constant $\rho_p>0$ such that $\lambda_{\min}(\Lambda_h)\geq \rho_p$ for every $h\in[H]$, then for any initial state \(x\in\Ssc\) and $n$ large enough, with probability at least $1-\xi$, 
\[
\operatorname{SubOpt}(\widehat{\pi} ; x)
\leq
O\left(\sqrt{\frac{\{d_r+\log (H / \xi)\}H^2}{n\rho}}\right)
+
O\left(\sqrt{\frac{(d_p+d_r)^2H^4\zeta_+}{(n+N)\rho_p}}\right).
\]
\end{corollary}

Complete reward observation corresponds to \(N=0\) and \(\Isc_h^r=[n]\), so the reward and transition terms are governed by the same sample size. To distinguish this specialization from the partial-observation procedure, we denote its transition design matrix, continuation estimator, uncertainty radius, Bellman backup, value function, and policy by \(\widetilde\Lambda_h\), \(\widetilde\beta_h\), \(\widetilde\Gamma_h\), \(\widetilde\B_h\), \(\widetilde V_h\), and \(\widetilde\pi_h\), respectively.

\begin{corollary}[\texorpdfstring{Complete reward-observation sample-size specialization}{Complete reward-observation sample-size specialization}]
\label{cor:complete_reward_observation_rate}
Under Assumptions~\ref{assump:link}--\ref{assump:sigma}, run Algorithm~\ref{alg:grasp_reward_observation} with \(N=0\) and \(\Isc_h^r=[n]\) for every \(h\), and set $\lambda=1$,
\[
\alpha_r=c_r\sqrt{d_r+\log(H/\xi)},
\qquad
\alpha_p=c_p(d_p+d_r)H\sqrt{\zeta},
\qquad
\zeta=\log\{2(d_r+d_p)Hn/\xi\},
\]
where $c_r,c_p>0$ are sufficiently large constants and $\xi\in(0,1)$. Then \(\widetilde\Gamma_h\), obtained by specializing the decomposed radius \(\widehat\Gamma_h\) in \eqref{def:gamma_reward_observation} to \(N=0\) and \(\Isc_h^r=[n]\), is a \(\xi\)-uncertainty quantifier for \(\widetilde\B_h\widetilde V_{h+1}\). Moreover, for any initial state $x\in\Ssc$ and $n$ large enough, the policy $\widetilde\pi=\{\widetilde\pi_h\}_{h=1}^H$ returned by Algorithm~\ref{alg:grasp_reward_observation} satisfies
\[
 \operatorname{SubOpt}(\widetilde\pi;x)
 \leq
 2\sum_{h=1}^H
 \E_{\pi^*}
 \left[
 \widetilde\Gamma_h(x_h,a_h)
 \mid x_1=x
 \right]
\]
with probability at least $1-\xi$. If, in addition, there exists a constant $\rho_p>0$ such that $\lambda_{\min}(\Lambda_h)\geq \rho_p$ for every $h\in[H]$, then
\[
\operatorname{SubOpt}\big(\widetilde \pi ;x \big)
\leq
O\left( \sqrt{\frac{\{d_r+\log (H/\xi)\}H^2}{n\rho}}\right)
+
O\left(  \sqrt{\frac{(d_p+d_r)^2H^4\log \left((d_p+d_r) H n / \xi\right)}{n\rho_p}}\right)
\]
with probability at least $1-\xi$. In addition,
\[
\max_{h\in [H]}\|\widetilde{\theta}_h-\theta_h^*\|_2\leq c\sqrt{\frac{d_r+\log(H/\xi)}{n\rho}}
\]
holds with probability at least $1-\xi$ for some constant $c>0$.
\end{corollary}

In the linear special case, the complete reward-observation specialization matches the PEVI \citep{jin2020pessimism} suboptimality order. Comparing Corollaries~\ref{cor:partial_reward_observation_rate} and~\ref{cor:complete_reward_observation_rate} also shows where transition-only trajectories help. In the partial reward-observation bound, the first term scales as \(\widetilde O\{\sqrt{d_rH^2/n}\}\) and reflects reward-estimation uncertainty, whereas the second term scales as \(\widetilde O\{\sqrt{(d_p+d_r)^2H^4/(n+N)}\}\) and reflects transition-estimation uncertainty. Thus the reward-estimation term remains governed by \(n\), whereas the transition-estimation term changes from a rate governed by \(n\) in the complete reward-observation specialization to one governed by \(n+N\).

A favorable regime occurs when transition estimation is the dominant source of error. Suppressing logarithmic and curvature factors, if \(d_p\gg d_r\) and \(N\gg nH^2d_p^2/d_r\), GRASP under partial reward observation achieves the simplified leading rate \(\widetilde O\{\sqrt{d_rH^2/n}\}\). This improves over the observed-reward-only transition-estimation rate \(\widetilde O\{\sqrt{(d_p+d_r)^2H^4/n}\}\).

In the step-level setting, only the first \(n\) trajectories are reward-eligible. At step \(h\), let
\[
n_h^{\rm obs}=|\Isc_h^r|=\sum_{\tau=1}^n O_h^\tau.
\]
All \(n+N\) trajectories contribute to transition estimation, while the additional \(N\) trajectories have no observed rewards.

Let \(\mathcal Z_h^b\) denote the support of \((x_h,a_h)\) under \(\pi^b\). Define the conditional and marginal reward-observation probabilities by
\[
p_h^{\rm obs}(x,a)=\mathbb P(O_h=1\mid x_h=x,a_h=a),
\qquad
q_h=\mathbb P(O_h=1)
=\E_{\pi^b}\{p_h^{\rm obs}(x_h,a_h)\}.
\]
Define the observed reward curvature and its minimum eigenvalue as
\[
\Sigma_h^{\rm obs}(\theta_h^*)
=\E_{\pi^b}\!\left[
\dot g\{\langle\phi_r(x_h,a_h),\theta_h^*\rangle\}
\phi_r(x_h,a_h)\phi_r(x_h,a_h)^\top
\mid O_h=1
\right],
\qquad
\rho_h^{\rm obs}
=\lambda_{\min}\{\Sigma_h^{\rm obs}(\theta_h^*)\}.
\]
Set
\[
\mathcal I_{\rm obs}
=\min_{h\in[H]}n_h^{\rm obs}\rho_h^{\rm obs}
\]
for the minimum observed reward information across steps.

\begin{assumption}[\texorpdfstring{Step-level reward observation and transition coverage}{Step-level reward observation and transition coverage}]
\label{assump:step_level_reward_observation_transition_coverage}
For every \(\tau\in[n]\) and \(h\in[H]\),
\[
\mathbb P(O_h^\tau=1\mid x_h^\tau,a_h^\tau,r_h^\tau)
=p_h^{\rm obs}(x_h^\tau,a_h^\tau),
\qquad
\inf_{\substack{h\in[H]\\(x,a)\in\mathcal Z_h^b}}
p_h^{\rm obs}(x,a)\geq \bar p>0.
\]
The \(n+N\) trajectories used for transition estimation are i.i.d.
under the behavior policy \(\pi^b\). The first \(n\) trajectories are
reward eligible, and their observation indicators are generated
according to the mechanism above, independently across trajectories.
Moreover, \(\lambda_{\min}(\Lambda_h)\geq\rho_p>0\) for every
\(h\in[H]\).
\end{assumption}

Conditioning on \(O_h=1\) reweights the behavior-policy design by \(p_h^{\rm obs}(x_h,a_h)/q_h\). Hence, Assumptions~\ref{assump:sigma} and~\ref{assump:step_level_reward_observation_transition_coverage} imply
\(
\rho_h^{\rm obs}\geq(\bar p/q_h)\rho.
\)

\begin{corollary}[\texorpdfstring{Step-level reward-observation sample-size specialization}{Step-level reward-observation sample-size specialization}]\label{cor:step_level_reward_observation_rate}
Suppose Assumptions~\ref{assump:link}--\ref{assump:step_level_reward_observation_transition_coverage} hold. For any initial state \(x\in\Ssc\), let \(\widehat\pi^{\rm obs}\) be the policy returned by GRASP (Algorithm~\ref{alg:grasp_step_level_reward_observation}) when, at each step \(h\), the reward model is fitted using only observations with \(O_h^\tau=1\), while the transition component is fitted using all \(n+N\) trajectories. If \(\lambda=1\), \(\alpha_r,\alpha_p\) are tuned as in Supplementary Theorem~\ref{thm:step_level_reward_observation} at confidence level \(\xi/3\), and \(n\bar p\) and \(n+N\) are sufficiently large, then with probability at least $1-\xi$,
\[
\operatorname{SubOpt}(\widehat\pi^{\rm obs};x)
\le
\widetilde O\!\left(
\sqrt{\frac{d_rH^2}{\mathcal I_{\rm obs}}}
\right)
+
\widetilde O\!\left(
\sqrt{\frac{(d_p+d_r)^2H^4}{(n+N)\rho_p}}
\right).
\]
\end{corollary}

Under the cross-trajectory sampling condition, \(n_h^{\rm obs}\) concentrates around \(nq_h\); together with \(\rho_h^{\rm obs}\ge(\bar p/q_h)\rho\), this gives \(\mathcal I_{\rm obs}\gtrsim n\bar p\rho\) with high probability. In particular, when reward observation is independent of \((x_h,a_h)\), \(\rho_h^{\rm obs}\ge\rho\), so the reward term incurs only the effective-sample-size loss \(n_h^{\rm obs}\asymp nq_h\).

The reward-side analysis is not tied to canonical GLMs. Supplementary~\ref{sec:m_estimator_extension} shows that the GRASP suboptimality argument extends to M-estimated reward models whenever the estimator provides a uniform confidence band for the conditional reward mean and its induced value class admits uniform transition concentration. These two conditions identify what is needed to incorporate other reward models without changing the transition-learning component. Supplementary~\ref{sec:beta_step_level_reward} verifies both conditions for beta regression with unknown scalar precision under step-level reward observation and derives explicit rates for unregularized and deterministically regularized reward estimators. This provides a concrete extension to bounded continuous rewards with jointly estimated mean and dispersion.

\section{Simulation Studies}
\label{sec:simulations}

We consider two settings: complete and partial reward observation. In the complete-observation setting, we examine binary and bounded continuous rewards with nonlinear conditional means. In the partial-observation setting, we assess whether transition-only trajectories improve policy learning when rewards are observed for only a subset of trajectories.

Let $d$ denote the state dimension. For each $h\in[H]$, the entries of $\theta_h^*\in\mathbb{R}^{d|\Asc|}$ are sampled independently from $\mathrm{Uniform}(-0.5,0.5)$. We index the actions by $\Asc=\{0,1,\ldots,|\Asc|-1\}$ and set $\phi_r=\phi_p=\phi$, where $\phi(x,a)\in\mathbb{R}^{d|\Asc|}$ places $x/\|x\|_2$ in the block corresponding to action $a$ and zeros elsewhere.

For both reward types, we use the logistic mean function $g(u)=\{1+\exp(-u)\}^{-1}$. Binary rewards are generated from $\mathrm{Bernoulli}\{m_h(x_h,a_h)\}$, while bounded continuous rewards are generated from a beta distribution with shape parameters $m_h(x_h,a_h)$ and $1-m_h(x_h,a_h)$.

For state transitions, we use rejection sampling. Candidate next states \(x'\) are sampled from $\mathrm{Uniform}(-0.5,0.5)^d$ and accepted with probability
\begin{equation*}
\alpha(x_h,a_h,x')
=
\left[
\frac{
\left\langle
(a_h+1)x_h+(a_h/d)\bone_d,
\exp(-x')
\right\rangle
}{
\sum_{j=1}^d\{(a_h+1)x'_j+a_h/d\}
}
\right]_0^1,
\end{equation*}
where \([u]_0^1=\min\{1,\max(0,u)\}\) and \(\bone_d\) is the \(d\)-dimensional vector of ones. We repeat the proposal step until an acceptance occurs and use the accepted proposal as \(x_{h+1}\).

The behavior policy selects the oracle optimal action with probability $70\%$ and a random action with probability $30\%$. The oracle action is computed from the data-generating MDP and is used only to generate the offline trajectories. In each simulation repetition, we generate 250 independent test trajectories under each learned policy and estimate its value by the average cumulative reward. Each configuration is repeated 100 times, and the figures report the average estimated policy value across repetitions.

For GRASP, binary and bounded continuous rewards are fitted by binomial GLM and beta regression, respectively. We set the transition-ridge parameter $\lambda=1$, $\xi=0.01$, and $c_r=c_p=c$. The common penalty multiplier $c$ is selected from $\{0.005,0.001,0.0005,0.0001\}$ by 5-fold cross-validation using a stepwise importance-sampling estimator \citep{precup2000eligibility,gottesman2019guidelines}.

\subsection{\texorpdfstring{Complete reward observation}{Complete reward observation}}

We vary the action-space size, state dimension, and number of episodes, with $|\Asc|\in\{2,3,4\}$, $d\in\{8,10,12\}$, and $n\in\{1000,1500,2000,2500\}$.

We compare four methods: GRASP, PEVI, Local Q-learning, and Global Q-learning. PEVI follows the linear-MDP construction of \citet{jin2020pessimism} and estimates Bellman targets by ridge regression under a linear approximation in $\phi(x,a)$. Local Q-learning fits a separate linear Q-function at each time step, whereas Global Q-learning fits one linear Q-function after pooling all time steps.

\begin{figure}[htbp]
    \centering
    \includegraphics[width=1\linewidth]{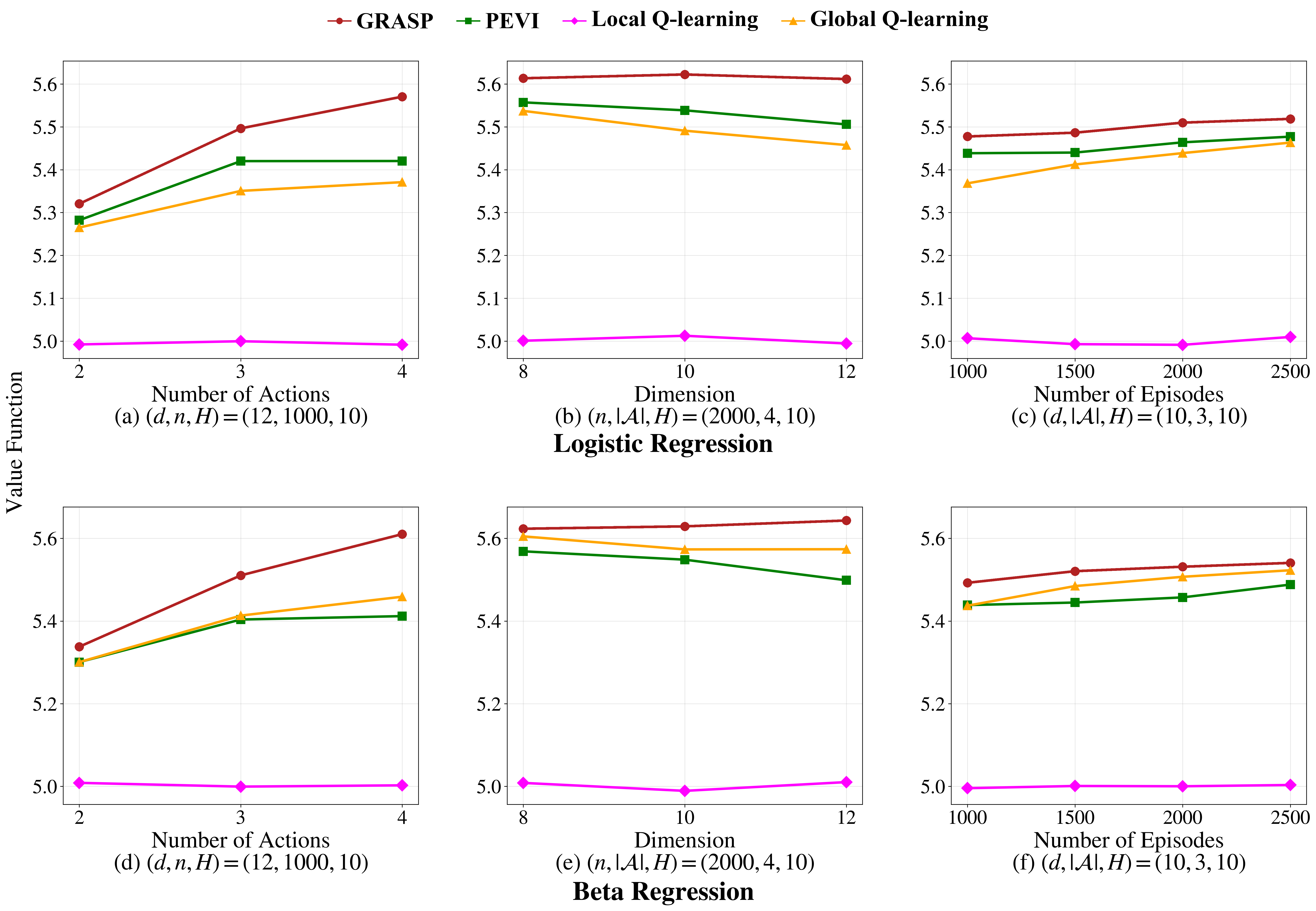}
    \caption{Average estimated policy value across 100 repetitions under complete reward observation. The top and bottom rows show binary and bounded continuous rewards, respectively. The columns vary the number of actions with $(d,n,H)=(12,1000,10)$, the state dimension with $(n,|\Asc|,H)=(2000,4,10)$, and the number of episodes with $(d,|\Asc|,H)=(10,3,10)$.}
    \label{fig:results_complete_reward_observation}
\end{figure}

Figure~\ref{fig:results_complete_reward_observation} compares the four methods as the action-space size, state dimension, and number of episodes vary. GRASP has the highest average estimated policy value in every displayed configuration, with its advantage over PEVI and Global Q-learning persisting across all three variations. The same qualitative pattern, with GRASP highest and Local Q-learning substantially lower, appears for both binary and bounded continuous rewards.

\subsection{\texorpdfstring{Partial reward observation}{Partial reward observation}}
\label{sec:simulations_partial_reward_observation}

We vary the observed-reward ratio \(n/(n+N)\), where \(n\) trajectories have observed rewards and the remaining \(N\) trajectories provide transition-only data. For binary rewards, we set \((d,n+N,|\Asc|,H)=(12,1000,4,8)\); for bounded rewards, we set \((12,1500,4,8)\). At each ratio, the reward-labeled trajectories are sampled uniformly before training.

We compare six methods: GRASP with full rewards, which treats all $n+N$ trajectories as reward observed; GRASP with reward-missing data, which uses the $n$ reward-labeled trajectories for reward estimation and all $n+N$ trajectories for transition estimation; and labeled-only versions of GRASP, PEVI, Local Q-learning, and Global Q-learning, each fitted using only the $n$ reward-labeled trajectories.

\begin{figure}[htbp]
    \centering
    \includegraphics[width=1\linewidth]{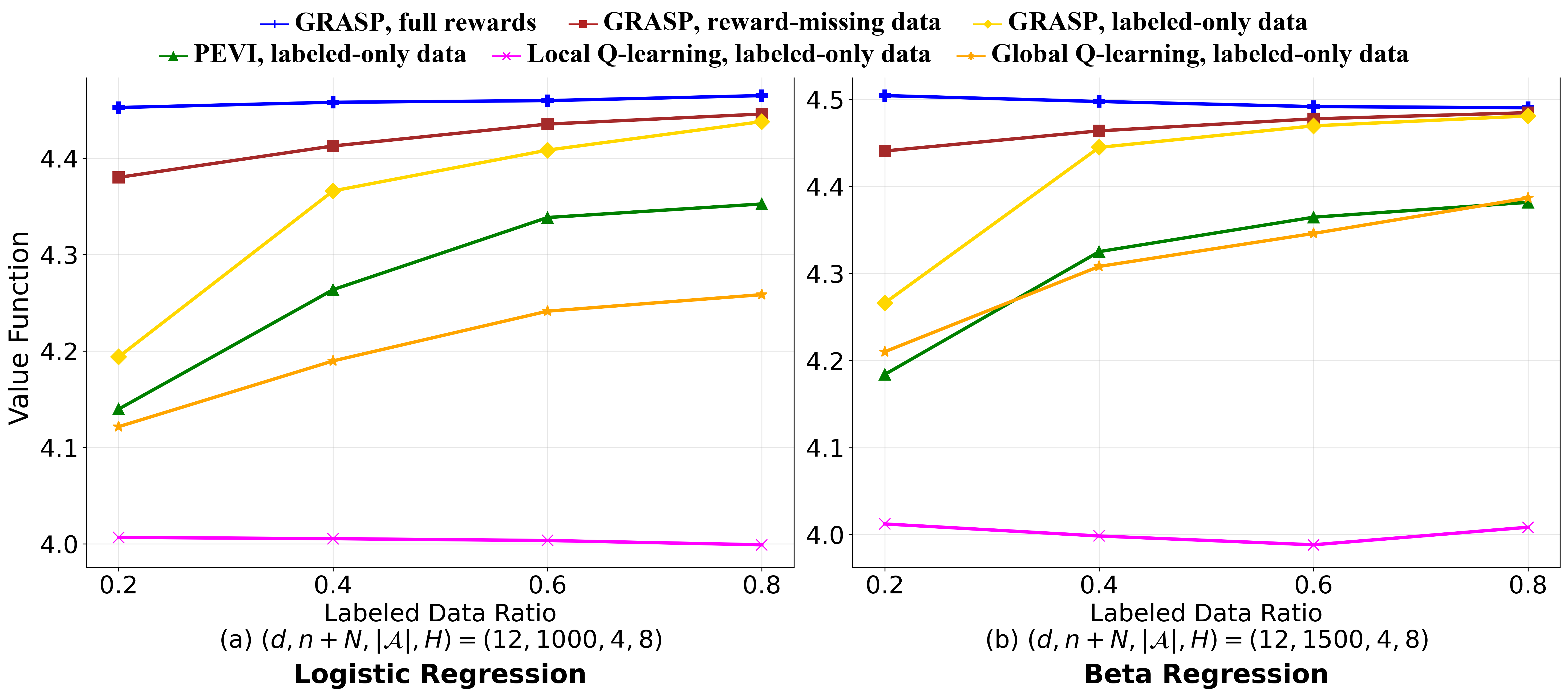}
    \caption{Average estimated policy value across 100 repetitions under partial reward observation as the observed-reward ratio varies. The left and right panels show binary and bounded continuous rewards, respectively.}
    \label{fig:results_partial_reward_observation}
\end{figure}

In Figure~\ref{fig:results_partial_reward_observation}, GRASP with full rewards provides an infeasible reference fitted as if rewards were observed for all $n+N$ trajectories. The average estimated value of GRASP with reward-missing data is closer to this reference as the labeled-data ratio increases. At every displayed ratio, it is higher than that of GRASP with labeled-only data, with the largest differences at low labeled-data ratios. Because these two methods use the same observed rewards and differ only in whether transition-only trajectories are retained, their comparison isolates the effect of retaining that transition information within GRASP. GRASP with reward-missing data also has a higher average estimated value than the labeled-only PEVI and Q-learning methods in all displayed settings.

\section{Optimizing DMT Decisions for MS}
\label{sec:real_data}

We apply GRASP-MDP to learn EDSS-based DMT decision policies for patients with MS. Evidence guiding longitudinal DMT sequencing, particularly decisions informed by disability outcomes, remains limited \citep{grand2018sequencing}. The timing of high-efficacy treatment is clinically relevant: observational studies have associated treatment started within 2 years of disease onset, or an early intensive rather than escalation strategy, with better subsequent disability outcomes \citep{he2020timing,harding2019clinical}. As clinical context for the policy analysis, we describe the timing of high-efficacy treatment observed in the study cohort.

For the analysis, we used an MS cohort curated from the EHRs at Mass General Brigham (MGB). We first applied a phenotyping algorithm \citep{Xiong2023KOMAP,gan2025knowledge} to identify $18,560$ patients with an MS diagnosis at a predefined specificity threshold of $90\%$ (MS.spec.90). Among these, $10,894$ patients were linked to the Comprehensive Longitudinal Investigation of Multiple Sclerosis at Brigham (CLIMB) registry \citep{gauthier2006model} via deidentified patient mapping. We retained the $9,495$ registry-linked patients with at least one structured EHR or natural language processing (NLP)-derived clinical note record. CLIMB provides relapse events and EDSS approximately every 6 months during registry follow-up \citep{CLIMB_EDSS2021}.

We identified DMT history using RxNorm prescription codes (Table~\ref{tab:dmt_classes}, Supplementary~\ref{sec:baseline_table}) and defined the index time $t=0$ as the date of the first recorded MS DMT prescription code. We required at least $2$ years of observable EHR follow-up after the index date and excluded patients with no recorded DMT prescription during this window, as well as three patients with missing baseline demographic characteristics. These criteria yielded a final cohort of $4,295$ patients.

Of the $4,295$ patients, $4,255$ had at least one documented EDSS during the $24$-month decision window and were split into a training set of $3,422$ patients (80\%) and a test set of 833 patients (20\%). EDSS observation remained intermittent within these patients: across the four study periods, $37.2\%$--$44.1\%$ of patient-periods had no observed EDSS. The remaining 40 patients had no EDSS records at any time during the decision window and were included as transition-only training patients. Cohort characteristics are summarized in Table~\ref{tab:baseline}.

\begin{table}[H]
\centering
\caption{Cohort characteristics. Values are mean (standard deviation; SD) for continuous variables and number (percentage) for categorical variables. The all-patient baseline EDSS summary excludes the 40 patients in the EDSS-unobserved group.}
\label{tab:baseline}
\footnotesize
\setlength{\tabcolsep}{3pt}
\begin{tabularx}{\linewidth}{>{\raggedright\arraybackslash}X *{4}{>{\centering\arraybackslash}p{2.5cm}}}
\toprule
Characteristic & All patients ($n=4,295$) & Train patients ($n=3,422$) & Test patients ($n=833$) & No EDSS ($n=40$) \\
\midrule
Female sex                         & 3,148 (73.3\%) & 2,501 (73.1\%) & 617 (74.1\%) & 30 (75.0\%) \\
Age at DMT index (years)           & 41.3 (11.5)    & 41.4 (11.5)    & 40.8 (11.6)  & 45.4 (12.2) \\
White race                         & 3,689 (85.9\%) & 2,934 (85.7\%) & 721 (86.6\%) & 34 (85.0\%) \\
Non-Hispanic ethnicity             & 4,247 (98.9\%) & 3,384 (98.9\%) & 824 (98.9\%) & 39 (97.5\%) \\
Baseline EDSS                     & 2.38 (2.11)    & 2.37 (2.11)    & 2.42 (2.15)  & --- \\
EHR follow-up (months)            & 143.2 (75.1)   & 144.1 (75.2)   & 142.4 (74.3) & 74.7 (46.9) \\
\bottomrule
\end{tabularx}
\end{table}

To align with the time resolution of EDSS in CLIMB, we summarized EHR data into four $6$-month periods, yielding horizon $H=4$. If multiple EDSS scores were recorded within a period, we used the latest score.

The state representation integrated heterogeneous EHR information, including structured clinical codes, concepts extracted from narrative notes, baseline demographic characteristics, and time-varying relapse history. It therefore combined coded and free-text information with both static patient characteristics and evolving disease activity. The resulting state vector had dimension $d=29$; details of feature screening, dimension reduction, and variable encoding are provided in Supplementary~\ref{sec:real_data_implementation}.

Within each period, we encoded DMT class as follows: $2$ if a high-efficacy DMT was prescribed; $1$ if a standard-efficacy DMT was prescribed without any high-efficacy DMT; and $0$ if no DMT was prescribed. For months without an active prescription record, we carried forward the most recent DMT class for up to $6$ months, because DMT prescriptions typically persist between recorded dispensing events. If no active prescription was observed within the $6$-month carry-forward window, the action was coded as $0$ (no active DMT observed).

We considered two EDSS-based rewards. Because higher EDSS indicates worse disability on a 0--10 scale, we reversed its direction for period-level rewards.
\begin{itemize}\tightlist
    \item \textbf{BETA}: A continuous reward defined as $r = 1 - \text{EDSS}/10$, clipped to $[0.001, 0.999]$.
    \item \textbf{BIO}: A binary reward indicating whether EDSS $\leq 4$. An EDSS score of 4.0 is a clinically interpretable ambulatory milestone, corresponding to a patient who is fully ambulatory without aid, self-sufficient, and able to walk approximately 500 meters despite relatively severe disability \citep{kurtzke1983rating}. We therefore use 4.0 as a prespecified cutoff.
\end{itemize}
Reporting BETA alongside BIO provides a continuous-outcome analysis that does not depend on a single EDSS cutoff.

\begin{figure}[htbp]
    \centering
    \includegraphics[width=1\linewidth]{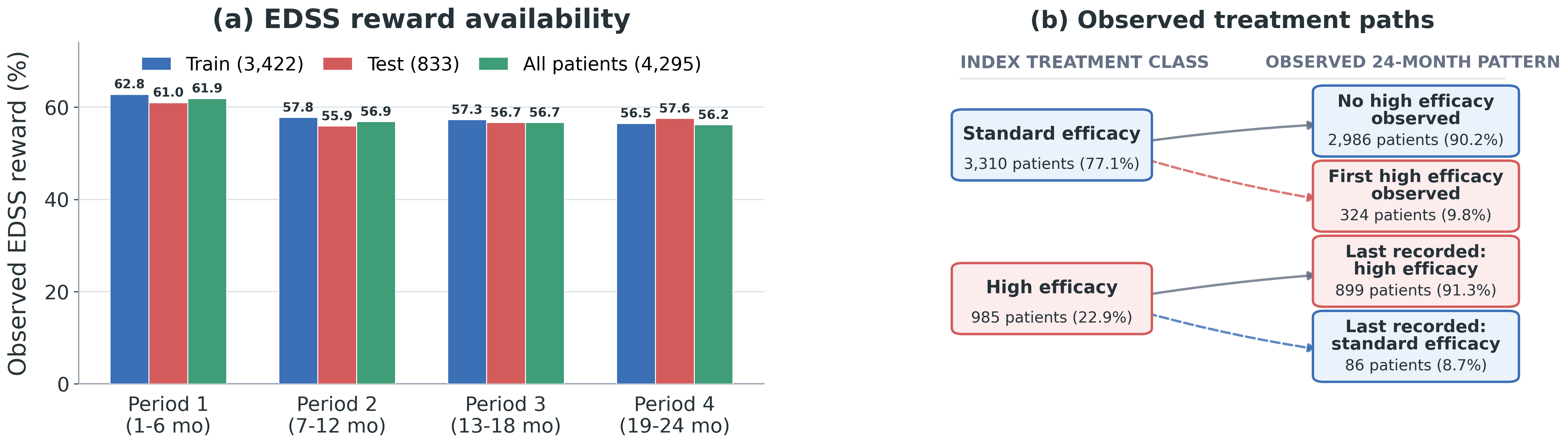}
    \caption{Study cohort summary. (a)~Per-step observed EDSS reward availability (\%) across the four $6$-month periods; the EDSS-unobserved group (40 patients) has $0\%$ observed-reward availability at all steps. (b)~Observed treatment paths during the $24$-month follow-up, shown separately for patients whose index DMT was standard efficacy and high efficacy. For the former, escalation is the first high-efficacy prescription; for the latter, the endpoint is the last explicitly recorded efficacy class.}
    \label{fig:cohort_summary}
\end{figure}

In the reconstructed real-world treatment record, 985 of 4,295 patients (22.9\%) initiated the study window with a high-efficacy DMT and 3,310 (77.1\%) with a standard-efficacy DMT. Among the latter, 324 (9.8\%) had a first high-efficacy prescription within 24 months, at a median of 11 months after the index DMT (interquartile range, 5--17 months). Thus, 1,309 patients (30.5\%) had any high-efficacy treatment recorded during the 24-month window. This time-to-high-efficacy measure is anchored to the first recorded DMT, not to MS onset. Figure~\ref{fig:cohort_summary} summarizes the step-specific availability of EDSS rewards and these observed DMT treatment paths.

We compare five learned policies: GRASP with reward-missing data, GRASP with labeled-only data, PEVI, Local Q-learning, and Global Q-learning. The proposed reward-missing procedure is the step-level GRASP method studied in Corollary~\ref{cor:step_level_reward_observation_rate} and Supplementary~\ref{sec:step_level_reward_observation}; it fits the reward model to observed EDSS at each step and the continuation component to all transitions from the 3,422 training patients and the 40 EDSS-unobserved patients. The labeled-only version uses the same observed rewards but discards transitions from patient-periods without observed EDSS; PEVI and the Q-learning methods are also fitted using only patient-periods with observed EDSS.

At each time step, we fitted a working beta-regression reward model with a logit mean link for BETA \citep{ferrari2004beta} and a binomial GLM with a logit link for BIO. Hyperparameter selection and implementation details for GRASP, the propensity models, and the baseline methods are provided in Supplementary~\ref{sec:real_data_implementation}.

We evaluated each learned policy on the held-out patient-periods using a period-specific inverse-probability-weighted score. At each period, the score uses patients whose recorded action matches the policy recommendation and whose EDSS is observed, with weights accounting for treatment assignment and EDSS observation. Separately, we estimated the mean outcome under recorded treatment as a descriptive reference using inverse EDSS-observation weights only, without reweighting the recorded treatment assignments. These estimates compare period-specific outcomes under the state distribution induced by recorded care; they do not estimate the value of following a learned policy over an intervened treatment trajectory. Full details of the estimands, estimators, identifying assumptions, and propensity models are provided in Supplementary~\ref{sec:real_data_implementation}.

\begin{figure}[htbp]
    \centering
    \includegraphics[width=0.86\linewidth]{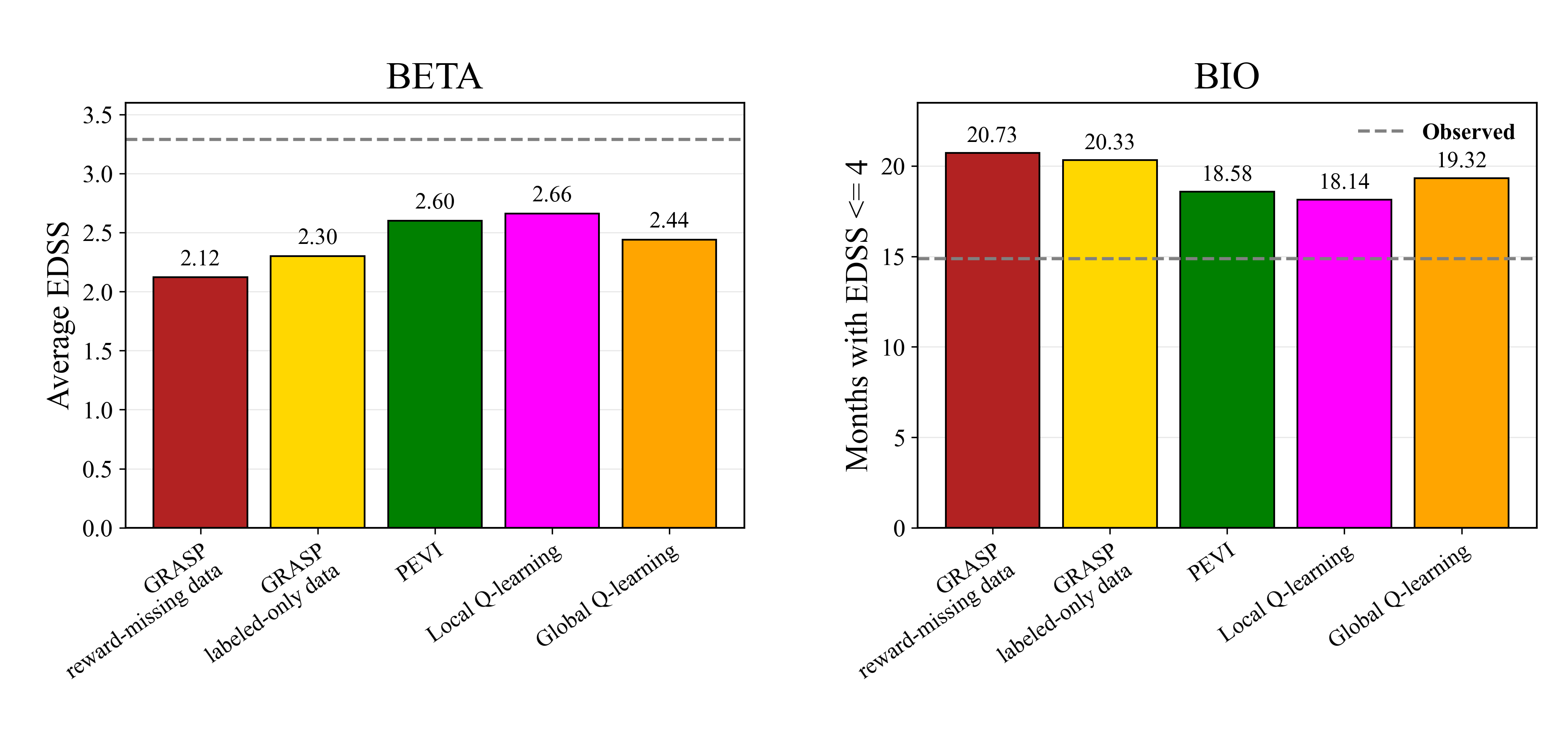}
    \caption{Estimated average EDSS over the 2-year follow-up (BETA; lower is better) and estimated months with EDSS $\leq 4$ (BIO; higher is better) for policies evaluated on held-out patient-periods. The dashed line shows the separately estimated recorded-treatment mean, adjusted for EDSS observation but not for treatment assignment.}
    \label{fig:real_data_results}
\end{figure}

Figure~\ref{fig:real_data_results} reports the point estimates on the clinically transformed scales. GRASP with reward-missing data has the numerically most favorable point estimate among the learned policies under both reward definitions: an average EDSS of $2.12$ under BETA and $20.73$ months with EDSS $\leq 4$ under BIO. The corresponding estimates for GRASP with labeled-only data are $2.30$ and $20.33$ months. The two versions differ only in whether transitions from patient-periods without observed EDSS are retained, so their comparison provides an empirical assessment of the contribution of retaining that transition information. PEVI and the Q-learning methods have less favorable point estimates than GRASP with reward-missing data in both analyses. The separately estimated recorded-treatment means are an average EDSS of $3.29$ and $14.87$ months with EDSS $\leq 4$. The learned-policy estimates are period-specific policy scores under the stated exchangeability, missing at random, positivity, and propensity-model assumptions; the recorded-treatment mean is a descriptive observed-care reference adjusted only for EDSS observation.

\section{Discussion and Conclusion}
\label{sec:conclusion}

Our findings support reward-transition separation as a useful design principle for offline RL with structured and intermittently observed outcomes. It preserves a tractable Bellman recursion while allowing transition-only observations to inform continuation values without imputing missing rewards. Such observations, however, cannot reduce reward-model uncertainty; their benefit is confined to transition estimation.

The framework also extends beyond the canonical-GLM formulation used in the main analysis. Supplementary~\ref{sec:m_estimator_extension} gives a high-level extension to M-estimated reward models, Supplementary~\ref{sec:beta_step_level_reward} provides a concrete step-level beta-regression result, Supplementary~\ref{sec:finite_mean_range} treats sub-Gaussian rewards with a finite conditional-mean range, and Supplementary~\ref{sec:relax} develops a regularized complete-observation formulation that relaxes the reward-curvature condition.

The method depends on the choice of feature maps, adequate coverage of relevant state-action regions, appropriate reward-model specification, and stable off-policy evaluation. The current guarantees do not cover nonignorable reward missingness or distribution shift between reward-labeled and transition-only trajectories; count rewards and other models outside the stated link and tail conditions also require separate concentration analyses.

The reward--transition separation is not specific to offline pessimism: the same decomposition supports online optimism through separate reward and transition bonuses. Supplementary~\ref{sec:online_appendix} develops the resulting GLSVI-UCB algorithm, establishes a sublinear regret guarantee, and evaluates it in an online simulation study. Future work could address nonignorable missingness and distribution shift and adapt the reward-transition decomposition to other online or offline linear-MDP algorithms \citep{xiong2022nearly,yang2020reinforcement}.

{\footnotesize
\bibliography{references}

@inproceedings{xu2025reinforcement,
  title={Reinforcement Learning on Dyads to Enhance Medication Adherence},
  author={Xu, Ziping and Jajal, Hinal and Choi, Sung Won and Nahum-Shani, Inbal and Shani, Guy and Psihogios, Alexandra M. and Hung, Pei-Yao and Murphy, Susan A.},
  booktitle={Artificial Intelligence in Medicine},
  series={Lecture Notes in Computer Science},
  volume={15734},
  pages={490--499},
  year={2025},
  publisher={Springer},
  doi={10.1007/978-3-031-95838-0_48},
  url={https://doi.org/10.1007/978-3-031-95838-0_48}
}

@inproceedings{xie2021bellman,
  title={Bellman-consistent pessimism for offline reinforcement learning},
  author={Xie, Tengyang and Cheng, Ching-An and Jiang, Nan and Mineiro, Paul and Agarwal, Alekh},
  booktitle={Advances in Neural Information Processing Systems},
  volume={34},
  pages={6683--6694},
  year={2021},
  url={https://proceedings.neurips.cc/paper/2021/hash/34f98c7c5d7063181da890ea8d25265a-Abstract.html}
}

@inproceedings{zheng2023semisupervisedofflinereinforcementlearning,
  title={Semi-supervised offline reinforcement learning with action-free trajectories},
  author={Zheng, Qinqing and Henaff, Mikael and Amos, Brandon and Grover, Aditya},
  booktitle={International Conference on Machine Learning},
  pages={42339--42362},
  year={2023},
  organization={PMLR}
}

@inproceedings{konyushkova2020semisupervisedrewardlearningoffline,
  title={Semi-supervised reward learning for offline reinforcement learning},
  author={Konyushkova, Ksenia and Zolna, Konrad and Aytar, Yusuf and Novikov, Alexander and Reed, Scott and Cabi, Serkan and de Freitas, Nando},
  booktitle={Offline Reinforcement Learning Workshop at NeurIPS},
  year={2020},
  url={https://offline-rl-neurips.github.io/pdf/32.pdf}
}

@article{gottesman2019guidelines,
  title={Guidelines for reinforcement learning in healthcare},
  author={Gottesman, Omer and Johansson, Fredrik and Komorowski, Matthieu and Faisal, Aldo and Sontag, David and Doshi-Velez, Finale and Celi, Leo Anthony},
  journal={Nature Medicine},
  volume={25},
  number={1},
  pages={16--18},
  year={2019},
  publisher={Nature Publishing Group US New York}
}

@inproceedings{yang2019sample,
  title={Sample-optimal parametric q-learning using linearly additive features},
  author={Yang, Lin and Wang, Mengdi},
  booktitle={International Conference on Machine Learning},
  pages={6995--7004},
  year={2019},
  organization={PMLR}
}

@article{mnih2015human,
  title={Human-level control through deep reinforcement learning},
  author={Mnih, Volodymyr and Kavukcuoglu, Koray and Silver, David and Rusu, Andrei A and Veness, Joel and Bellemare, Marc G and Graves, Alex and Riedmiller, Martin and Fidjeland, Andreas K and Ostrovski, Georg and others},
  journal={Nature},
  volume={518},
  number={7540},
  pages={529--533},
  year={2015},
  publisher={Nature Publishing Group},
  doi={10.1038/nature14236},
  url={https://doi.org/10.1038/nature14236}
}

@misc{levine2020offline,
  title={Offline reinforcement learning: Tutorial, review, and perspectives on open problems},
  author={Levine, Sergey and Kumar, Aviral and Tucker, George and Fu, Justin},
  journal={arXiv preprint arXiv:2005.01643},
  year={2020}
}

@inproceedings{duan2020minimax,
  title={Minimax-optimal off-policy evaluation with linear function approximation},
  author={Duan, Yaqi and Jia, Zeyu and Wang, Mengdi},
  booktitle={International Conference on Machine Learning},
  pages={2701--2709},
  year={2020},
  organization={PMLR}
}

@inproceedings{yang2020reinforcement,
  title={Reinforcement learning in feature space: Matrix bandit, kernels, and regret bound},
  author={Yang, Lin and Wang, Mengdi},
  booktitle={International Conference on Machine Learning},
  pages={10746--10756},
  year={2020},
  organization={PMLR}
}

@article{tropp2015introduction,
  title={An introduction to matrix concentration inequalities},
  author={Tropp, Joel A},
  journal={Foundations and Trends in Machine Learning},
  volume={8},
  number={1--2},
  pages={1--230},
  year={2015},
  doi={10.1561/2200000048},
  url={https://doi.org/10.1561/2200000048}
}

@article{jin2020pessimism,
  title={Is Pessimism Provably Efficient for Offline Reinforcement Learning?},
  author={Jin, Ying and Yang, Zhuoran and Wang, Zhaoran},
  journal={Mathematics of Operations Research},
  volume={50},
  number={4},
  pages={2738--2793},
  year={2024},
  doi={10.1287/moor.2022.0216},
  url={https://doi.org/10.1287/moor.2022.0216}
}

@book{mccullagh2019generalized,
  title={Generalized Linear Models},
  edition={2},
  author={McCullagh, Peter and Nelder, John A.},
  year={1989},
  publisher={Chapman \& Hall}
}

@inproceedings{jin2020provably,
  title={Provably efficient reinforcement learning with linear function approximation},
  author={Jin, Chi and Yang, Zhuoran and Wang, Zhaoran and Jordan, Michael I},
  booktitle={Conference on Learning Theory},
  pages={2137--2143},
  year={2020},
  organization={PMLR}
}

@inproceedings{wang2019optimism,
  title={Optimism in reinforcement learning with generalized linear function approximation},
  author={Wang, Yining and Wang, Ruosong and Du, Simon S and Krishnamurthy, Akshay},
  booktitle={International Conference on Learning Representations},
  year={2021},
  url={https://openreview.net/forum?id=CBmJwzneppz}
}

@incollection{vershynin2010introduction,
  title={Introduction to the non-asymptotic analysis of random matrices},
  author={Vershynin, Roman},
  booktitle={Compressed Sensing: Theory and Applications},
  editor={Eldar, Yonina C. and Kutyniok, Gitta},
  pages={210--268},
  publisher={Cambridge University Press},
  year={2012},
  doi={10.1017/CBO9780511794308.006},
  url={https://doi.org/10.1017/CBO9780511794308.006}
}

@inproceedings{NIPS2011_e1d5be1c,
 author = {Abbasi-yadkori, Yasin and P\'{a}l, D\'{a}vid and Szepesv\'{a}ri, Csaba},
 booktitle = {Advances in Neural Information Processing Systems},
 editor = {J. Shawe-Taylor and R. Zemel and P. Bartlett and F. Pereira and K. Q. Weinberger},
 pages = {},
 publisher = {Curran Associates, Inc.},
 title = {Improved Algorithms for Linear Stochastic Bandits},
 url = {https://proceedings.neurips.cc/paper/2011/file/e1d5be1c7f2f456670de3d53c7b54f4a-Paper.pdf},
 volume = {24},
 year = {2011}
}

@inproceedings{hu2023provable,
  title={The provable benefits of unsupervised data sharing for offline reinforcement learning},
  author={Hu, Hao and Yang, Yiqin and Zhao, Qianchuan and Zhang, Chongjie},
  booktitle={The Eleventh International Conference on Learning Representations},
  year={2023},
  url={https://openreview.net/forum?id=MTTPLcwvqTt}
}

@inproceedings{zanette2021provable,
  title={Provable benefits of actor-critic methods for offline reinforcement learning},
  author={Zanette, Andrea and Wainwright, Martin J and Brunskill, Emma},
  booktitle={Advances in Neural Information Processing Systems},
  volume={34},
  pages={13626--13640},
  year={2021}
}

@inproceedings{xiong2022nearly,
  title={Nearly minimax optimal offline reinforcement learning with linear function approximation: Single-agent mdp and markov game},
  author={Xiong, Wei and Zhong, Han and Shi, Chengshuai and Shen, Cong and Wang, Liwei and Zhang, Tong},
  booktitle={The Eleventh International Conference on Learning Representations},
  year={2023},
  url={https://openreview.net/forum?id=UP_GHHPw7rP}
}

@inproceedings{Wang2020Reinforcement,
author = {Wang, Ruosong and Salakhutdinov, Russ and Yang, Lin F.},
title = {Reinforcement Learning with General Value Function Approximation: Provably Efficient Approach via Bounded Eluder Dimension},
booktitle = {Advances in Neural Information Processing Systems},
volume = {33},
pages = {6123--6135},
year = {2020},
url = {https://proceedings.neurips.cc/paper/2020/hash/440924c5948e05070663f88e69e8242b-Abstract.html}
}

@article{khan2023adaptive,
  title={Adaptive Normalization for {IPW} Estimation},
  author={Khan, Samir and Ugander, Johan},
  journal={Journal of Causal Inference},
  volume={11},
  number={1},
  pages={20220019},
  year={2023},
  doi={10.1515/jci-2022-0019},
  url={https://doi.org/10.1515/jci-2022-0019}
}

@article{ross2022combining,
  title={Reflection on Modern Methods: Combining Weights for Confounding and Missing Data},
  author={Ross, Rachael K. and Breskin, Alexander and Breger, Tiffany L. and Westreich, Daniel},
  journal={International Journal of Epidemiology},
  volume={51},
  number={2},
  pages={679--684},
  year={2022},
  doi={10.1093/ije/dyab205},
  url={https://doi.org/10.1093/ije/dyab205}
}

@article{grand2018sequencing,
  title={Sequencing of high-efficacy disease-modifying therapies in multiple sclerosis: perspectives and approaches},
  author={Grand, Francois and Yeung, Michael and Morrow, Sarah A and Lee, Liesly and Emond, Francois and Ward, Brian J and Laneuville, Pierre and Schecter, Robyn and others},
  journal={Neural Regeneration Research},
  volume={13},
  number={11},
  pages={1871--1874},
  year={2018},
  publisher={Medknow},
  doi={10.4103/1673-5374.239433}
}

@article{he2020timing,
  title={Timing of high-efficacy therapy for multiple sclerosis: a retrospective observational cohort study},
  author={He, Anna and Merkel, Bernd and Brown, James William L and Zhovits Ryerson, Lana and Kister, Ilya and Malpas, Charles B and Sharmin, Sifat and Horakova, Dana and Havrdova, Eva Kubala and Spelman, Tim and others},
  journal={The Lancet Neurology},
  volume={19},
  number={4},
  pages={307--316},
  year={2020},
  doi={10.1016/S1474-4422(20)30067-3},
  url={https://doi.org/10.1016/S1474-4422(20)30067-3}
}

@article{harding2019clinical,
  title={Clinical outcomes of escalation vs early intensive disease-modifying therapy in patients with multiple sclerosis},
  author={Harding, Katharine and Williams, Owain and Willis, Mark and Hrastelj, Jane and Rimmer, Alice and Joseph, Fady and Tomassini, Valentina and Wardle, Mark and Pickersgill, Trevor and Robertson, Neil and others},
  journal={JAMA Neurology},
  volume={76},
  number={5},
  pages={536--541},
  year={2019},
  doi={10.1001/jamaneurol.2018.4905},
  url={https://doi.org/10.1001/jamaneurol.2018.4905}
}

@article{chard2017resolving,
  title={Resolving the clinico-radiological paradox in multiple sclerosis},
  author={Chard, Declan and Trip, S Anand},
  journal={F1000Research},
  volume={6},
  pages={1828},
  year={2017},
  doi={10.12688/f1000research.11932.1}
}

@article{kurtzke1983rating,
  title={Rating neurologic impairment in multiple sclerosis: an expanded disability status scale (EDSS)},
  author={Kurtzke, John F},
  journal={Neurology},
  volume={33},
  number={11},
  pages={1444--1452},
  year={1983},
  publisher={Lippincott Williams \& Wilkins},
  doi={10.1212/WNL.33.11.1444},
  url={https://doi.org/10.1212/WNL.33.11.1444}
}

@article{liu2021disease,
  title={Disease modifying therapies in relapsing-remitting multiple sclerosis: a systematic review and network meta-analysis},
  author={Liu, Zhuoyi and Liao, Qiao and Wen, Haicheng and Zhang, Yihao},
  journal={Autoimmunity Reviews},
  volume={20},
  number={6},
  pages={102826},
  year={2021},
  publisher={Elsevier},
  doi={10.1016/j.autrev.2021.102826}
}

@misc{Xiong2023KOMAP,
  title={Knowledge-driven online multimodal automated phenotyping system},
  author={Xiong, Xin and Sweet, Sara Morini and Liu, Molei and Hong, Chuan and Bonzel, Clara-Lea and Panickan, Vidul Ayakulangara and Zhou, Doudou and Wang, Linshanshan and Costa, Lauren and Ho, Yuk-Lam and others},
  howpublished={medRxiv preprint},
  year={2023},
  doi={10.1101/2023.09.29.23296239},
  url={https://www.medrxiv.org/content/10.1101/2023.09.29.23296239v1},
  note={Updated March 5, 2025}
}

@article{gauthier2006model,
  title={A model for the comprehensive investigation of a chronic autoimmune disease: the multiple sclerosis CLIMB study},
  author={Gauthier, Susan A and Glanz, Bonnie I and Mandel, Micha and Weiner, Howard L},
  journal={Autoimmunity Reviews},
  volume={5},
  number={8},
  pages={532--536},
  year={2006},
  publisher={Elsevier},
  doi={10.1016/j.autrev.2006.02.012},
  url={https://doi.org/10.1016/j.autrev.2006.02.012}
}

@article{CLIMB_EDSS2021,
author = {Brian C Healy and Bonnie I Glanz and Elyse Swallow and James Signorovitch and Kaitlin Hagan and Diego Silva and Corey Pelletier and Tanuja Chitnis and Howard Weiner},
title ={Confirmed disability progression provides limited predictive information regarding future disease progression in multiple sclerosis},

journal = {Multiple Sclerosis Journal – Experimental, Translational and Clinical},
volume = {7},
number = {2},
pages = {2055217321999070},
year = {2021},
doi = {10.1177/2055217321999070},
    note ={PMID: 33953937},
URL = {
        https://doi.org/10.1177/2055217321999070
},
eprint = {
        https://doi.org/10.1177/2055217321999070
}}

@inproceedings{li2010contextual,
  title={A Contextual-Bandit Approach to Personalized News Article Recommendation},
  author={Li, Lihong and Chu, Wei and Langford, John and Schapire, Robert E},
  booktitle={Proceedings of the 19th International Conference on World Wide Web},
  pages={661--670},
  year={2010},
  organization={ACM}
}

@article{russo2018tutorial,
  title={A Tutorial on {Thompson} Sampling},
  author={Russo, Daniel J and Van Roy, Benjamin and Kazerouni, Abbas and Osband, Ian and Wen, Zheng},
  journal={Foundations and Trends in Machine Learning},
  volume={11},
  number={1},
  pages={1--96},
  year={2018},
  publisher={Now Publishers},
  doi={10.1561/2200000070},
  url={https://doi.org/10.1561/2200000070}
}

@inproceedings{osband2016generalization,
  title={Generalization and Exploration via Randomized Value Functions},
  author={Osband, Ian and Van Roy, Benjamin and Wen, Zheng},
  booktitle={International Conference on Machine Learning},
  pages={2377--2386},
  year={2016},
  organization={PMLR}
}

@article{klasnja2015microrandomized,
  title={Micro-Randomized Trials: An Experimental Design for Developing Just-in-Time Adaptive Interventions},
  author={Klasnja, Predrag and Hekler, Eric B and Shiffman, Saul and Boruvka, Audrey and Almirall, Daniel and Tewari, Ambuj and Murphy, Susan A},
  journal={Health Psychology},
  volume={34},
  number={S},
  pages={1220--1228},
  year={2015},
  publisher={American Psychological Association},
  doi={10.1037/hea0000305},
  url={https://doi.org/10.1037/hea0000305}
}

@inproceedings{sawarni2024generalized,
  title={Generalized linear bandits with limited adaptivity},
  author={Sawarni, Ayush and Das, Nirjhar and Barman, Siddharth and Sinha, Gaurav},
  booktitle={Advances in Neural Information Processing Systems},
  volume={37},
  pages={8329--8369},
  year={2024},
  doi={10.52202/079017-0267},
  url={https://proceedings.neurips.cc/paper_files/paper/2024/hash/0faa0019b0a8fcab8e6476bc43078e2e-Abstract-Conference.html}
}

@inproceedings{filippi2010parametric,
 author = {Filippi, Sarah and Cappe, Olivier and Garivier, Aur\'{e}lien and Szepesv\'{a}ri, Csaba},
 booktitle = {Advances in Neural Information Processing Systems},
 editor = {J. Lafferty and C. Williams and J. Shawe-Taylor and R. Zemel and A. Culotta},
 pages = {},
 publisher = {Curran Associates, Inc.},
 title = {Parametric Bandits: The Generalized Linear Case},
 url = {https://proceedings.neurips.cc/paper_files/paper/2010/file/c2626d850c80ea07e7511bbae4c76f4b-Paper.pdf},
 volume = {23},
 year = {2010}
}

@inproceedings{li2017provably,
  title={Provably optimal algorithms for generalized linear contextual bandits},
  author={Li, Lihong and Lu, Yu and Zhou, Dengyong},
  booktitle={International Conference on Machine Learning},
  pages={2071--2080},
  year={2017},
  organization={PMLR}
}

@InProceedings{faury2020improved,
  title = 	 {Improved Optimistic Algorithms for Logistic Bandits},
  author =       {Faury, Louis and Abeille, Marc and Calauzenes, Clement and Fercoq, Olivier},
  booktitle = 	 {Proceedings of the 37th International Conference on Machine Learning},
  pages = 	 {3052--3060},
  year = 	 {2020},
  editor = 	 {III, Hal Daumé and Singh, Aarti},
  volume = 	 {119},
  series = 	 {Proceedings of Machine Learning Research},
  month = 	 {13--18 Jul},
  publisher =    {PMLR},
  url = 	 {https://proceedings.mlr.press/v119/faury20a.html}
}

@article{ferrari2004beta,
author = {Silvia Ferrari and Francisco Cribari-Neto},
title = {Beta Regression for Modelling Rates and Proportions},
journal = {Journal of Applied Statistics},
volume = {31},
number = {7},
pages = {799--815},
year = {2004},
publisher = {Taylor \& Francis},
doi = {10.1080/0266476042000214501},
URL = {https://doi.org/10.1080/0266476042000214501},
eprint = {https://doi.org/10.1080/0266476042000214501}
}

@inproceedings{precup2000eligibility,
author = {Precup, Doina and Sutton, Richard S. and Singh, Satinder P.},
title = {Eligibility Traces for Off-Policy Policy Evaluation},
year = {2000},
isbn = {1558607072},
publisher = {Morgan Kaufmann Publishers Inc.},
address = {San Francisco, CA, USA},
booktitle = {Proceedings of the Seventeenth International Conference on Machine Learning},
pages = {759–766},
numpages = {8},
series = {ICML '00}
}

@misc{gan2025knowledge,
  title={Knowledge Graph-Guided Identification of Multiple Sclerosis and Therapeutic Trend Analysis: Real-World Evidence from Two Large Healthcare Systems},
  author={Gan, Ziming and Zhu, Wen and Tang, Weijing and Sweet, Sara Morini and Morris, Michele and Han, Yunqing and Chen, Chenyi and Lu, Junwei and Song, Emily and Moro, Mohammed and others},
  howpublished={medRxiv preprint},
  year={2025},
  publisher={Cold Spring Harbor Laboratory Press},
  doi={10.1101/2025.11.12.25340116},
  url={https://www.medrxiv.org/content/10.1101/2025.11.12.25340116v1},
  note={Preprint}
}
}

\newpage
\appendix

\setcounter{section}{0}
\renewcommand{\thesection}{S\arabic{section}}
\renewcommand{\thesubsection}{\thesection.\arabic{subsection}}

\numberwithin{equation}{section}
\counterwithin{theorem}{section}
\counterwithin{corollary}{section}
\counterwithin{assumption}{section}
\counterwithin{lemma}{section}
\counterwithin{algorithm}{section}
\renewcommand{\thefigure}{S\arabic{figure}}
\renewcommand{\thetable}{S\arabic{table}}

\setcounter{figure}{0}
\setcounter{table}{0}

\makeatletter
\renewcommand{\theHsection}{supp.\arabic{section}}
\renewcommand{\theHsubsection}{supp.\arabic{section}.\arabic{subsection}}
\renewcommand{\theHequation}{supp.eq.\arabic{section}.\arabic{equation}}
\renewcommand{\theHtheorem}{supp.thm.\arabic{section}.\arabic{theorem}}
\renewcommand{\theHcorollary}{supp.cor.\arabic{section}.\arabic{corollary}}
\renewcommand{\theHassumption}{supp.assump.\arabic{section}.\arabic{assumption}}
\renewcommand{\theHlemma}{supp.lem.\arabic{section}.\arabic{lemma}}
\renewcommand{\theHalgorithm}{supp.alg.\arabic{section}.\arabic{algorithm}}
\renewcommand{\theHfigure}{supp.fig.\arabic{figure}}
\renewcommand{\theHtable}{supp.tab.\arabic{table}}
\@ifundefined{theHALG@line}{}{%
  \renewcommand{\theHALG@line}{supp.algline.\theHalgorithm.\arabic{ALG@line}}%
}
\makeatother

\phantomsection
\addcontentsline{toc}{section}{Supplementary Material}
\begin{center}
{\Large\bfseries Supplementary Material}\\[0.4em]
{\large\bfseries for \textit{Generalized Linear Markov Decision Process}}
\end{center}

\noindent
This document provides supplementary material for the manuscript \textit{Generalized Linear Markov Decision Process}. It contains additional algorithmic details, the online extension and experiment, DMT classification and real-data implementation details, the unbounded-reward extension, full proofs of the main theoretical results, supporting technical lemmas, and additional generalization results, including M-estimated reward models, step-level reward observation, a concrete beta-regression verification, a relaxation of the reward-curvature assumption, and the online regret proof. For compact proof notation, let \(\Isc_h^p=[n+N]\) denote the transition index set at every step; when \(N=0\), this reduces to \(\Isc_h^p=[n]\). The proofs are ordered by dependency so that shared arguments are stated once and then specialized through \(\Isc_h^r\) and \(\Isc_h^p\). The material is organized as follows.

\vspace{0.8em}

\begin{center}
\begingroup
\hypersetup{linkcolor={blue}}
\small
\newcommand{\suppcontentsline}[4]{%
  \noindent\hangindent=#2\hangafter=1
  \makebox[#2][l]{\textbf{\hyperref[#1]{#3}}}%
  \textbf{\hyperref[#1]{#4}}%
  \nobreak\leaders\hbox{\normalfont\kern0.35em.\kern0.35em}\hfill
  \nobreak\textbf{\hyperref[#1]{\pageref*{#1}}}\par}
\newcommand{\suppcontentsubline}[3]{%
  \noindent\hspace*{1.75em}%
  \makebox[4.7em][l]{\hyperref[#1]{#2}}%
  \hyperref[#1]{#3}%
  \nobreak\leaders\hbox{\normalfont\kern0.35em.\kern0.35em}\hfill
  \nobreak\hyperref[#1]{\pageref*{#1}}\par}
\noindent{\large\bfseries Contents}\par
\vspace{0.35em}
\hrule
\vspace{0.7em}
\setlength{\parskip}{0.2em}
\suppcontentsline{sec:offline_algorithms_appendix}{3.8em}{S1}{Offline Algorithm Details}
\suppcontentsline{sec:online_appendix}{3.8em}{S2}{Online Extension}
\suppcontentsubline{sec:online}{S2.1}{Online Algorithm}
\suppcontentsubline{sec:online_theory}{S2.2}{Regret Analysis}
\suppcontentsubline{sec:online_experiment}{S2.3}{Online Experiment}
\suppcontentsline{sec:baseline_table}{3.8em}{S3}{DMT Classification}
\suppcontentsubline{sec:real_data_implementation}{S3.1}{Real-Data Implementation and Policy Summaries}
\suppcontentsline{sec:finite_mean_range}{3.8em}{S4}{Sub-Gaussian Rewards with Finite Mean Range}
\suppcontentsline{sec:proofs_appendix}{3.8em}{S5}{Proofs}
\suppcontentsubline{sec:proof_bellman}{S5.1}{Bellman Completeness}
\suppcontentsubline{sec:shared_bellman_uncertainty}{S5.2}{Uniform Transition Calculation}
\suppcontentsubline{sec:proof_thm_partial_reward_observation}{S5.3}{GRASP Partial Reward-Observation Guarantees}
\suppcontentsubline{sec:proof_cor_complete_reward_observation_specialization}{S5.4}{Complete Reward-Observation Specializations}
\suppcontentsubline{sec:proof_cor_partial_reward_observation}{S5.5}{GRASP Partial Reward-Observation Rate Corollaries}
  \suppcontentsline{sec:generalization_results}{3.8em}{S6}{Generalization Results}
  \suppcontentsubline{sec:m_estimator_extension}{S6.1}{M-estimation Reward Models}
  \suppcontentsubline{sec:step_level_reward_observation}{S6.2}{Step-Level Reward Observation}
  \suppcontentsubline{sec:beta_step_level_reward}{S6.3}{Beta Regression with Step-Level Reward Observation}
  \suppcontentsubline{sec:relax}{S6.4}{Relaxing the Reward-Curvature Assumption}
  \suppcontentsubline{sec:proof_regret_online}{S6.5}{Online Regret Guarantee}
  \suppcontentsline{sec:technical_lemmas}{3.8em}{S7}{Technical Lemmas}
  \suppcontentsubline{lem:sigma}{S7.1}{Reward Information Concentration}
  \suppcontentsubline{lem:gradient_norm_bound}{S7.2}{Reward Score Concentration}
  \suppcontentsubline{lem:init}{S7.3}{Reward-Estimator Error}
  \suppcontentsubline{lem:g}{S7.4}{Reward Prediction Bound}
  \suppcontentsubline{lem:norm}{S7.5}{Bounded Bellman Coefficients}
  \suppcontentsubline{lem:concen_self_normalized}{S7.6}{Self-Normalized Concentration}
  \suppcontentsubline{lem:self_normalized_value_concentration}{S7.7}{Value-Process Concentration}
  \suppcontentsubline{lem:cover}{S7.8}{Euclidean Covering Number}
  \suppcontentsubline{lem:cover0}{S7.9}{Value-Class Covering Number}
  \suppcontentsubline{lem:observed_sample_information}{S7.10}{Observed-Sample Information under Missing at Random and Positivity}
  \suppcontentsubline{lem:rough_bound}{S7.11}{Regularization Bias Bound}
  \suppcontentsubline{lem:init1}{S7.12}{Regularized Reward-Estimator Error without Curvature}
  \suppcontentsubline{lem:g1}{S7.13}{Regularized Reward Prediction Bound without Curvature}
  \suppcontentsubline{lem:ucb}{S7.14}{Optimism for GLSVI-UCB}
  \suppcontentsubline{lem:theta_err_online}{S7.15}{Online Reward Estimation Error}
  \suppcontentsubline{lem:online_value_cover}{S7.16}{Online Value-Class Covering Number}
  \suppcontentsubline{lem:online_current_value_concentration}{S7.17}{Current-Value Uniform Concentration}
  \suppcontentsubline{lem:beta_err_online}{S7.18}{Online Transition Estimation Error}
  \suppcontentsubline{lem:concen}{S7.19}{Empirical Covariance Concentration}
\endgroup
\end{center}
\hypersetup{linkcolor={blue}}

\newpage

\section{Offline Algorithm Details}
\label{sec:offline_algorithms_appendix}

This section gives the step-level reward-observation version of the GRASP recursion in Section~\ref{sec:algorithms}.

\begin{algorithm}[H]
\caption[\texorpdfstring{GRASP for offline data with step-level reward observation}{GRASP for offline data with step-level reward observation}]{GRASP for offline data with step-level reward observation}
\label{alg:grasp_step_level_reward_observation}
\begin{algorithmic}[1]
\State Input: offline dataset \(\Dsc\); hyperparameters \(\lambda,\alpha_r,\alpha_p\).
\State Initialization: set \(\widehat V_{H+1}^{\rm obs}(x)\leftarrow 0\).
\For{step $h=H,H-1,\ldots,1$}
    \State Estimate \(\widetilde\theta_h^{\rm obs}\) using only the reward-observation set \(\Isc_h^r\).
    \State Estimate \(\widehat\beta_h\) from all \(n+N\) trajectories by the pooled transition regression in \eqref{est:beta_reward_observation}, with \(\widehat V_{h+1}^{\rm obs}\) replacing \(\widehat V_{h+1}\).
    \State Construct the step-level reward radius \(\widetilde\Gamma_{r,h}^{\rm obs}\) and set \(\widehat\Gamma_h^{\rm obs}=\widetilde\Gamma_{r,h}^{\rm obs}+\widehat\Gamma_{p,h}\).
    \State Form \(\widehat Q_h^{\rm obs}\) by the pessimistic update \eqref{eq:q_hat}, replacing \(g(\phi_r^\top\widetilde\theta_h)\) and \(\widehat\Gamma_h\) with \(g(\phi_r^\top\widetilde\theta_h^{\rm obs})\) and \(\widehat\Gamma_h^{\rm obs}\).
    \State Choose \(\widehat\pi_h^{\rm obs}\) greedily with respect to \(\widehat Q_h^{\rm obs}\) and set \(\widehat V_h^{\rm obs}(x)=\langle \widehat Q_h^{\rm obs}(x,\cdot),\widehat\pi_h^{\rm obs}(\cdot\mid x)\rangle_{\Asc}\).
\EndFor
\State Output: \(\widehat\pi^{\rm obs}=\{\widehat\pi_h^{\rm obs}\}_{h=1}^H\).
\end{algorithmic}
\end{algorithm}

\section{Online Extension}
\label{sec:online_appendix}

\subsection{Online Algorithm}
\label{sec:online}

Although the main text focuses on offline learning under partial reward observation, the GRASP-MDP structure also naturally yields an online optimistic value-iteration algorithm. The construction parallels the offline algorithms in Section~\ref{sec:algorithms}: at each step $h$, the reward component is estimated through the generalized linear reward model, while the transition component is estimated through a least-squares Bellman regression using the linear transition features. The main difference is the direction of uncertainty adjustment. Offline GRASP subtracts a pessimism penalty to avoid overestimating poorly covered actions, whereas the online algorithm adds upper-confidence bonuses to encourage exploration.

In the online RL setting, an agent interacts with the episodic MDP defined in Section~\ref{sec:problem} over episodes $t\in[T]$. At each episode, the agent starts from a fixed initial state $x$, follows a policy $\hat\pi_{\cdot,t}$, observes a trajectory $\{(x_{h,t},a_{h,t},r_{h,t},x_{h+1,t})\}_{h=1}^H$, and then updates the value estimates using all data collected up to episode $t$. Let $\Fsc_{t-1}$ denote the interaction history before episode $t$, so that $\widehat\pi_{\cdot,t}$ is $\Fsc_{t-1}$-measurable. We measure performance by the conditional pseudo-regret
\[
\mathcal{R}^{\rm ps}_T(x)
:=
\sum_{t=1}^T
\left\{V_1^*(x)-V_1^{\widehat\pi_{\cdot,t}}(x)\right\},
\]
where $V_1^{\widehat\pi_{\cdot,t}}(x)$ is the value in the underlying MDP conditional on the policy selected from $\Fsc_{t-1}$. Thus $\mathcal{R}^{\rm ps}_T(x)$ is random through the data-dependent policies, and the probability statement below is over the online interaction history.

Algorithm~\ref{alg:online} gives the resulting Generalized Least-Squares Value Iteration with upper confidence bounds (GLSVI-UCB). The reward bonus and transition bonus use the same two uncertainty components as GRASP across reward-availability settings, but are added rather than subtracted. The reward bonus is based on the self-normalized geometry of the reward feature covariance, while the transition bonus is based on the linear transition feature covariance.

\begin{algorithm}
\caption{GLSVI-UCB.}\label{alg:online}
\begin{algorithmic}[1]
\State Input: confidence parameters $
\gamma_r, \gamma_p,
$
and reward-parameter radius
\(
M
\).

\State Initialize
\(
\bar Q_{h,0}(x,a) \equiv H-h+1
\)
for all $h\leq H$, and
\(
\bar Q_{H+1,t}(x,a)\equiv 0
\)
for all $1 \leq t \leq T$.

\For{$t=1,2,\cdots,T$}
    \State Set the optimistic policy
    \[
    \hat{\pi}_{h,t}(x)
    \coloneqq
    \arg\max_{a\in\Asc} \bar Q_{h,t-1}(x,a)
    \]
    for each $h\in[H]$.

    \State Execute
    \(
    \hat{\pi}_{\cdot,t}
    \)
    from the fixed initial state $x$ and collect one trajectory
    \(
    \{(x_{h,t},a_{h,t},r_{h,t},x_{h+1,t})\}_{h=1}^H.
    \)

    \For{$h=H,H-1,\cdots,1$}
        \State Estimate the reward parameter by the GLM loss
        \begin{equation}\label{eq:glm_est}
\widehat \theta_{h,t}
\coloneqq
\arg \min_{\|\theta\|_2\le M}
\sum_{\tau=1}^{t}
\big(
-r_{h,\tau}
\langle
\phi_r(x_{h,\tau},a_{h,\tau}),
\theta
\rangle
+
G(
\langle
\phi_r(x_{h,\tau},a_{h,\tau}),
\theta
\rangle
)
\big).
\end{equation}

        \State Compute the empirical feature covariance matrices
        \[
        \Lambda_{p,h,t}
        \coloneqq
        \sum_{\tau=1}^t
        \phi_p(x_{h,\tau},a_{h,\tau})
        \phi_p(x_{h,\tau},a_{h,\tau})^\top
        +
        \bI_{d_p},
        \]
        \[
        \Lambda_{r,h,t}
        \coloneqq
        \sum_{\tau=1}^t
        \phi_r(x_{h,\tau},a_{h,\tau})
        \phi_r(x_{h,\tau},a_{h,\tau})^\top
        +
        \bI_{d_r}.
        \]

        \State Estimate the transition component by least-squares Bellman regression:
        \[
        \bar V_{h+1,t}(x)
        :=
        \max_{a\in\Asc}
        \bar Q_{h+1,t}(x,a), \quad
        \widehat\beta_{h,t}
        =
        \Lambda_{p,h,t}^{-1}
        \sum_{\tau=1}^{t}
        \phi_p(x_{h,\tau},a_{h,\tau})
        \bar V_{h+1,t}(x_{h+1,\tau}).
        \]

        \State Construct the optimistic Q-function
        \[
        \begin{aligned}
        \bar{Q}_{h,t}(x,a)
        \coloneqq
        \max\Bigg\{0,\,
        \min\Bigg\{
        &H-h+1, g\!\left(
        \phi_r(x,a)^\top\widehat{\theta}_{h,t}
        \right)
        +
        \phi_p(x,a)^\top\widehat \beta_{h,t}\\
        &\quad
        +
        \gamma_r
        \|\phi_r(x,a)\|_{\Lambda_{r,h,t}^{-1}}
        +
        \gamma_p
        \|\phi_p(x,a)\|_{\Lambda_{p,h,t}^{-1}}
        \Bigg\}
        \Bigg\}.
        \end{aligned}
        \]
    \EndFor
\EndFor

\State Output: optimistic policies
\(
\{\hat\pi_{\cdot,t}\}_{t=1}^T.
\)

\end{algorithmic}
\end{algorithm}

\subsection{Regret Analysis}
\label{sec:online_theory}

The regret analysis follows the same reward-transition decomposition as the offline analysis, with optimism replacing pessimism. The first confidence term controls the error of the generalized linear reward estimator, and the second controls the Bellman regression error associated with the linear transition component. Together, these two terms ensure that $\bar Q_{h,t}$ is an upper confidence bound for $Q_h^*$ with high probability.
\begin{assumption}\label{assump:theta_online}
   For any $h\in [H]$, $\|\theta_h^*\|_2\leq M$ for some $M\geq1$ and $\big \|\mu_{h}(\Ssc)\big\|\leq \sqrt{d_p}$.
\end{assumption}
\begin{assumption}\label{assump:g_online}
    There exist constants $0<k<K<\infty$ such that $k\leq \dot{g}(\langle \phi_r(x,a),\theta\rangle)\leq K$ for all $x\in \Ssc$, $a\in \Asc$ and $\|\theta \|_2\le M$.
\end{assumption}
\begin{remark}\label{rmk:self_concordance}
    The global positive lower bound on $\dot g$ can be replaced under a self-concordance-type condition. In particular, suppose that $0<\dot g\le K$,
    \[
   \inf_{x\in \Ssc,a\in \Asc,h\in [H]}\dot{g}(\langle\phi_r(x,a),\theta_h^*\rangle)\ge k_{\mathrm{loc}}>0,
    \]
    and
    \[
    |\ddot{g}|\le R\dot{g}
    \]
    for some $R>0$. Because $|(\log \dot g)'|\le R$, the feature normalization and Assumption~\ref{assump:theta_online} imply that, for every $\theta$ in the optimization domain $\Theta_M=\{\theta\in\R^{d_r}:\|\theta\|_2\le M\}$,
    \[
    \dot g(\langle\phi_r(x,a),\theta\rangle)
    \ge k_{\mathrm{loc}}\exp(-2RM)
    \eqqcolon k_{\mathrm{eff}}.
    \]
    Hence the same proof and regret bound hold with $k_{\mathrm{eff}}$ in place of the global constant $k$ in Assumption~\ref{assump:g_online}.
\end{remark}

The following theorem shows that GLSVI-UCB inherits the two-component complexity of GRASP-MDPs. The first term is driven by reward learning under the generalized linear model, while the second term is driven by transition learning under the linear transition model.

For $t\in[T]$ and $p\in(0,1)$, write
\[
A_t(p)\coloneqq d_r\log(4Mt)+\log(TH/p).
\]
For a fixed numerical constant $c_p\ge1$, also write
\[
\zeta_{\rm on}(p)
:=
\log\!\left(
\frac{3(c_p+1)(d_r+d_p)(1+M+K+k^{-1})TH}{p}
\right).
\]

\begin{theorem}[Regret for GLSVI-UCB]\label{thm:regret_online}
Under the standing bounded-reward and feature-normalization conditions and Assumptions \ref{assump:theta_online} and \ref{assump:g_online}, for any fixed $p_0\in(0,1)$, if we set
    \[
    \gamma_r=K\sqrt{4M^2+\frac{4+8\sqrt{A_T(p_0/3)}}{k}+\frac{16A_T(p_0/3)}{k^2}}
    \quad\text{and}\quad
    \gamma_p=c_p(d_r+d_p)H\sqrt{\zeta_{\rm on}(p_0)},
    \]
    where $c_p>0$ is sufficiently large, then, for any fixed $x\in \Ssc$,
    \begin{align*}
     \mathcal{R}^{\rm ps}_T(x)&\leq H+2H\sqrt{T}\Big(\gamma_r\sqrt{2d_r\log(1+T/d_r)}+\gamma_p\sqrt{2d_p\log(1+T/d_p)}\Big)+\sqrt{2\log (6/p_0)TH^3}\\
     &=O\Bigg(H\sqrt{T\,d_r\log(1+T/d_r)}\,K\sqrt{M^2+\frac{1+\sqrt{A_T(p_0/3)}}{k}+\frac{A_T(p_0/3)}{k^2}}\\
     &+\sqrt{TH^4d_p(d_r+d_p)^2\zeta_{\rm on}(p_0)\log(1+T/d_p)}\Bigg)\\
     &=\widetilde{O}\Big(\sqrt{TH^2d_r^2}+\sqrt{TH^4d_p(d_r+d_p)^2}\Big)
    \end{align*}
    with probability at least $1-p_0$.
\end{theorem}

\subsection{Online Experiment}
\label{sec:online_experiment}

To evaluate the practical performance of GLSVI-UCB (Algorithm~\ref{alg:online}) in the online RL setting, we conduct a simulation study motivated by the ADAPTS-HCT study design described by \citet{xu2025reinforcement}, which targets dyadic mobile health (mHealth) interventions for enhancing medication adherence among adolescent and young adult (AYA) patients after hematopoietic cell transplantation (HCT). The complete simulator specification, including the data-generating equations, feature construction, and calibration procedure, is provided by \citet{xu2025reinforcement}; here we summarize only the components needed to describe our online comparison.

The simulation environment models dyadic interactions between AYA patients and their caregivers over a 14-week trial period ($W=14$). At each decision point, the agent selects actions from multiple intervention components. The environment dynamics are calibrated from the ROADMAP dataset using generalized estimating equation (GEE) models fitted to real clinical data. The state representation has dimension $d=11$, capturing medication adherence, psychological distress, treatment burden for both the AYA and caregiver, the dyadic relationship quality, and temporal features.

We evaluate algorithms under two experimental settings:
\begin{itemize}\tightlist
    \item \textbf{Target-Only}: Only the AYA-targeted intervention component is optimized by the learning algorithm, while the other two components (game and caregiver) follow a fixed micro-randomized trial (MRT) policy with randomization probability $p=0.5$.
    \item \textbf{3-Action}: All three intervention components---AYA target, game, and caregiver---are simultaneously optimized by the learning algorithm.
\end{itemize}

In each setting, $T=25$ dyads are sequentially recruited and followed over $W=14$ weeks. Each experiment is replicated 100 times.

We compare GLSVI-UCB against a comprehensive set of baseline algorithms spanning contextual bandits, value-based RL, and fixed-randomization designs:
\begin{itemize}\tightlist
    \item \textbf{Linear Upper Confidence Bound (LinUCB)} \citep{li2010contextual}: A contextual bandit algorithm that maintains a linear model of the expected reward and selects actions using upper confidence bound exploration. We evaluate the standard UCB rule and our $\epsilon$-greedy action-selection variant.
    \item \textbf{LSVI-UCB} \citep{jin2020provably}: Least-Squares Value Iteration with UCB exploration, a standard linear MDP algorithm using a single feature map for both reward and transition. We evaluate our softmax and $\epsilon$-greedy action-selection variants.
    \item \textbf{Bayesian Thompson Sampling (Bayes Pool)} \citep{russo2018tutorial}: Our fully pooled Bayesian Thompson-sampling baseline, which samples reward parameters from their posterior distribution and pools information across dyads.
    \item \textbf{Bandit Thompson Sampling (TS; Pool / No Pool)} \citep{russo2018tutorial}: Our contextual-bandit Thompson-sampling implementations with and without information pooling across dyads.
    \item \textbf{RLSVI-H} \citep{osband2016generalization}: Our finite-horizon implementation of Randomized Least-Squares Value Iteration, denoted RLSVI-H, which uses posterior sampling for exploration in episodic MDPs.
    \item \textbf{Reward Learning}: A model-based approach that learns the reward function and selects actions to maximize learned expected rewards.
    \item \textbf{MRT ($p=0.5$)} \citep{klasnja2015microrandomized}: A fixed-randomization design where each action is selected with probability $0.5$, serving as the non-adaptive baseline.
\end{itemize}

\begin{figure}[htbp]
    \centering
    \includegraphics[width=1\linewidth]{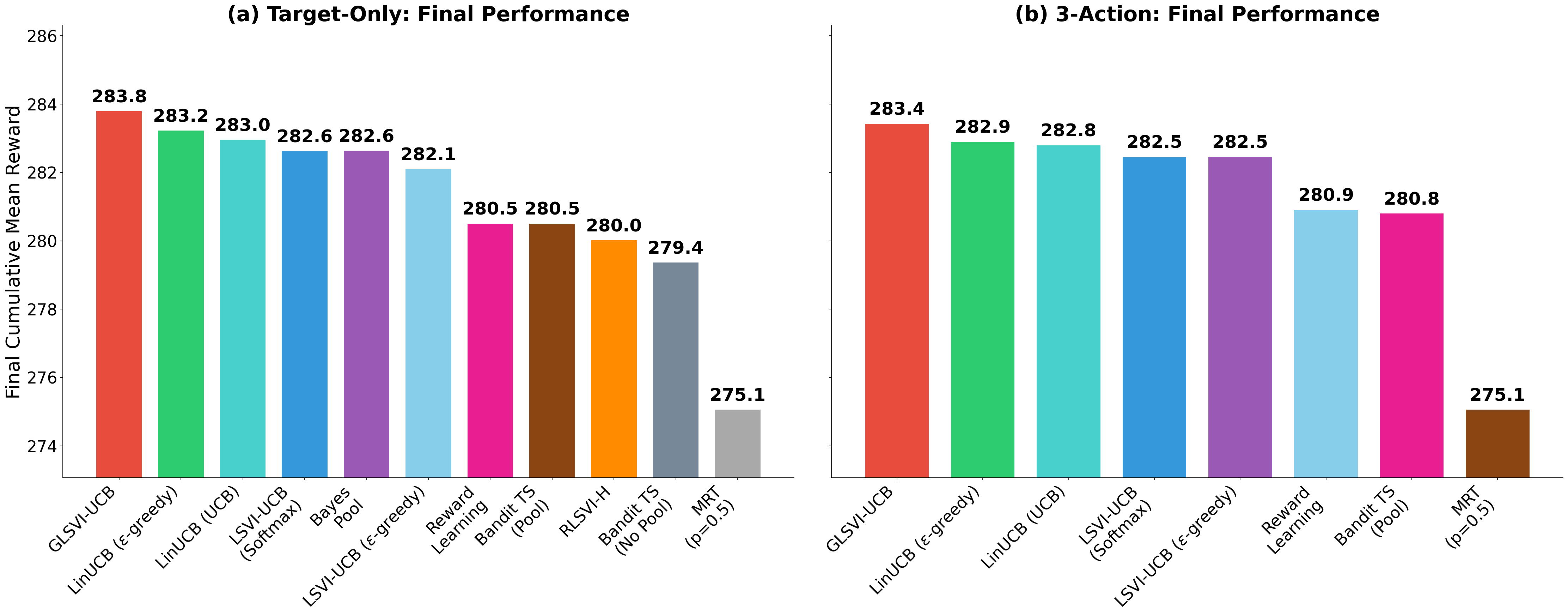}
    \caption{Final cumulative mean reward across all algorithms in the ADAPTS-HCT simulation environment. (a) Target-Only setting where only the AYA intervention component is optimized. (b) 3-Action setting where all three intervention components are simultaneously optimized. GLSVI-UCB achieves the highest cumulative reward in both settings.}
    \label{fig:online_barplot}
\end{figure}

Figure~\ref{fig:online_barplot} presents the final cumulative mean reward achieved by each algorithm across both experimental settings. In the Target-Only setting (Figure~\ref{fig:online_barplot}a), GLSVI-UCB achieves the highest cumulative reward of $283.8$, outperforming all baselines. The closest competitors are LinUCB ($\epsilon$-greedy) at $283.2$ and LinUCB (UCB) at $283.0$, followed by LSVI-UCB (Softmax) and Bayes Pool at $282.6$. The fixed MRT baseline achieves the lowest reward of $275.1$, demonstrating the substantial benefit of adaptive learning.

In the 3-Action setting (Figure~\ref{fig:online_barplot}b), where all three intervention components are simultaneously optimized, GLSVI-UCB again achieves the best performance with a cumulative reward of $283.4$. The performance advantage over the best contextual bandit baseline (LinUCB $\epsilon$-greedy at $282.9$) remains consistent, suggesting that GLSVI-UCB effectively leverages the transition structure in the environment to improve policy optimization. Notably, the gap between adaptive algorithms and the fixed MRT baseline ($275.1$) is substantial in both settings, underscoring the value of online learning for personalizing mHealth interventions.

\section{DMT Classification}
\label{sec:baseline_table}

For cohort construction, we grouped disease-modifying therapies (DMTs) into high-efficacy and standard-efficacy classes using the medication names and RxNorm identifiers listed in Table~\ref{tab:dmt_classes}.

\begin{table}[H]
\centering
\caption{DMT Classes and Medication Codes Used for Cohort Construction.}
\label{tab:dmt_classes}
\footnotesize
\setlength{\tabcolsep}{4pt}
\begin{tabularx}{\linewidth}{l >{\raggedright\arraybackslash}X >{\raggedright\arraybackslash}X c}
\toprule
Generic name & Brand name(s) & Drug class & RxNorm ID \\
\midrule
\multicolumn{4}{l}{\textbf{High-efficacy}} \\
ocrelizumab & Ocrevus & B cell depletion & 1876366 \\
ofatumumab & Arzerra / Kesimpta & B cell depletion & 712566 \\
rituximab & Rituxan & B cell depletion & 121191 \\
ublituximab & Briumvi & B cell depletion & 2626349 \\
natalizumab & Tysabri / Tyruko & alpha4-integrin blocker & 354770 \\
alemtuzumab & Lemtrada / Campath & chemotherapy & 117055 \\
mitoxantrone & Novantrone & cell proliferation blocker & 7005 \\
cladribine & Mavenclad & purine analog & 44157 \\
\addlinespace
\multicolumn{4}{l}{\textbf{Standard-efficacy}} \\
dimethyl fumarate & Tecfidera & fumarate & 1373478 \\
diroximel fumarate & Vumerity & fumarate & 2261783 \\
monomethyl fumarate & Bafiertam & fumarate & 1546433 \\
glatiramer acetate & Copaxone / Glatopa & glatiramer acetate & 214582 \\
interferon beta-1a & Avonex / Rebif & interferon-beta & 75917 \\
interferon beta-1b & Betaseron / Extavia & interferon-beta & 72257 \\
peginterferon beta-1a & Plegridy & interferon-beta & 1546168 \\
fingolimod & Gilenya / Tascenso & S1P receptor modulator & 1012892 \\
ozanimod & Zeposia & S1P receptor modulator & 2288236 \\
ponesimod & Ponvory & S1P receptor modulator & 2532300 \\
siponimod & Mayzent & S1P receptor modulator & 2121085 \\
teriflunomide & Aubagio & pyrimidine blocker & 1310520 \\
\bottomrule
\end{tabularx}
\end{table}

\subsection{Real-Data Implementation and Policy Summaries}
\label{sec:real_data_implementation}

\noindent\textbf{Treatment definitions and timing.}
The observed, or ``real-world,'' treatment is the DMT class reconstructed from the EHR prescription record using Table~\ref{tab:dmt_classes}. In each month, a recorded high-efficacy DMT takes precedence over a standard-efficacy DMT; the latest class is carried forward for at most 6 months, after which treatment is coded as no active DMT observed. The four decision actions summarize months 1--6, 7--12, 13--18, and 19--24 after the first recorded DMT. Time to high efficacy is the number of months from that index prescription to the first month classified as high efficacy. It describes the available prescription record and is not necessarily time since MS onset.

\noindent\textbf{State construction.}
Through a knowledge network \citep{Xiong2023KOMAP}, we selected $971$ EHR features relevant to MS, including 287 structured EHR codes and 684 narrative concepts, and summarized these features by monthly counts. The 49 RxNorm medication codes were reserved for action encoding; the remaining 922 features were used as candidate state covariates.

Prior to dimensionality reduction, we defined prevalence as the fraction of non-zero patient-months and removed features with prevalence below 5\%, retaining $50$ clinically prevalent covariates. We then applied principal component analysis (PCA) to project the EHR features into a lower-dimensional latent space, retaining $24$ components that explain $90.5\%$ of the total variance.

In addition to the 24 EHR-derived PCA components, the state included four baseline demographic variables, sex, race, ethnicity, and age at index ($t=0$), and one time-varying clinical variable, a relapse indicator for each period defined as whether any relapse occurred within the 6-month window. Sex, ethnicity, and relapse were encoded as binary indicators. Because White patients accounted for the majority of the cohort, race was dichotomized as White versus non-White and encoded as a binary indicator. Age at index was retained as a continuous variable. After dimension reduction and clinical-variable augmentation, the total state dimension is $d = 24 + 4 + 1 = 29$.

\noindent\textbf{Model fitting and tuning.}
We compared GRASP with reward-missing data, GRASP with labeled-only data, PEVI, Local Q-learning, and Global Q-learning. The reward-missing GRASP procedure used patient-periods with observed EDSS for reward estimation and all transitions from the 3,422 training patients and the 40 fully EDSS-unobserved patients for continuation estimation. The other four methods used only patient-periods with observed EDSS. For the GRASP procedures, we selected the pessimism parameter \(c\) from \(\{0,0.005,0.001,0.0005,0.0001\}\) using 5-fold cross-validation with the period-specific policy score defined below. We fixed the transition-ridge parameter \(\lambda=1\) and set \(\xi=0.01\), as in the simulation studies. At each period, the reward model was a beta regression with logit mean link and scalar precision for BETA and a binomial GLM with logit link for BIO, with a separate \(\ell_2\) penalty selected by 5-fold cross-validation to maximize held-out log-likelihood. The propensity model used \(\ell_2\)-regularized multinomial logistic regression with a cross-validation-selected penalty; the Q-learning baselines used leave-one-out cross-validation-selected ridge regularization.

Supplementary~\ref{sec:beta_step_level_reward} gives finite-sample guarantees for an exact conditional beta model when step-level rewards are missing at random, without regularization and with a prespecified deterministic penalty. Because the clipped EDSS transformation takes values on a finite set, beta regression is used here as a continuous working approximation. The data-adaptive reward penalties and pessimism parameter are practical tuning choices rather than objects covered by those guarantees.

\noindent\textbf{Period-specific policy evaluation.}
Let \((X_h,A_h,R_h,O_h)\) denote the state, recorded action, reward, and reward-observation indicator at period \(h\), and let \(R_h(a)\) denote the potential period-$h$ reward under action $a$. For a deterministic evaluation policy, the period-specific target is
\[
\Psi_{\mathrm{period}}(\pi)
=
\sum_{h=1}^H
\E_{X_h\sim d_h^b}
\left[
\E\{R_h\mid X_h,A_h=\pi_h(X_h)\}
\right],
\]
where \(d_h^b\) is the distribution of \(X_h\) under the observed clinician policy. For a stochastic evaluation policy, the inner conditional mean is replaced by \(\sum_{a\in\Asc}\pi_h(a\mid X_h)\E\{R_h\mid X_h,A_h=a\}\). Thus, \(\Psi_{\mathrm{period}}(\pi)\) is a period-specific policy score, not the standard MDP value under trajectories induced by \(\pi\).

For a test set of \(n_{\rm test}\) patients, we estimated this target using
\begin{equation*}
\widehat\Psi_{\mathrm{period}}(\pi)
=
\sum_{h=1}^{H}
\frac{\sum_{i=1}^{n_{\rm test}} w_h^{(i)}r_h^{(i)}}
{\sum_{i=1}^{n_{\rm test}} w_h^{(i)}},
\qquad
w_h^{(i)}
=
\frac{\mathbf{1}\{\pi_h(x_h^{(i)})=a_h^{(i)}\}o_h^{(i)}}
{\max\{\hat e_h(a_h^{(i)},1\mid x_h^{(i)}),10^{-3}\}} .
\end{equation*}
This H\'ajek/self-normalized form uses the joint treatment-observation propensity \(\hat e_h(a,1\mid x)=\Pr(A_h=a,O_h=1\mid X_h=x)\) \citep{khan2023adaptive,ross2022combining}. We factorized it as \(P(a,o\mid x)=P(a\mid x)P(o\mid a,x)\), fitting period-specific regularized multinomial treatment models and logistic observation models with standardized inputs. The same fitted propensity models were used during hyperparameter selection and final evaluation. The target estimands are defined using the corresponding untruncated propensities; the lower bound of \(10^{-3}\) is a finite-sample numerical stabilization device.

Under consistency, period-specific conditional exchangeability \(R_h(a)\perp A_h\mid X_h\), missing at random \(R_h\perp O_h\mid(X_h,A_h)\), and positivity for recommended actions,
\[
\E\left[
\frac{\mathbf 1\{A_h=\pi_h(X_h)\}O_hR_h}
{e_h(A_h,1\mid X_h)}
\right]
=
\E_{X_h\sim d_h^b}
\left[\E\{R_h\mid X_h,A_h=\pi_h(X_h)\}\right].
\]
This identity does not identify the value of intervening to follow \(\pi\) over the full trajectory because earlier actions are not reweighted. Accordingly, Table~\ref{tab:bio_action_distribution} compares learned and recorded recommendations at held-out states generated under recorded care; these are neither simulated treatment trajectories nor causal estimates of replacing clinician decisions.

The recorded-treatment reference in Figure~\ref{fig:real_data_results} was
estimated separately rather than by substituting each patient's recorded
action into the learned-policy estimator. Its period-specific target is
\[
\Psi_{\rm recorded}
=
\sum_{h=1}^H
\E_{X_h\sim d_h^b}
\left[
\sum_{a\in\Asc}\pi_h^b(a\mid X_h)
\E\{R_h\mid X_h,A_h=a\}
\right].
\]
We estimated it by
\[
\widehat\Psi_{\rm recorded}
=
\sum_{h=1}^H
\frac{\sum_{i=1}^{n_{\rm test}}v_h^{(i)}r_h^{(i)}}
{\sum_{i=1}^{n_{\rm test}}v_h^{(i)}},
\qquad
v_h^{(i)}
=
\frac{o_h^{(i)}}
{\max\{\widehat p_h^{\rm obs}(x_h^{(i)},a_h^{(i)}),10^{-3}\}},
\]
where \(\widehat p_h^{\rm obs}(x,a)\) is the fitted model for
\(\Pr(O_h=1\mid X_h=x,A_h=a)\).
Thus the recorded-treatment mean corrects for EDSS observation without
reweighting the empirical distribution of recorded treatments. It is a
descriptive observed-care reference, not an estimate of a longitudinal
dynamic clinician-policy value.

\begin{table}[H]
\centering
\caption{Action distributions for recorded care and GRASP with reward-missing data under the BIO reward setting at held-out period-specific states (\(n=833\)). Each cell reports count (row percentage).}
\label{tab:bio_action_distribution}
\footnotesize
\setlength{\tabcolsep}{4pt}
\begin{tabular}{clccc}
\toprule
Period & Action source & No active DMT observed & Standard efficacy & High efficacy \\
\midrule
1 (months 1--6) & Recorded clinician & 80 (9.6\%) & 558 (67.0\%) & 195 (23.4\%) \\
                 & GRASP, reward-missing & 48 (5.8\%) & 760 (91.2\%) & 25 (3.0\%) \\
\addlinespace
2 (months 7--12) & Recorded clinician & 387 (46.5\%) & 287 (34.5\%) & 159 (19.1\%) \\
                  & GRASP, reward-missing & 59 (7.1\%) & 734 (88.1\%) & 40 (4.8\%) \\
\addlinespace
3 (months 13--18) & Recorded clinician & 370 (44.4\%) & 310 (37.2\%) & 153 (18.4\%) \\
                   & GRASP, reward-missing & 83 (10.0\%) & 741 (89.0\%) & 9 (1.1\%) \\
\addlinespace
4 (months 19--24) & Recorded clinician & 419 (50.3\%) & 267 (32.1\%) & 147 (17.6\%) \\
                   & GRASP, reward-missing & 139 (16.7\%) & 664 (79.7\%) & 30 (3.6\%) \\
\bottomrule
\end{tabular}
\end{table}

\section{Sub-Gaussian Rewards with Finite Mean Range}
\label{sec:finite_mean_range}

This section explains how our analysis extends beyond uniformly bounded rewards by allowing sub-Gaussian reward noise, while requiring the conditional mean reward to have a finite range.

\begin{assumption}[Sub-Gaussian reward noise] \label{assump:noise} For all $h\in[H]$, $x\in\Ssc$, and $a\in\Asc$, the reward noise \[ r_h(x,a)-g\!\left(\left\langle \phi_r(x,a),\theta_h^* \right\rangle\right) \] is sub-Gaussian with uniformly bounded sub-Gaussian norm. That is, there exists a constant $\sigma_r>0$ such that \[ \left\| r_h(x,a)-g\!\left(\left\langle \phi_r(x,a),\theta_h^* \right\rangle\right) \right\|_{\psi_2} \le \sigma_r \] for all $h\in[H]$, $x\in\Ssc$, and $a\in\Asc$, where $\|\cdot\|_{\psi_2}$ denotes the sub-Gaussian Orlicz norm. \end{assumption}

This extension replaces the bounded realized rewards commonly assumed in offline RL \citep[e.g.][]{jin2020pessimism,xie2021bellman} with uniformly sub-Gaussian conditional noise and a finite range for the mean reward on the relevant linear-predictor region. Bounded noise is sub-Gaussian, whereas sub-Gaussian noise need not be bounded. The finite mean range motivates the normalization of $g$ described below.

To formalize this extension, we take $g_{\max}$ as an arbitrary constant larger than
$\sup_{|u|\leq \sup_{h\in[H]}\|\theta_h^*\|_2}g(u)$ and $g_{\min}$ as an arbitrary constant smaller than $\inf_{|u|\leq \sup_{h\in[H]}\|\theta_h^*\|_2}g(u)$. Define the normalized mean function by
\[
g_{nrm}(u)
=
\frac{g(u)-g_{\min}}{g_{\max}-g_{\min}}.
\]
Write $D:=g_{\max}-g_{\min}$. For every policy $\pi$, the corresponding
normalized value and suboptimality satisfy
\begin{align*}
&r_{h,nrm}:=\frac{r_h-g_{\min}}{D},
\quad
V_{h,nrm}^{\pi}(x)
=\frac{V_h^\pi(x)-(H-h+1)g_{\min}}{D},
\\&
\operatorname{SubOpt}_{nrm}(\pi;x)
=\frac{\operatorname{SubOpt}(\pi;x)}{D}.
\end{align*}
The affine shift is common to all policies, so normalization does not change
the optimal policy or any greedy comparison.
The same update applies after replacing the reward component by \(g_{nrm}\). Specifically, for \(\Isc_h^r\subseteq\Isc_h^p\),
\begin{equation}\label{est:beta_reward_observation_finite_mean_range}
  \widehat \beta_{h,nrm}\coloneqq (\widehat \Lambda_{h} + \lambda \bI_{d_p})^{-1} \sum_{\tau\in\Isc_h^p} \phi_p(x_h^{\tau},a_h^{\tau})  \widehat V_{h+1,nrm}(x_{h+1}^{\tau} ) \,,
\end{equation}
For a continuation value \(V\), define the normalized empirical backup by
\[
(\widehat\B_{h,nrm}V)(x,a)
:=
g_{nrm}\{\phi_r(x,a)^\top\widetilde\theta_h\}
+\langle\phi_p(x,a),\widehat\beta_{h,nrm}(V)\rangle,
\]
where \(\widehat\beta_{h,nrm}(V)\) denotes the right-hand side of
\eqref{est:beta_reward_observation_finite_mean_range} with
\(\widehat V_{h+1,nrm}\) replaced by \(V\). We use the tilded notation
\(\widetilde\B_{h,nrm}\) in the complete-observation
specialization. The normalized uncertainty quantifier is
\begin{equation}\label{def:gamma_reward_observation_finite_mean_range}
    \widehat{\Gamma}_{h,nrm}
    =
    \frac{\widehat{\Gamma}_h}{g_{\max}-g_{\min}}
    =
    \frac{\widetilde \Gamma_{r,h}+\widehat \Gamma_{p,h}}{g_{\max}-g_{\min}}
    =
    \widetilde{\Gamma}_{r,h,nrm}+\widehat{\Gamma}_{p,h,nrm}.
\end{equation}

Rather than restating Algorithm~\ref{alg:grasp_reward_observation}, the
normalized procedure applies that algorithm with three substitutions:
\begin{enumerate}[(i)]
\item retain the reward estimator \(\widetilde\theta_h\) from
\eqref{est:theta}, but replace the reward component \(g\) in the Bellman
backup by \(g_{nrm}\);
\item replace the continuation update, empirical Bellman backup, and
uncertainty radius by their normalized counterparts above, including
\eqref{est:beta_reward_observation_finite_mean_range} and
\eqref{def:gamma_reward_observation_finite_mean_range}, with
\(\widehat V_{H+1,nrm}\equiv0\); and
\item apply the same clipped pessimistic \(Q\)-update and greedy policy update
on the normalized scale, denoting the output by
\(\widehat\pi_{nrm}=\{\widehat\pi_{h,nrm}\}_{h=1}^H\).
\end{enumerate}
Use \(N=0\) and \(\Isc_h^r=\Isc_h^p=[n]\) for complete reward observation,
and \(\Isc_h^r=[n]\), \(\Isc_h^p=[n+N]\) for trajectory-level partial
observation. Convert normalized suboptimality back to the original reward
scale by multiplying by \(D\).

The corresponding finite-sample guarantees have the same dependence on the reward and transition sample sizes as in the bounded-reward case.
\begin{theorem}[\texorpdfstring{GRASP under finite mean range}{GRASP under finite mean range}]
\label{thm:partial_reward_observation_finite_mean_range}
Under Assumptions \ref{assump:link}, \ref{assump:sigma} and \ref{assump:noise}, set $\lambda=1$, $\alpha_r=c_r \sqrt{d_r+\log(H/\xi)}$, $\alpha_p=c_p(g_{\max}-g_{\min})\left(d_p+d_r\right) H \sqrt{\zeta_+}$, where $\zeta_+=\log \left(2\left(d_r+d_p\right) H(n+N) / \xi\right)$, $c_r>0$ is sufficiently large as a function of the fixed link and reward-noise constants, $c_p>0$ is a sufficiently large numerical constant, and $\xi \in (0,1)$. Then $\widehat \Gamma_{h,nrm}$ in \eqref{def:gamma_reward_observation_finite_mean_range} is a $\xi$-uncertainty quantifier for $\widehat{\B}_{h,nrm}\widehat{V}_{h+1,nrm}$. For any $x \in \Ssc$ and \(n\) large enough, the resulting policy $\widehat \pi_{nrm} = \{ \widehat \pi_{h,nrm} \}_{h=1}^H$ satisfies
$$
\operatorname{SubOpt}(\widehat{\pi}_{nrm} ; x)
\leq
2(g_{\max}-g_{\min})\sum_{h=1}^H \mathbb{E}_{\pi^*}\left[
\widehat{\Gamma}_{h,nrm}(x_h, a_h) \mid x_1=x\right]
$$
with probability at least $1-\xi$.
\end{theorem}

\begin{corollary}
\label{cor:partial_reward_observation_finite_mean_range_rate}
Under the assumptions of Theorem \ref{thm:partial_reward_observation_finite_mean_range}, if there exists $\rho_p>0$ such that $\lambda_{\min}(\Lambda_h) \geq \rho_p$ for every $h\in[H]$, then, for $n$ large enough,
\begin{align*}
\operatorname{SubOpt}(\widehat{\pi}_{nrm} ; x)  &\leq O\left(\sqrt{\frac{\{d_r+\log (H / \xi)\} H^2}{n\rho}}\right) \\ &+ O\left(\sqrt{\frac{(g_{\max}-g_{\min})^2\left(d_p+d_r\right)^2 H^4 \log \left(2\left(d_r+d_p\right) H(n+N) / \xi\right)}{(n+N)\rho_p}}\right)
\end{align*}
with probability at least $1-\xi$. As in the bounded-reward case, the leading
transition-estimation rate is governed by the pooled sample size \(n+N\)
rather than the reward-observed sample size \(n\).
\end{corollary}

\begin{theorem}[\texorpdfstring{Complete reward-observation specialization under finite mean range}{Complete reward-observation specialization under finite mean range}]
\label{thm:complete_reward_observation_finite_mean_range}
Under Assumptions \ref{assump:link}, \ref{assump:sigma} and \ref{assump:noise}, apply the normalized procedure with \(\Isc_h^r=\Isc_h^p=[n]\). Set $\lambda=1$, $\alpha_r=c_r \sqrt{d_r+\log(H/\xi)}$, $\alpha_p=c_p(g_{\max}-g_{\min})\left(d_p+d_r\right) H \sqrt{\zeta}$, where $\zeta=\log \left(2\left(d_r+d_p\right) H n / \xi\right)$, $c_r>0$ is sufficiently large as a function of the fixed link and reward-noise constants, $c_p>0$ is a sufficiently large numerical constant, and $\xi \in (0,1)$ is the confidence parameter. Then the specialized radius $\widetilde \Gamma_{h,nrm}$ from \eqref{def:gamma_reward_observation_finite_mean_range} is a $\xi$-uncertainty quantifier for $\widetilde{\B}_{h,nrm}\widetilde{V}_{h+1,nrm}$. For any $x \in \Ssc$ and $n$ large enough, the resulting policy $\widetilde \pi_{nrm} = \{ \widetilde \pi_{h,nrm} \}_{h=1}^H$ satisfies
 $$
 \operatorname{SubOpt}\big(\widetilde \pi_{nrm} ;x \big) \leq 2(g_{\max}-g_{\min}) \sum_{h=1}^H\E_{\pi^{*}}\Bigl[ \widetilde \Gamma_{h,nrm}(x_h, a_h)\mid x_1=x\Bigr]
 $$
with probability at least $1-\xi$.
\end{theorem}
\begin{corollary}\label{cor:complete_reward_observation_finite_mean_range_rate}
Under the assumptions of Theorem~\ref{thm:complete_reward_observation_finite_mean_range}, if there exists $\rho_p>0$ such that $\lambda_{\min}(\Lambda_h)\geq\rho_p$ for every $h\in[H]$, then, for $n$ large enough,
\begin{align*}
  \operatorname{SubOpt}\big(\widetilde \pi_{nrm} ;x \big)  &\leq O\left( \sqrt{\frac{\{d_r+\log (H/\xi)\}H^2}{n\rho}}\right) \\&+O\left(  \sqrt{\frac{(g_{\max}-g_{\min})^2(d_p+d_r)^2H^4\log \left((d_p+d_r) H n / \xi\right)}{n\rho_p}}\right)
\end{align*}
with probability at least $1-\xi$.
\end{corollary}

Corollaries~\ref{cor:partial_reward_observation_finite_mean_range_rate} and~\ref{cor:complete_reward_observation_finite_mean_range_rate} show that the transition-estimation term scales with the reward range $g_{\max}-g_{\min}$ and may therefore dominate when that range is large. Under partial reward observation, this term is governed by the pooled transition sample size $n+N$, so transition-only observations continue to reduce transition uncertainty in the finite-mean-range setting.

\section{Proofs}
\label{sec:proofs_appendix}

\paragraph*{Proof roadmap.}
After Bellman completeness, we establish the uniform transition calculation
used by the offline results. We then prove
Theorem~\ref{thm:partial_reward_observation_suboptimality} with separate
reward and transition index sets.
Corollary~\ref{cor:complete_reward_observation_rate} follows by
setting \(N=0\). The step-level extension is collected in
Section~\ref{sec:step_level_reward_observation}.

\paragraph*{Additional notation.} For clarity about constants, let
$R_\theta=1+\max_{h\in[H]}\|\theta_h^*\|_2$ and
$\underline g_{R_\theta}=\inf_{|u|\le R_\theta}\dot g(u)>0$.
Reward-confidence constants below may depend on the fixed link quantities
$(L,\underline g_{R_\theta})$ and the uniform conditional sub-Gaussian norm of
the centered reward noise, and sample-size thresholds may additionally
depend on fixed model quantities such as $(R_\theta,\rho)$. They do not depend
on $n,N,H,d_r,d_p$, or the confidence level.
\subsection{Proof of Proposition~\ref{prop:bellman_completeness}}
\label{sec:proof_bellman}

\begin{proof}
Fix any \(h\in[H]\) and any bounded measurable function
\(V:\Ssc\to\mathbb{R}\). By the GRASP-MDP transition model, we have
\begin{align*}
(\mathbb{B}_h V)(x,a)
&=
g\!\left(\left\langle
\phi_r(x,a),\theta_h^*
\right\rangle\right)
+
\int_{\Ssc}V(x')\,\P_h(dx'\mid x,a)\\
&=
g\!\left(\left\langle
\phi_r(x,a),\theta_h^*
\right\rangle\right)
+
\int_{\Ssc}
V(x')
\left\langle
\phi_p(x,a),\mu_h(dx')
\right\rangle\\
&=
g\!\left(\left\langle
\phi_r(x,a),\theta_h^*
\right\rangle\right)
+
\left\langle
\phi_p(x,a),
\int_{\Ssc}V(x')\,\mu_h(dx')
\right\rangle.
\end{align*}
Define
\[
\beta_h(V)
:=
\int_{\Ssc}V(x')\,\mu_h(dx')
\in\mathbb{R}^{d_p}.
\]
It follows that
\[
(\mathbb{B}_h V)(x,a)
=
f_{\theta_h^*,\beta_h(V)}(x,a).
\]
We conclude that
\[
\mathbb{B}_hV\in\mathcal F.
\]

In particular, the Bellman equations give
\[
Q_h^\pi=\mathbb{B}_hV_{h+1}^\pi
\qquad\text{and}\qquad
Q_h^*=\mathbb{B}_hV_{h+1}^*.
\]
Under the standing boundedness and measurability assumptions,
both \(V_{h+1}^\pi\) and \(V_{h+1}^*\) are bounded measurable.
Hence,
\[
Q_h^\pi\in\mathcal F
\qquad\text{and}\qquad
Q_h^*\in\mathcal F
\]
for every \(h\in[H]\). Specifically,
\[Q_h^*(x,a)
=
(\mathbb{B}_hV_{h+1}^*)(x,a)
=
g\!\left(\left\langle\phi_r(x,a),\theta_h^*\right\rangle\right)
+
\left\langle\phi_p(x,a),\beta_h^*\right\rangle,
\]
where
\[
\beta_h^*
:=
\int_{\Ssc}V_{h+1}^*(x')\,\mu_h(dx').
\]
\end{proof}

\subsection{Uniform Transition Calculation}
\label{sec:shared_bellman_uncertainty}
We give the calculation in complete-observation notation; the proof of
Theorem~\ref{thm:partial_reward_observation_suboptimality} replaces the \(n\)
transition samples below by \(n+N\). Let \(\E_{\Dsc}\) denote expectation over the sampling distribution of \(\Dsc\). For
\[
\beta_h=\int_{\Ssc}\widetilde V_{h+1}(x')\,\mu_h(\mathrm{d}x'),
\qquad
\Omega_h=(\widetilde\Lambda_h+\lambda\bI_{d_p})^{-1},
\]
write the transition-estimation error as
\((\mathrm{ii})=\langle\phi_p(x,a),\beta_h-\widetilde\beta_h\rangle\).
By the definition of \(\widetilde\beta_h\),
$$
\begin{aligned}
(\mathrm{ii}) = &  \phi_p(x, a)^\top \beta_{h} -  \phi_p(x, a)^\top  \Omega_{h}  \sum_{\tau=1}^{n} \phi_p(x_h^{\tau},a_h^{\tau})  \widetilde V_{h+1}(x_{h+1}^{\tau} ) \\
= &  \underbrace{ \phi_p(x, a)^\top \beta_{h} - \phi_p(x, a)^\top \Omega_{h} \sum_{\tau=1}^{n}  \phi_p(x_h^{\tau},a_h^{\tau}) \phi_p(x_h^{\tau},a_h^{\tau})^\top \beta_{h}}_{(\mathrm{iii})} \\
& - \underbrace{ \phi_p(x, a)^\top  \Omega_{h} \sum_{\tau=1}^{n}\phi_p(x_h^{\tau},a_h^{\tau}) \big(  \widetilde V_{h+1}(x_{h+1}^{\tau}) - \phi_p(x_h^{\tau},a_h^{\tau})^\top \beta_{h} \big)}_{(\mathrm{iv})}.
\end{aligned}
$$
Then by Lemma \ref{lem:norm}, we have 
$\|\beta_{h}\|_2 \leq H \sqrt{d_p}$, and 
$$
\begin{aligned}
|(\mathrm{iii}) | = & \big | \phi_p(x, a)^\top \beta_{h} - \phi_p(x, a)^\top  ( \widetilde \Lambda_{h} + \lambda \bI_{d_p})^{-1}  \widetilde \Lambda_{h} \beta_{h} \big | = \big | \lambda \phi_p(x, a)^\top  ( \widetilde  \Lambda_{h} + \lambda \bI_{d_p})^{-1} \beta_{h}	\big| \\
\leq & H \sqrt{\lambda  d_p} \sqrt{\phi_p(x, a)^\top  ( \widetilde \Lambda_{h} + \lambda \bI_{d_p})^{-1} \phi_p(x, a)}  \,.
\end{aligned}
$$
We then bound $|(\mathrm{iv})|$. To simplify the notation, for $h \in[H]$ and $\tau \in[n]$, and any value function $V: \mathcal{S} \rightarrow[0, H]$, we define
\begin{equation*}
   \epsilon_{h}^{\tau}(V) = V(x_{h+1}^{\tau})-\E \big[V (x_{h+1}) \mid x_h = x_{h}^{\tau}, a_h = a_{h}^{\tau} \big]\,.
 \end{equation*}
We then have
$$
\begin{aligned}
|(\mathrm{iv})| ={}& \left|\phi_p(x, a)^\top  \Omega_{h} \sum_{\tau=1}^{n}  \phi_p(x_h^{\tau},a_h^{\tau})  \epsilon_{h}^{\tau} (  \widetilde V_{h+1})\right|\\
& \leq  \| \phi_p(x, a)\|_{\Omega_{h}}  \underbrace{ \Big \| \sum_{\tau=1}^{n}  \phi_p(x_h^{\tau},a_h^{\tau})  \epsilon_{h}^{\tau}  (  \widetilde V_{h+1}) \Big \|_{\Omega_{h}}}_{(\mathrm{v})}.
\end{aligned}
$$
We then bound term $(\mathrm{v})$ via concentration inequalities. An obstacle is that $\widetilde{V}_{h+1}$ depends on $\big\{(x_h^{\tau}, a_h^{\tau})\big\}_{\tau=1}^{n}$ via $\big\{(x_{h^{\prime}}^{\tau}, a_{h^{\prime}}^{\tau})\big\}_{\tau \in[n], h^{\prime}>h}$, as it is constructed based on the dataset $\Dsc$. To this end, we resort to uniform concentration inequalities. Specifically, for all $h \in[H]$, we define the function class
$$
\begin{aligned}
& \mathcal{V}_h(R, B, J_r, J_p, \rho, \lambda)=  \Big\{V_h(x ; \theta, \beta, \Sigma, \Lambda,\gamma_r,\gamma_p): \mathcal{S} \rightarrow[0, H] \text { with } \\
 & \quad \quad  \quad \quad \quad \quad \quad \quad \quad  \|\theta\|_2 \leq R, \|\beta\|_2 \leq B, \gamma_r \in[0, J_r], \gamma_p \in[0, J_p], \Sigma \succeq \rho \bI_{d_r}, \Lambda \succeq \lambda \bI_{d_p} \Big\} \,, \\
& \text { where } V_h(x ; \theta, \beta, \Sigma, \Lambda,\gamma_r,\gamma_p ) = \max_{a \in \Asc}\Big\{\min \big\{ f_r(x,a;\theta,\Sigma,\gamma_r) +  f_p(x,a;\beta,\Lambda,\gamma_p)   , H-h+1\big\}^{+}\Big\} \\
& \text{ with } f_r(x,a;\theta,\Sigma,\gamma_r) =  g( \langle \phi_r(x,a),\theta \rangle) - \gamma_r\dot g(\langle \phi_r(x,a),\theta \rangle)\sqrt{\phi_r(x, a)^\top\Sigma^{-1} \phi_r(x, a)} \\
& \text{ and } f_p(x,a;\beta,\Lambda,\gamma_p) =  \langle\phi_p(x, a), \beta\rangle - \gamma_p \cdot \sqrt{\phi_p(x, a)^\top\Lambda^{-1} \phi_p(x, a)}\,.
\end{aligned}
$$
For all $\epsilon >0$, let $\mathcal{N}_{h}(\epsilon; R, B, J_r, J_p,\rho, \lambda)$ be the minimal $\epsilon$-cover of $\mathcal{V}_{h}(R, B, J_r, J_p,\rho, \lambda)$ with respect to the supremum norm. In other words, for any function $V \in \mathcal{V}_{h}(R, B, J_r,J_p,\rho, \lambda)$, there exists a function $V^{\dagger} \in \mathcal{N}_{h}(\epsilon ;R, B, J_r, J_p,\rho, \lambda)$ such that
$$
\sup _{x \in \mathcal{S}}\big|V(x)-V^{\dagger}(x)\big| \leq \epsilon \,.
$$
Meanwhile, among all $\epsilon$-covers of $\mathcal{V}_{h}(R, B, J_r, J_p,\rho, \lambda)$ defined by such a property, we choose $\mathcal{N}_{h}(\epsilon; R, B, J_r,J_p,\rho, \lambda)$ as the one with the minimal cardinality.

Recall the construction of $\widetilde V_h$ in Algorithm \ref{alg:grasp_reward_observation}. Set
\[
R_0:=1+\max_{h\in[H]}\|\theta_h^*\|_2.
\]
For sufficiently large $n$, Lemma~\ref{lem:init} gives $\|\widetilde\theta_h\|_2\le R_0$ simultaneously over $h$ with probability at least $1-\xi/8$. By \eqref{eq:sigma}, invoked at the same split confidence level, we have $\lambda_{\min} \big(\frac{1}{n}\widetilde \Sigma_{h}(\widetilde \theta_h) \big) \geq \rho/4$ with probability at least $1 - \xi/8$, where $\rho = \min_{h\in[H]}\lambda_{\min}\big(\Sigma_{h}(\theta_h^{*}) \big) >0$. By Lemma \ref{lem:norm}, we have $\| \widetilde \beta_h\|_2 \leq H \sqrt{n/\lambda} \coloneqq B_0$. Finally, we take $J_r = 2 \alpha_{r}$ and $J_p = 2 \alpha_{p}$. Under these events, the preceding class matches the algorithmic reward penalty and we have
$$
\widetilde{V}_{h+1} \in \mathcal{V}_{h+1}(R_0, B_0, J_r, J_p,  n\rho/4, \lambda).
$$
Here $\lambda>0$ is the regularization parameter and $\alpha_{r}, \alpha_{p} >0$ are the scaling parameters, which are specified
in the corresponding theorem. For notational simplicity, we use $\mathcal{V}_{h+1}$ and $\mathcal{N}_{h+1}(\epsilon)$ to denote $ \mathcal{V}_{h+1}(R_0, B_0, J_r, J_p, n\rho/4, \lambda)$ and $\mathcal{N}_{h+1}(\epsilon ; R_{0}, B_{0}, J_r, J_p, n\rho/4, \lambda)$, respectively. As a result, there exists a function $V_{h+1}^{\dagger} \in \mathcal{N}_{h+1}(\epsilon)$ such that
$$
\sup _{x \in \mathcal{S}} |\widetilde{V}_{h+1}(x)-V_{h+1}^{ \dagger}(x)| \leq \epsilon.
$$
Hence, given $\widetilde{V}_{h+1}$ and $V_{h+1}^{ \dagger}$, we have 
\begin{equation*}
\begin{aligned}
\E\big[ |V_{h+1}^{\dagger}(x_{h+1}) - \widetilde{V}_{h+1} (x_{h+1})| \mid  x_{h}=x, a_{h}=a \big]  \leq \epsilon, \quad \forall(x, a) \in \mathcal{S} \times \Asc, \forall h \in[H]   \,.
\end{aligned}
\end{equation*}
Here the conditional expectation is induced by the transition kernel $\P_{h}(\cdot \mid x, a)$. As a result, for all $h \in[H]$, we have
$$
\big|\epsilon_{h}^{\tau}(\widetilde{V}_{h+1})-\epsilon_{h}^{\tau}(V_{h+1}^{\dagger})\big| \leq 2 \epsilon,  \forall \tau \in[n] \,.
$$

By the Cauchy-Schwarz inequality, for any two vectors $a, b \in \mathbb{R}^{d}$ and any positive definite matrix $\Lambda \in \mathbb{R}^{d \times d}$, it holds that $\|a+b\|_{\Lambda}^{2} \leq$ $2 \|a\|_{\Lambda}^{2}+2 \|b\|_{\Lambda}^{2} .$ Hence, for all $h \in[H]$, we have
\begin{equation*}
\begin{aligned}
     &  |(\mathrm{v})|^{2} \leq 2 \underbrace{ \Big \| \sum_{\tau=1}^{n}  \phi_p(x_h^{\tau},a_h^{\tau})  \epsilon_{h}^{\tau}  (   V^{\dagger}_{h+1}) \Big \|_{\Omega_{h}}^2	}_{ b^2( \{ V_{h+1}^{\dagger}\}) }
    +  \underbrace{2 \Big \| \sum_{\tau=1}^{n}  \phi_p(x_h^{\tau},a_h^{\tau}) \big( \epsilon_{h}^{\tau}  (   \widetilde V_{h+1}) -  \epsilon_{h}^{\tau}  (   V^{\dagger}_{h+1}) \big) \Big \|_{\Omega_{h}}^2	}_{(\mathrm{vi})}
\end{aligned}
\label{eq:iii}   
\end{equation*}
To bound $(\mathrm{vi})$, let \(X_h\) be the matrix with rows
\(\phi_p(x_h^\tau,a_h^\tau)^\top\), and let \(d_h\) collect the \(n\)
differences
\(\epsilon_h^\tau(\widetilde V_{h+1})
-\epsilon_h^\tau(V_{h+1}^\dagger)\). Since
\[
0\preceq
X_h(X_h^\top X_h+\lambda\bI_{d_p})^{-1}X_h^\top
\preceq \bI_n,
\]
we obtain
\begin{equation}
(\mathrm{vi})
=2d_h^\top X_h\Omega_hX_h^\top d_h
\le 2\|d_h\|_2^2
\le 8n\epsilon^2,
\label{eq:iii_2}
\end{equation}
We then bound $b( \{ V_{h+1}^{\dagger}\})$ via uniform concentration inequalities. Applying Lemma \ref{lem:self_normalized_value_concentration} and the union bound, for any fixed $h \in [H]$, we have 
$$
\begin{aligned}
 & \P_{\Dsc} \bigg(  \sup _{V \in \mathcal{N}_{h+1}(\epsilon)} \Big \| \sum_{\tau=1}^{n}  \phi_p(x_h^{\tau},a_h^{\tau})  \epsilon_{h}^{\tau}  (V) \Big \|_{\Omega_{h}}^2 > H^{2} (2 \log (1 / \delta)+d_p \log (1+ n/ \lambda)) \bigg)\\
 & \leq   \delta  | \mathcal{N}_{h+1}(\epsilon ) |.
\end{aligned}
$$
For all $\xi \in(0, 1)$ and all $\epsilon > 0$, we set  $\delta =\xi/( 4 H |\mathcal{N}_{h+1}(\epsilon)| )$.  
Hence, for any fixed $h \in [H]$, it holds that
\begin{equation}
\begin{aligned}
& \sup _{V \in \mathcal{N}_{h+1}(\epsilon)} \Big \| \sum_{\tau=1}^{n}  \phi_p(x_h^{\tau},a_h^{\tau})  \epsilon_{h}^{\tau}  (V) \Big \|_{\Omega_{h}}^2 \leq H^{2} (2 \log (1 / \delta)+d_p \log (1+ n/ \lambda))  \\
& \leq H^2  \Bigl ( 2 \log \bigl ( 4 H   | \mathcal{N}_{h+1} (\epsilon ) | / \xi  \bigr )   + d_p  \log (1 + n / \lambda )  \Bigr )
\end{aligned}
   \label{eq:bound_term3_4}
\end{equation}
with probability at least $1 - \xi /(4 H)$, which is taken with respect to $\P_{\Dsc}$. Using the union bound again, \eqref{eq:bound_term3_4} holds for all $h \in [H]$ with probability at least $1 - \xi/4$. Combining \eqref{eq:iii_2} and \eqref{eq:bound_term3_4}, we have 
$$ |(\mathrm{v})|^{2} \leq  2H^2  \Bigl ( 2 \log \bigl ( 4 H   | \mathcal{N}_{h+1} (\epsilon ) | / \xi  \bigr )   + d_p  \log (1 + n / \lambda )  \Bigr )      + 8 n \epsilon^2.$$
Applying Lemma \ref{lem:cover0} with
$R_0 = 1+\max_{h\in[H]}\|\theta_h^* \|_2$, $B_0 = H \sqrt{n/\lambda}$, $\epsilon = H \sqrt{d_p }/\sqrt n$, $J_r=2c_r\sqrt{d_r+\log(H/\xi)}$, $J_p=2c_p(d_p+d_r)H\sqrt\zeta$, $\zeta =\log \big(2 (d_r+d_p) H n / \xi\big)$,
and $\lambda = 1$. If $n$ is sufficiently large and $c_p$ is chosen
sufficiently large relative to $c_r$, then
$$
\begin{aligned}
 & \log \big|\mathcal{N}_{h}(\epsilon ; R_0, B_0, J_r, J_p, n\rho/4, \lambda)\big| \\
&\leq d_r\log\!\left(1+
\frac{8L\{1+J_r\sqrt{4/(n\rho)}\}R_0}{\epsilon}\right)
+d_p\log(1+8B_0/\epsilon)\\
&\quad+d_r^2\log\!\left(1+
\frac{128L^2d_r^{1/2}J_r^2}{n\rho\epsilon^2}\right)
+d_p^2\log\!\left(1+
\frac{32d_p^{1/2}J_p^2}{\lambda\epsilon^2}\right)\\
&\leq C(d_r+d_p)^2\{\log(c_p+1)+\zeta\}.
\end{aligned}
$$
Here $C>0$ is universal once the fixed model constants $(L,\rho,R_0)$ are specified, and the last inequality holds above a sample-size threshold that may depend on those constants. The displayed entropy bound implies that
$$
\begin{aligned}
  |(\mathrm{v})|^{2} &  \leq 2H^2  \Bigl (2  \log \bigl ( 4 H   | \mathcal{N}_{h+1} (\epsilon ) | / \xi  \bigr )   + d_p  \log (1 + n / \lambda )  \Bigr )  + 8  n \epsilon^2 \\
 & \leq C'(d_r+d_p)^2H^2\zeta\{1+\log(c_p+1)\}\\
 & \leq (d_r + d_p)^2H^2c_p^2\zeta/4
\end{aligned}
$$
when $ c_p \geq 1$ is sufficiently large. We then have $|(\mathrm{iv})| \leq \alpha_{p}/2  \| \phi_p(x, a)\|_{\Omega_{h}}$. Since $ H \sqrt{ \lambda d_p}  \leq  \alpha_{p}/2$, combining the bounds for $(\mathrm{iii})$ and $(\mathrm{iv})$ gives
$$
|(\mathrm{ii})| = \big| \langle\phi_p(x, a), \beta_{h} - \widetilde \beta_{h} \rangle \big|   \leq \alpha_{p}  \| \phi_p(x, a) \|_{\Omega_{h}} =  \widetilde \Gamma_{p,h}(x, a).
$$
The parameter, reward-curvature, and uniform transition events used above
have total failure probability at most \(\xi/2\). Consequently,
\[
\left|\left\langle
\phi_p(x,a),\beta_h-\widetilde\beta_h
\right\rangle\right|
\le \widetilde\Gamma_{p,h}(x,a)
\]
simultaneously over \(h,x,a\) with probability at least \(1-\xi/2\).

\subsection{Proof of Theorems~\ref{thm:partial_reward_observation_suboptimality} and~\ref{thm:partial_reward_observation_finite_mean_range}}
\label{sec:proof_thm_partial_reward_observation}

We apply Section~\ref{sec:shared_bellman_uncertainty} using
\(\Isc_h^r=[n]\) for reward learning and
\(\Isc_h^p=[n+N]\) for transition learning. Once the resulting total
Bellman-uncertainty event is established, the pessimism argument is
deterministic.

\begin{proof}
We give the bounded-reward argument. The finite-mean-range result is identical
after applying the affine normalization in
Supplementary~\ref{sec:finite_mean_range}.

For \(\xi\in(0,1)\), let
\[
\mathcal C_h(z):=\max\{\min\{z,H-h+1\},0\}.
\]
The algorithm can be written as
\[
\widehat Q_h(x,a)
=
\mathcal C_h\!\left(
(\widehat\B_h\widehat V_{h+1})(x,a)
-
\widehat\Gamma_h(x,a)
\right).
\]
Consider the joint event on which, simultaneously for every \(h,x,a\),
\[
\left|
(\widehat\B_h\widehat V_{h+1})(x,a)
-
(\B_h\widehat V_{h+1})(x,a)
\right|
\le
\widehat\Gamma_h(x,a).
\tag{E}
\label{eq:partial-uq-event}
\]
The reward part of this event follows from Lemma~\ref{lem:g}, invoked with
failure probability $\xi/2$. For the transition part, apply
Section~\ref{sec:shared_bellman_uncertainty} with confidence parameter \(\xi\)
and \(m=n+N\) transition samples, using
\[
\widehat\Omega_h=(\widehat\Lambda_h+\lambda\bI_{d_p})^{-1},
\qquad
B_0=H\sqrt{m/\lambda},
\qquad
\epsilon=H\sqrt{d_p/m},
\]
and retain the reward-design lower bound
\(\widetilde\Sigma_h(\widetilde\theta_h)\succeq n\rho\bI_{d_r}/2\).
Lemma~\ref{lem:cover0} gives covering entropy
\(O\{(d_r+d_p)^2\zeta_+\}\); the additional term
\(\log\{1+m/(n\rho)\}\) is absorbed by \(\zeta_+\) above the stated
sample-size threshold. The finite-net union bound in
\eqref{eq:bound_term3_4} therefore gives
\[
\left|
\left\langle
\phi_p(x,a),
\beta_h(\widehat V_{h+1})-\widehat\beta_h
\right\rangle
\right|
\le
\widehat\Gamma_{p,h}(x,a)
\]
simultaneously over \(h,x,a\). Combining the reward and transition bounds proves
\eqref{eq:partial-uq-event} with probability at least \(1-\xi\).

On this event, abbreviate
\[
b_h(x,a):=(\B_h\widehat V_{h+1})(x,a),
\qquad
\widehat b_h(x,a):=(\widehat\B_h\widehat V_{h+1})(x,a).
\]
Because the relevant conditional reward means (normalized in the
finite-mean-range case) and continuation values are bounded,
\(b_h(x,a)\in[0,H-h+1]\), and hence \(\mathcal C_h(b_h)=b_h\).
Event~\eqref{eq:partial-uq-event} gives
\[
\widehat b_h-\widehat\Gamma_h
\le b_h
\quad\text{and}\quad
\widehat b_h-\widehat\Gamma_h
\ge b_h-2\widehat\Gamma_h.
\]
Since \(\mathcal C_h\) is monotone and one-Lipschitz,
\[
0
\le
\iota_h(x,a)
:=
(\B_h\widehat V_{h+1})(x,a)-\widehat Q_h(x,a)
\le
2\widehat\Gamma_h(x,a).
\tag{P}
\label{eq:partial-pessimism-residual}
\]

The finite-horizon performance-difference identity used by
PEVI gives
\[
\begin{aligned}
\operatorname{SubOpt}(\widehat\pi;x)
&=
-\sum_{h=1}^H
\E_{\widehat\pi}[\iota_h(x_h,a_h)\mid x_1=x]
+
\sum_{h=1}^H
\E_{\pi^*}[\iota_h(x_h,a_h)\mid x_1=x]\\
&\quad+
\sum_{h=1}^H
\E_{\pi^*}\!\left[
\left\langle
\widehat Q_h(x_h,\cdot),
\pi_h^*(\cdot\mid x_h)-\widehat\pi_h(\cdot\mid x_h)
\right\rangle_{\Asc}
\middle|x_1=x
\right].
\end{aligned}
\]
The first term is nonpositive by \eqref{eq:partial-pessimism-residual}, and the
last term is nonpositive because \(\widehat\pi_h\) is greedy with respect to
\(\widehat Q_h\). Applying the upper bound in
\eqref{eq:partial-pessimism-residual} to the remaining term yields
\[
\operatorname{SubOpt}(\widehat\pi;x)
\le
2\sum_{h=1}^H
\E_{\pi^*}\!\left[
\widehat\Gamma_h(x_h,a_h)\mid x_1=x
\right].
\]
The same argument with the normalized Bellman operator gives
\(\operatorname{SubOpt}_{nrm}\le2\sum_h\E_{\pi^*}\widehat\Gamma_{h,nrm}\).
Multiplying by $g_{\max}-g_{\min}$ proves
Theorem~\ref{thm:partial_reward_observation_finite_mean_range} in original reward units.
\end{proof}

\subsection{Complete Reward-Observation Specializations}
\label{sec:proof_cor_complete_reward_observation_specialization}
\begin{proof}
Set \(N=0\) in Theorem~\ref{thm:partial_reward_observation_suboptimality}.
Then \(\Isc_h^r=\Isc_h^p=[n]\), the hatted quantities become their tilded
counterparts, and \(\zeta_+=\zeta\). This gives
Corollary~\ref{cor:complete_reward_observation_rate}. Applying the
same specialization after affine normalization gives
Theorem~\ref{thm:complete_reward_observation_finite_mean_range}.

For the rate statements, invoke the corresponding uncertainty-quantifier
result at confidence level \(\xi/2\). Lemma~\ref{lem:concen} and
\(\lambda_{\min}(\Lambda_h)\ge\rho_p\) give, simultaneously over \(h\),
\[
\lambda_{\min}(n^{-1}\widetilde\Lambda_h)\ge\rho_p/2
\]
with failure probability at most \(\xi/4\). Lemma~\ref{lem:init} and
\eqref{eq:sigma}, with their total failure probability set to \(\xi/4\),
give both
\[
\lambda_{\min}\!\left(
n^{-1}\widetilde\Sigma_h(\widetilde\theta_h)
\right)\ge\rho/4
\]
and the stated bound on
\(\max_h\|\widetilde\theta_h-\theta_h^*\|_2\). On the intersection of these
events, substituting the two covariance lower bounds into the reward and
transition radii proves the rate in
Corollary~\ref{cor:complete_reward_observation_rate}. For the normalized
finite-mean-range result, conversion to original units cancels the factor
\(D^{-1}\) in the reward radius and contributes one factor \(D\) to the
transition term, proving
Corollary~\ref{cor:complete_reward_observation_finite_mean_range_rate}.
\end{proof}

\subsection{Proof of Corollaries~\ref{cor:partial_reward_observation_rate} and~\ref{cor:partial_reward_observation_finite_mean_range_rate}}
\label{sec:proof_cor_partial_reward_observation}
\begin{proof}
Apply Theorem~\ref{thm:partial_reward_observation_suboptimality} (or its
finite-mean-range counterpart) at confidence level $\xi/2$, and allocate the
remaining failure probability $\xi/2$ to the empirical population-covariance events
used below. Constant factors inside the logarithms are absorbed by the tuning
constants.
For the pooled-transition estimator, Lemma~\ref{lem:concen}, with \(n\)
replaced by \(n+N\), and the assumption
\(\lambda_{\min}(\Lambda_h)\ge\rho_p\) imply, simultaneously over \(h\),
\[
\widehat\Gamma_{p,h}(x,a)
=
O\left(
\sqrt{
\frac{(d_p+d_r)^2H^2
\log\{2(d_r+d_p)H(n+N)/\xi\}}
{(n+N)\rho_p}
}
\right).
\]
Lemma~\ref{lem:g} and the reward-curvature event give
\[
\widetilde\Gamma_{r,h}(x,a)
=
O\left(
\sqrt{\frac{d_r+\log(H/\xi)}{n\rho}}
\right).
\]
Substituting
\(\widehat\Gamma_h=\widetilde\Gamma_{r,h}+\widehat\Gamma_{p,h}\)
into Theorem~\ref{thm:partial_reward_observation_suboptimality} and summing
over \(h\) proves the two-term bound in
Corollary~\ref{cor:partial_reward_observation_rate}. Repeating the same
calculation for the normalized operator and then using
\(\operatorname{SubOpt}=(g_{\max}-g_{\min})\operatorname{SubOpt}_{nrm}\)
proves Corollary~\ref{cor:partial_reward_observation_finite_mean_range_rate}.
The conversion cancels the normalization in the reward radius and contributes
one factor \(g_{\max}-g_{\min}\) to the transition term.
For both the bounded-reward and normalized finite-mean-range recursions,
\(\alpha_r\asymp\sqrt{d_r+\log(H/\xi)}\). After converting the normalized
result to the original reward scale, the reward contribution is
\(O\{\sqrt{(d_r+\log(H/\xi))/(n\rho)}\}\).
\end{proof}
The remaining supplement sections collect extensions that use the same reward-transition decomposition but add modeling ingredients, including M-estimated rewards, step-level reward observation, beta regression, relaxed curvature, and online optimism.

\section{Generalization Results}
\label{sec:generalization_results}
We collect the M-estimation extension, step-level reward observation, a concrete beta-regression verification, the reward-curvature relaxation, and the online regret proof as generalization results.

\subsection{M-estimation Reward Models}
\label{sec:m_estimator_extension}
This subsection gives a general theorem for replacing the canonical GLM reward estimator by an M-estimator satisfying abstract reward-confidence and transition-concentration conditions.
The main text uses canonical GLM losses so that the reward-side confidence
radius has an explicit closed form. The pessimistic value-iteration argument
only needs two ingredients: a uniformly valid confidence band for the
conditional reward mean and transition concentration for the value class
generated by that reward estimator. We formalize this high-level template
below.

For each step $h$, let
\(\Theta_h\) be the reward-parameter space and let
$m_h:\Ssc\times\Asc\times\Theta_h\to[0,1]$ be a parametric reward mean model,
and assume that the true conditional reward mean can be written as
\[
    \E(r_h\mid x_h=x,a_h=a)=m_h(x,a;\theta_h^\dagger)
\]
for some $\theta_h^\dagger\in\Theta_h$. Let $\widetilde\theta_h^M$ denote an
M-estimator based only on the observed rewards at step $h$, for example a
solution of an estimating equation or a minimizer of a convex empirical loss.
For a value function $V$, write
\[
    \beta_h(V)=\int_{\Ssc} V(x')\,\mu_h({\rm d}x')
\]
so that the transition component remains
$\E\{V(x_{h+1})\mid x_h=x,a_h=a\}=\langle \phi_p(x,a),\beta_h(V)\rangle$.

\begin{assumption}[M-estimator reward confidence]
\label{assump:m_estimator_reward}
For any confidence level \(\xi_r\in(0,1)\), the reward estimator
\(\widetilde\theta_h^M\) and a nonnegative radius
\(\widetilde\Gamma_{M,r,h}(x,a)\) can be tuned so that, with probability at
least \(1-\xi_r\),
\[
\left|
m_h(x,a;\theta_h^\dagger)
-
m_h(x,a;\widetilde\theta_h^M)
\right|
\le
\widetilde\Gamma_{M,r,h}(x,a)
\]
for all \((x,a)\in\Ssc\times\Asc\) and \(h\in[H]\).
\end{assumption}

For the two recursions below, the complete-observation version uses the \(n\)
reward-observed trajectories for both components, whereas the
trajectory-level partial-observation version uses the same \(n\) trajectories
for reward estimation and all \(n+N\) trajectories for transition estimation.
Define GRASP under complete reward observation with an M-estimated reward model by replacing
$g(\langle\phi_r(x,a),\widetilde\theta_h\rangle)$ with
$m_h(x,a;\widetilde\theta_h^M)$ and replacing the GLM reward radius with
$\widetilde\Gamma_{M,r,h}$. Thus, with $\widetilde V_{H+1}^M\equiv0$,
\[
\begin{aligned}
    \widetilde\beta_h^M
    &=
    (\widetilde\Lambda_h+\lambda\bI_{d_p})^{-1}
    \sum_{\tau=1}^n
    \phi_p(x_h^\tau,a_h^\tau)
    \widetilde V_{h+1}^M(x_{h+1}^\tau),\\
    \widetilde\Gamma_{M,h}(x,a)
    &=
    \widetilde\Gamma_{M,r,h}(x,a)
    +
    \widetilde\Gamma_{p,h}(x,a),\\
    \widetilde Q_h^M(x,a)
    &=
    \min\left\{
        m_h(x,a;\widetilde\theta_h^M)
        +
        \phi_p(x,a)^\top\widetilde\beta_h^M
        -
        \widetilde\Gamma_{M,h}(x,a),
        H-h+1
    \right\}^+,
\end{aligned}
\]
and $\widetilde\pi_h^M$ is greedy with respect to
$\widetilde Q_h^M$, with
$\widetilde V_h^M(x)=\langle\widetilde Q_h^M(x,\cdot),
\widetilde\pi_h^M(\cdot\mid x)\rangle_{\Asc}$.
For the pooled-transition version, set
$\widehat V_{H+1}^M\equiv0$ and, backward for $h=H,\ldots,1$, define
\[
\begin{aligned}
    \widehat\beta_h^M
    & =
    (\widehat\Lambda_h+\lambda\bI_{d_p})^{-1}
    \sum_{\tau=1}^{n+N}
    \phi_p(x_h^\tau,a_h^\tau)
    \widehat V_{h+1}^M(x_{h+1}^\tau),\\
    \widehat\Gamma_{M,h}(x,a)
    & =
    \widetilde\Gamma_{M,r,h}(x,a)+\widehat\Gamma_{p,h}(x,a),\\
    \widehat Q_h^M(x,a)
    & =
    \min\left\{
        m_h(x,a;\widetilde\theta_h^M)
        +\phi_p(x,a)^\top\widehat\beta_h^M
        -\widehat\Gamma_{M,h}(x,a),
        H-h+1
    \right\}^+,\\
    \widehat\pi_h^M(\cdot\mid x)
    &\in
    \arg\max_{\nu\in\Delta(\Asc)}
    \langle\widehat Q_h^M(x,\cdot),\nu\rangle_{\Asc},\\
    \widehat V_h^M(x)
    &=\langle\widehat Q_h^M(x,\cdot),
    \widehat\pi_h^M(\cdot\mid x)\rangle_{\Asc}.
\end{aligned}
\]

\begin{assumption}[Transition concentration for the M-based value class]
\label{assump:m_estimator_transition}
For any confidence level $\xi_p\in(0,1)$, the transition radii
$\widetilde\Gamma_{p,h}$ and $\widehat\Gamma_{p,h}$ can be tuned so that, with
probability at least $1-\xi_p$,
\[
\begin{aligned}
    \left|
        \left\langle
            \phi_p(x,a),
            \beta_h(\widetilde V_{h+1}^M)-\widetilde\beta_h^M
        \right\rangle
    \right|
    &\le
    \widetilde\Gamma_{p,h}(x,a),\\
    \left|
        \left\langle
            \phi_p(x,a),
            \beta_h(\widehat V_{h+1}^M)-\widehat\beta_h^M
        \right\rangle
    \right|
    &\le
    \widehat\Gamma_{p,h}(x,a)
\end{aligned}
\]
for all $(x,a)\in\Ssc\times\Asc$ and $h\in[H]$.
\end{assumption}

Assumption~\ref{assump:m_estimator_transition} is the self-normalized transition
concentration requirement for the two algorithm-generated value functions. It
must be verified for the value-function class induced by each concrete
M-estimator. Smooth finite-dimensional parameterization makes such a covering
argument possible. Section~\ref{sec:beta_step_level_reward} verifies the same
reward-confidence and induced-value transition-concentration ingredients for
beta regression under its stated step-level model and regularity conditions.

\begin{theorem}[Suboptimality with M-estimated rewards]
\label{thm:m_estimator_suboptimality}
Suppose the transition kernel is linear in $\phi_p$ and
Assumptions~\ref{assump:m_estimator_reward}--\ref{assump:m_estimator_transition}
hold with $\xi_r=\xi_p=\xi/2$. Then, with probability at least $1-\xi$, the
following two statements hold for every initial state $x\in\Ssc$.

First, the policy $\widetilde\pi^M=\{\widetilde\pi_h^M\}_{h=1}^H$ returned by
GRASP under complete reward observation with M-estimated rewards satisfies
\[
    \operatorname{SubOpt}(\widetilde\pi^M;x)
    \le
    2\sum_{h=1}^H
    \E_{\pi^*}\!\left[
        \widetilde\Gamma_{M,h}(x_h,a_h)
        \mid x_1=x
    \right].
\]

Second, the policy $\widehat\pi^M=\{\widehat\pi_h^M\}_{h=1}^H$ returned by
the pooled-transition GRASP recursion under trajectory-level partial reward
observation satisfies
\[
    \operatorname{SubOpt}(\widehat\pi^M;x)
    \le
    2\sum_{h=1}^H
    \E_{\pi^*}\!\left[
        \widehat\Gamma_{M,h}(x_h,a_h)
        \mid x_1=x
    \right].
\]
\end{theorem}

\begin{proof}
Let \(\Esc_M\) be the intersection of the reward-confidence event in
Assumption~\ref{assump:m_estimator_reward} and the transition event in
Assumption~\ref{assump:m_estimator_transition}. By the union bound,
\(\P(\Esc_M)\ge1-\xi\).

Define the population Bellman backup
\[
(\B_h^M V)(x,a)
=
m_h(x,a;\theta_h^\dagger)
+
\langle\phi_p(x,a),\beta_h(V)\rangle.
\]
For the complete-observation recursion, define
\[
(\widetilde\B_h^M\widetilde V_{h+1}^M)(x,a)
=
m_h(x,a;\widetilde\theta_h^M)
+
\langle\phi_p(x,a),\widetilde\beta_h^M\rangle.
\]
On \(\Esc_M\), the triangle inequality gives
\[
\left|
(\B_h^M\widetilde V_{h+1}^M)(x,a)
-
(\widetilde\B_h^M\widetilde V_{h+1}^M)(x,a)
\right|
\le
\widetilde\Gamma_{M,r,h}(x,a)
+
\widetilde\Gamma_{p,h}(x,a)
=
\widetilde\Gamma_{M,h}(x,a).
\]
Thus \(\widetilde\Gamma_{M,h}\) is an uncertainty quantifier at the
algorithm-generated value function.

For the pooled-transition recursion, similarly define
\[
(\widehat\B_h^M\widehat V_{h+1}^M)(x,a)
=
m_h(x,a;\widetilde\theta_h^M)
+
\langle\phi_p(x,a),\widehat\beta_h^M\rangle.
\]
The same two confidence events imply
\[
\left|
(\B_h^M\widehat V_{h+1}^M)(x,a)
-
(\widehat\B_h^M\widehat V_{h+1}^M)(x,a)
\right|
\le
\widetilde\Gamma_{M,r,h}(x,a)
+
\widehat\Gamma_{p,h}(x,a)
=
\widehat\Gamma_{M,h}(x,a).
\]
Hence \(\widehat\Gamma_{M,h}\) is also an uncertainty quantifier at the
corresponding algorithm-generated value function.

Apply the clipping argument in
\eqref{eq:partial-pessimism-residual} separately to the two recursions. Their
Bellman residuals satisfy
\[
0\le\widetilde\iota_h^M(x,a)
\le2\widetilde\Gamma_{M,h}(x,a),
\qquad
0\le\widehat\iota_h^M(x,a)
\le2\widehat\Gamma_{M,h}(x,a).
\]
The performance-difference identity in the proof of
Theorem~\ref{thm:partial_reward_observation_suboptimality}, together with
greediness of \(\widetilde\pi^M\) and \(\widehat\pi^M\), now yields the two
claimed bounds.
\end{proof}

\subsection{Step-Level Reward Observation}
\label{sec:step_level_reward_observation}
This subsection gives the full uncertainty-radius statement and proof for the
step-level reward-observation setting introduced in
Section~\ref{subsec:grasp_reward_observation} and
Corollary~\ref{cor:step_level_reward_observation_rate}. Under the
supplementary convention stated above, \(\Isc_h^p=[n+N]\). We use
\(O_h^\tau\), \(\Isc_h^r\), \(n_h^{\rm obs}\),
\(p_h^{\rm obs}\), \(q_h\), \(\Sigma_h^{\rm obs}\),
\(\rho_h^{\rm obs}\), and \(\mathcal I_{\rm obs}\) as defined in the
main text, together with the reward-observation and transition-coverage
conditions in
Assumption~\ref{assump:step_level_reward_observation_transition_coverage}.
No separate step-level observation mechanism is introduced here.

For the appendix argument, at each step let
\(\widetilde\theta_h^{\rm obs}\) minimize the reward
loss~\eqref{est:theta} over \(\Isc_h^r\), and define
\[
    \widetilde\Sigma_h^{\rm obs}(\theta)
    =
    \sum_{\tau\in\Isc_h^r}
    \dot g\!\left(
        \left\langle\phi_r(x_h^\tau,a_h^\tau),\theta\right\rangle
    \right)
    \phi_r(x_h^\tau,a_h^\tau)
    \phi_r(x_h^\tau,a_h^\tau)^\top.
\]
The corresponding observed-sample reward uncertainty radius is
\[
    \widetilde\Gamma_{r,h}^{\rm obs}(x,a;\theta)
    =
    \alpha_r
    \sqrt{
        \dot g\!\left(
            \left\langle
                \phi_r(x,a),\theta
            \right\rangle
        \right)^2
        \phi_r(x,a)^\top
        \left(
            \widetilde\Sigma_h^{\rm obs}
            (\theta)
        \right)^{-1}
        \phi_r(x,a)
    }.
\]
For the radius used by the algorithm, abbreviate
\[
    \widetilde\Gamma_{r,h}^{\rm obs}(x,a)
    \coloneqq
    \widetilde\Gamma_{r,h}^{\rm obs}(x,a;\widetilde\theta_h^{\rm obs}).
\]
The transition estimator \(\widehat\beta_h\) and transition uncertainty radius
\(\widehat\Gamma_{p,h}\) are computed by the pooled transition step in
Algorithm~\ref{alg:grasp_step_level_reward_observation}. Set
\[
    \widehat\Gamma_h^{\rm obs}(x,a;\theta)
    =
    \widetilde\Gamma_{r,h}^{\rm obs}(x,a;\theta)
    +
    \widehat\Gamma_{p,h}(x,a),
    \qquad
    \widehat\Gamma_h^{\rm obs}(x,a)
    \coloneqq
    \widehat\Gamma_h^{\rm obs}(x,a;\widetilde\theta_h^{\rm obs}).
\]
Let
\[
    \mathcal N_{\rm obs}
    =
    \bigcap_{h=1}^H\left\{n_{h}^{\rm obs}\ge \max \left(m_{\rm obs}, \frac{nq_h}{2}\right)\right\},
\]
where $m_{\rm obs}$ is large enough for the observed-sample concentration
bounds in Lemma~\ref{lem:observed_sample_information}. The supporting lemma
also gives
\(\rho_h^{\rm obs}\ge(\bar p/q_h)\rho\).
Additionally, define \[
\mathcal C =
\{\min_h\lambda_{\min}\{(n+N)^{-1}\widehat\Lambda_h\}\ge\rho_p/2\}.
\]

\begin{theorem}[\texorpdfstring{GRASP with step-level reward observation}{GRASP with step-level reward observation}]\label{thm:step_level_reward_observation}
Suppose
Assumptions~\ref{assump:link}--\ref{assump:step_level_reward_observation_transition_coverage}
hold, and suppose \(\Isc_h^p=[n+N]\) at every step, so all pooled
trajectories contribute a state-action-next-state triple to the transition
regression. If $\lambda=1$,
\[
    \alpha_r
    =
    c_r\sqrt{d_r+\log(H/\xi)},
    \qquad
    \alpha_p
    =
    c_p(d_p+d_r)H
    \sqrt{\log\!\left(2(d_p+d_r)H(n+N)/\xi\right)},
\]
then
\[
\P\!\left(
\mathcal N_{\rm obs}
\cap
\left\{
\operatorname{SubOpt}(\widehat\pi^{\rm obs};x)
>
2\sum_{h=1}^H
\E_{\pi^*}\!\left[
\widehat \Gamma_h^{\rm obs}(x_h,a_h)
\mid x_1=x
\right]
\right\}
\right)
\le \xi.
\]
Thus, outside a joint failure event of probability at most \(\xi\), occurrence
of \(\mathcal N_{\rm obs}\) implies
\[
    \operatorname{SubOpt}(\widehat\pi^{\rm obs};x)
    \le
    2\sum_{h=1}^H
    \E_{\pi^*}\!\left[
        \widehat \Gamma_h^{\rm obs}(x_h,a_h)
        \mid x_1=x
    \right].
\]
The transition-coverage clause of
Assumption~\ref{assump:step_level_reward_observation_transition_coverage}
further yields
\[
    \operatorname{SubOpt}(\widehat\pi^{\rm obs};x)
    \le
    \widetilde O\!\left(
        \sqrt{\frac{d_rH^2}{\mathcal I_{\rm obs}}}
    \right)
    +
    \widetilde O\!\left(
        \sqrt{\frac{(d_p+d_r)^2H^4}{(n+N)\rho_p}}
    \right).
\]
This bound holds unconditionally with probability at least
$1-\xi-\P(\mathcal N_{\rm obs}^c)-\P(\mathcal C^c)$.
Moreover,
positivity gives \(q_h\ge\bar p\); if
\(m_{\rm obs}\le n\bar p/2\), a Chernoff bound and a union bound yield
\[
\P(\mathcal N_{\rm obs}^c)
\le H\exp(-n\bar p/8).
\]
A matrix Bernstein inequality and a union bound also yield
\[
\mathbb P(\mathcal C^c)
\le
2Hd_p
\exp\left\{
-c(n+N)\rho_p^2
\right\},
\]
for some universal constant $c>0$.
Applying the preceding claims at confidence level \(\xi/3\) and taking
\(n\bar p\) and \(n+N\) sufficiently large so that both of the last two
upper bounds are at most \(\xi/3\) therefore
gives an unconditional probability of at least \(1-\xi\).
\end{theorem}

\begin{proof}
For each fixed step and observation-indicator vector satisfying
\(\mathcal N_{\rm obs}\), Lemma~\ref{lem:observed_sample_information}
and a union bound over \(h\) give
\[
\mathcal E_{\rm obs}
=
\bigcap_{h=1}^H
\left\{
\lambda_{\min}\!\left(
(n_h^{\rm obs})^{-1}
\widetilde\Sigma_h^{\rm obs}(\theta_h^*)
\right)
\ge
3\rho_h^{\rm obs}/4
\right\}
\]
with conditional failure probability at most \(\xi/4\). Averaging over the
indicator vectors gives
\(\P(\mathcal E_{\rm obs}^c\cap\mathcal N_{\rm obs})\le\xi/4\).
On this event, the reward-localization and estimation argument in
Lemma~\ref{lem:g}, with
\(n_h^{\rm obs}\) and \(\rho_h^{\rm obs}\) in place of \(n\) and \(\rho\),
gives, jointly over \(h,x,a\),
\[
\left|
g\{\phi_r(x,a)^\top\widetilde\theta_h^{\rm obs}\}
-
g\{\phi_r(x,a)^\top\theta_h^*\}
\right|
\le
\widetilde\Gamma_{r,h}^{\rm obs}(x,a)
\]
with conditional failure probability at most \(\xi/4\) for every indicator
vector in \(\mathcal N_{\rm obs}\), hence with joint failure probability at
most \(\xi/4\) on that event.

The same localization argument also gives within $\mathcal{N}_{\rm obs}$,
\[
\lambda_{\min}\!\left\{
\widetilde\Sigma_h^{\rm obs}(\widetilde\theta_h^{\rm obs})
\right\}
\ge
\frac12 n_h^{\rm obs}\rho_h^{\rm obs}
\ge
\frac{1}{4}n\bar p \rho, 
\] where we utilize Lemma~\ref{lem:observed_sample_information}.
Thus the value-class covering calculation uses
\( n\bar p \rho/4\) as its reward-matrix eigenvalue lower bound.
Applying the pooled-transition concentration argument from
Theorem~\ref{thm:partial_reward_observation_suboptimality},
with the value class defined by
\(\widetilde\theta_h^{\rm obs}\) and
\(\widetilde\Sigma_h^{\rm obs}(\widetilde\theta_h^{\rm obs})\)
gives, under the original unconditional trajectory law and with failure
probability at most \(\xi/2\),
\[
\left|
\left\langle
\phi_p(x,a),
\beta_h(\widehat V_{h+1}^{\rm obs})-\widehat\beta_h
\right\rangle
\right|
\le
\widehat\Gamma_{p,h}(x,a)
\]
simultaneously over \(h,x,a\).

Combining the two componentwise bounds gives
\[
\begin{aligned}
(\widehat\B_{h,\rm obs}\widehat V_{h+1}^{\rm obs})(x,a)
&:=
g\{\phi_r(x,a)^\top\widetilde\theta_h^{\rm obs}\}
+\langle\phi_p(x,a),\widehat\beta_h\rangle,\\
\left|
(\widehat\B_{h,\rm obs}\widehat V_{h+1}^{\rm obs})(x,a)
-
(\B_h\widehat V_{h+1}^{\rm obs})(x,a)
\right|
&\le
\widehat\Gamma_h^{\rm obs}(x,a).
\end{aligned}
\]
On \(\mathcal N_{\rm obs}\) and the intersection of these reward and transition
events, \(\widehat\Gamma_h^{\rm obs}\) is an uncertainty quantifier at the
algorithm-generated value function. Applying the clipping and
performance-difference argument in
\eqref{eq:partial-pessimism-residual} yields
\[
\operatorname{SubOpt}(\widehat\pi^{\rm obs};x)
\le
2\sum_{h=1}^H
\E_{\pi^*}\!\left[
\widehat\Gamma_h^{\rm obs}(x_h,a_h)
\mid x_1=x
\right].
\]

Finally, the reward-curvature event gives
\[
\widetilde\Gamma_{r,h}^{\rm obs}(x,a)
=
\widetilde O\!\left(
\sqrt{\frac{d_r}{n_h^{\rm obs}\rho_h^{\rm obs}}}
\right),
\]
whereas transition covariance concentration gives
\[
\widehat\Gamma_{p,h}(x,a)
=
\widetilde O\!\left(
\sqrt{\frac{(d_p+d_r)^2H^2}{(n+N)\rho_p}}
\right).
\]
Summing over \(h\) proves the displayed two-term rate. The two reward failures
intersected with \(\mathcal N_{\rm obs}\) are at most \(\xi/4\) each, and the
unconditional transition failure is at most \(\xi/2\). Their union therefore
has probability at most \(\xi\) jointly with \(\mathcal N_{\rm obs}\).
Under the cross-trajectory independence condition in
Assumption~\ref{assump:step_level_reward_observation_transition_coverage},
standard concentration gives \(n_h^{\rm obs}\asymp nq_h\) uniformly over
\(h\). Since
\(\rho_h^{\rm obs}\ge(\bar p/q_h)\rho\), it follows that
\(\mathcal I_{\rm obs}\gtrsim n\bar p\rho\).
When \(p_h^{\rm obs}(x,a)\equiv q_h\), the conditional observed-sample
information equals the full-sample information, so
\(\rho_h^{\rm obs}\ge\rho\); only the observed sample size is reduced.
\end{proof}

\subsection{Beta Regression with Step-Level Reward Observation}
\label{sec:beta_step_level_reward}

We now verify the reward-confidence and induced-value transition-concentration
ingredients for a beta reward model \citep{ferrari2004beta} with an
unknown scalar precision and step-level reward observation.  We first give the
unregularized result and then a deterministic-regularization extension.  Both
results concern the exact conditional beta model specified below.  The
discrete EDSS-derived reward in
Section~\ref{sec:real_data} uses beta regression as a continuous working
approximation rather than asserting this sampling model.

For \(z=(x,a)\), let
\[
m_\beta(z;\vartheta)
=
\operatorname{expit}\{\phi_r(z)^\top\theta\},
\qquad
\kappa(\vartheta)=\exp(\nu),
\qquad
\vartheta=(\theta^\top,\nu)^\top\in\mathbb R^{d_\beta},
\quad d_\beta=d_r+1.
\]
Here \(\operatorname{expit}(u)=\{1+\exp(-u)\}^{-1}\).
For \(y\in(0,1)\), define
\[
f_\beta(y\mid z;\vartheta)
=
\frac{\Gamma\{\kappa(\vartheta)\}}
{\Gamma\{\kappa(\vartheta)m_\beta(z;\vartheta)\}
 \Gamma[\kappa(\vartheta)\{1-m_\beta(z;\vartheta)\}]}
y^{\kappa(\vartheta)m_\beta(z;\vartheta)-1}
(1-y)^{\kappa(\vartheta)\{1-m_\beta(z;\vartheta)\}-1}
\]
and \(\ell_\beta(y,z;\vartheta)=-\log f_\beta(y\mid z;\vartheta)\).
Let
\[
P_\beta=\operatorname{diag}(\bI_{d_r},0),
\qquad
a_\beta(z;\vartheta)
=
\nabla_\vartheta m_\beta(z;\vartheta)
=
\bigl[
m_\beta(z;\vartheta)\{1-m_\beta(z;\vartheta)\}\phi_r(z)^\top,
0
\bigr]^\top .
\]

\begin{assumption}[Regular beta reward model]
\label{assump:beta_reward_regular}
For each \(h\in[H]\), the following conditions hold.
\begin{enumerate}[(i)]
\item Conditional on \(z_h=(x_h,a_h)\),
\[
r_h\sim
\operatorname{Beta}\!\left(
\kappa_h^*m_{\beta,h}^*,
\kappa_h^*\{1-m_{\beta,h}^*\}
\right),
\qquad
m_{\beta,h}^*
=\operatorname{expit}\{\phi_r(z_h)^\top\theta_h^*\},
\]
where \(\vartheta_h^*=(\theta_h^{*\top},\log\kappa_h^*)^\top\) is an
interior point of a compact parameter set
\(\Psi_\beta=\Theta_\beta\times[\nu_-,\nu_+]\).
\item There is an \(\epsilon_\beta\in(0,1/2)\) such that
\[
\epsilon_\beta
\le m_\beta(z;\vartheta)
\le 1-\epsilon_\beta
\]
for all \(z\) and \(\vartheta\in\Psi_\beta\).  The conditional beta model is
identifiable under the observed state-action design.
\item With
\[
\mathcal J_{\beta,h}^{\rm obs}
=
\E_{\pi^b}\!\left[
\nabla_\vartheta^2
\ell_\beta(r_h,z_h;\vartheta_h^*)
\mid O_h=1
\right],
\qquad
\rho_{\beta,h}^{\rm obs}
=\lambda_{\min}(\mathcal J_{\beta,h}^{\rm obs}),
\]
we have \(\rho_{\beta,h}^{\rm obs}>0\).
\end{enumerate}
\end{assumption}

Assumption~\ref{assump:beta_reward_regular} keeps both beta shape parameters
uniformly away from zero and infinity.  Consequently, the beta score and
likelihood derivatives through third order have uniformly bounded
sub-exponential norms.  The information condition covers the joint mean and
scalar-precision parameters.

For a prespecified deterministic reward penalty
\(\lambda_{\beta,h}\ge0\), distinct from the transition-ridge parameter
\(\lambda\), define
\[
\widehat\vartheta_{\beta,h}^{\rm reg}
\in
\arg\min_{\vartheta=(\theta^\top,\nu)^\top\in\Psi_\beta}
\left[
\frac{1}{n_h^{\rm obs}}
\sum_{\tau\in\Isc_h^r}
\ell_\beta(r_h^\tau,z_h^\tau;\vartheta)
+
\frac{\lambda_{\beta,h}}{2}\|\theta\|_2^2
\right].
\]
Write \(\widehat\theta_{\beta,h}^{\rm reg}\) for the first \(d_r\)
components of \(\widehat\vartheta_{\beta,h}^{\rm reg}\); as below, the
superscript is omitted when \(\lambda_{\beta,h}=0\).
Let its summed penalized observed information be
\[
\widehat{\mathcal J}_{\beta,h}^{\rm reg}(\vartheta)
=
\sum_{\tau\in\Isc_h^r}
\nabla_\vartheta^2\ell_\beta(r_h^\tau,z_h^\tau;\vartheta)
+
n_h^{\rm obs}\lambda_{\beta,h}P_\beta .
\]
Write \(B_\theta=\sup_{\theta\in\Theta_\beta}\|\theta\|_2\), and define
\[
\begin{aligned}
\widehat\Gamma_{\beta,r,h}^{\rm reg}(z)
&=
\alpha_\beta
\left\|
a_\beta(z;\widehat\vartheta_{\beta,h}^{\rm reg})
\right\|_{
\widehat{\mathcal J}_{\beta,h}^{\rm reg}
(\widehat\vartheta_{\beta,h}^{\rm reg})^{-1}}
+
b_{\beta,h},\\
\alpha_\beta
&=
C_\beta\sqrt{d_\beta+\log(2H/\xi)},
\qquad
b_{\beta,h}
=
C_\beta
\frac{\lambda_{\beta,h}B_\theta}
{\rho_{\beta,h}^{\rm obs}},
\end{aligned}
\]
where \(C_\beta>0\) depends only on the fixed constants in
Assumption~\ref{assump:beta_reward_regular}. A known lower bound on
\(\rho_{\beta,h}^{\rm obs}\) may be substituted in \(b_{\beta,h}\). The inverse
is well defined on the high-probability event in the results below. When
\(\lambda_{\beta,h}=0\), we omit the superscript \({\rm reg}\) and
\(b_{\beta,h}=0\).

\begin{lemma}[Beta likelihood event]
\label{lem:beta_likelihood_event}
Let
\[
\rho_{\beta,\min}^{\rm obs}
=
\min_{h\in[H]}\rho_{\beta,h}^{\rm obs},
\qquad
D_\beta=d_\beta+\log(8H/\xi),
\]
where \(\xi\in(0,1)\). There are constants \(C_{\rm lik},c_0,C_0>0\)
and fixed convex neighborhoods \(U_h\) whose closures lie in
\({\rm int}(\Psi_\beta)\) such that, with
\[
m_{\rm BR}=\left\lceil C_{\rm lik}D_\beta\right\rceil,
\qquad
\mathcal N_\beta
=
\left\{\min_{h\in[H]}n_h^{\rm obs}\ge m_{\rm BR}\right\},
\]
there is an event \(\mathcal L_\beta\) satisfying
\[
\P(\mathcal L_\beta^c\cap\mathcal N_\beta)\le\xi/2.
\]
On \(\mathcal L_\beta\), simultaneously over \(h\),
\[
\left\|
\frac{1}{n_h^{\rm obs}}
\sum_{\tau\in\Isc_h^r}
\nabla_\vartheta\ell_\beta(r_h^\tau,z_h^\tau;\vartheta_h^*)
\right\|_{(\mathcal J_{\beta,h}^{\rm obs})^{-1}}
\le
C_0\sqrt{\frac{D_\beta}{n_h^{\rm obs}}},
\]
and, for every \(\vartheta\in U_h\),
\[
c_0\mathcal J_{\beta,h}^{\rm obs}
\preceq
\frac{1}{n_h^{\rm obs}}
\sum_{\tau\in\Isc_h^r}
\nabla_\vartheta^2
\ell_\beta(r_h^\tau,z_h^\tau;\vartheta)
\preceq
C_0\mathcal J_{\beta,h}^{\rm obs}.
\]
The unregularized estimator \(\widehat\vartheta_{\beta,h}\) lies in \(U_h\).
Moreover, if
\[
\max_{h\in[H]}
\lambda_{\beta,h}B_\theta^2
\le c_{\rm loc}
\]
for a sufficiently small fixed \(c_{\rm loc}>0\), then
\(\widehat\vartheta_{\beta,h}^{\rm reg}\in U_h\) as well.
The constants may depend on
\(\epsilon_\beta,[\nu_-,\nu_+]\), the compact parameter set, its
identification gap, and \(\rho_{\beta,\min}^{\rm obs}\), but not on
\(n,N,H\), or \(\xi\).
\end{lemma}

\begin{proof}
Fix \(h\in[H]\), and write
\[
\mathcal O_h
:=
\sigma\!\left(O_h^1,\ldots,O_h^n\right),
\qquad
m_h:=n_h^{\rm obs},
\qquad
\Isc_h^r=\{\tau\in[n]:O_h^\tau=1\}.
\]
Throughout the proof, we condition on \(\mathcal O_h\) and work on the
event \(\{m_h\ge m_{\rm BR}\}\).

Let \(Z_h=(X_h,A_h)\) and let \(R_h\) denote the corresponding reward for a generic trajectory under \(\pi^b\).

For \(\vartheta\in\Psi_\beta\), define the empirical and population
observed-data risks by
\[
\widehat{\mathcal R}_h(\vartheta)
:=
\frac{1}{m_h}
\sum_{\tau\in\Isc_h^r}
\ell_\beta
\left(
r_h^\tau,z_h^\tau;\vartheta
\right),
\]
and
\[
\mathcal R_h(\vartheta)
:=
\mathbb E_{\pi^b}
\left[
\ell_\beta(R_h,Z_h;\vartheta)
\,\middle|\,
O_h=1
\right].
\]
Also define
\[
\widehat{\mathcal J}_h(\vartheta)
:=
\nabla_\vartheta^2
\widehat{\mathcal R}_h(\vartheta),
\qquad
\mathcal J_h(\vartheta)
:=
\nabla_\vartheta^2
\mathcal R_h(\vartheta),
\]
so that
\[
\mathcal J_h(\vartheta_h^*)
=
\mathcal J_{\beta,h}^{\rm obs}.
\]

By Assumption~\ref{assump:beta_reward_regular}, the beta mean is uniformly bounded away from
\(0\) and \(1\), and the precision parameter ranges over a compact
subset of \((0,\infty)\). Consequently, \(-\log R_h\) and
\(-\log(1-R_h)\) have uniformly bounded sub-exponential norms under the
observed-design law.

Moreover, the beta log-likelihood and its first three derivatives are
combinations of bounded feature terms, polygamma functions evaluated
over a compact subset of \((0,\infty)\), and the random variables
\(\log R_h\) and \(\log(1-R_h)\). Using \(\|\cdot\|_{\psi_1}\) to denote the sub-exponential Orlicz norm, for \(j=0,1,2,3\), there
exists a constant \(K_j<\infty\), independent of \(h\), such that
\[
\left\|
\sup_{\vartheta\in\Psi_\beta}
\left\|
\nabla_\vartheta^j
\ell_\beta(R_h,Z_h;\vartheta)
\right\|_{\rm op}
\right\|_{\psi_1}
\le K_j.
\]

Correct specification and the information identity give
\[
\mathbb E_{\pi^b}
\left[
\nabla_\vartheta
\ell_\beta(R_h,Z_h;\vartheta_h^*)
\,\middle|\,
O_h=1
\right]
=0
\]
and
\[
\mathbb E_{\pi^b}
\left[
\nabla_\vartheta
\ell_\beta(R_h,Z_h;\vartheta_h^*)
\nabla_\vartheta
\ell_\beta(R_h,Z_h;\vartheta_h^*)^\top
\,\middle|\,
O_h=1
\right]
=
\mathcal J_{\beta,h}^{\rm obs}.
\]

\paragraph*{Score concentration.}
Let
\[
\widehat s_h
:=
\nabla_\vartheta
\widehat{\mathcal R}_h(\vartheta_h^*)
=
\frac{1}{m_h}
\sum_{\tau\in\Isc_h^r}
s_h^\tau,
\]
where
\[
s_h^\tau
:=
\nabla_\vartheta
\ell_\beta
\left(
r_h^\tau,z_h^\tau;\vartheta_h^*
\right).
\]
Define the whitened scores
\[
X_h^\tau
:=
\left(
\mathcal J_{\beta,h}^{\rm obs}
\right)^{-1/2}
s_h^\tau.
\]
Conditionally on \(\mathcal O_h\),
\[
\mathbb E[X_h^\tau\mid\mathcal O_h]=0,
\qquad
\mathbb E[
X_h^\tau(X_h^\tau)^\top
\mid\mathcal O_h
]
=
I_{d_\beta}.
\]
The derivative-envelope bound above implies that
\[
\sup_{\|u\|_2=1}
\left\|
u^\top X_h^\tau
\right\|_{\psi_1}
\le K_{\rm sc}
\]
for a finite constant \(K_{\rm sc}\).

Applying the Bernstein inequality on a \(1/2\)-net of
\(\mathbb S^{d_\beta-1}\), followed by a union bound, yields
\[
\left\|
\widehat s_h
\right\|_{
(\mathcal J_{\beta,h}^{\rm obs})^{-1}
}
\le
C_{\rm sc}
\left[
\sqrt{
\frac{
d_\beta+\log(8H/\xi)
}{
m_h
}
}
+
K_{\rm sc}
\frac{
d_\beta+\log(8H/\xi)
}{
m_h
}
\right]
\]
with conditional probability at least \(1-\xi/(8H)\).

By choosing \(m_{\rm BR}\) sufficiently large, the second term is
absorbed into the first, and hence
\begin{equation}
\label{eq:beta-score-bound}
\left\|
\widehat s_h
\right\|_{
(\mathcal J_{\beta,h}^{\rm obs})^{-1}
}
\le
C_{\rm sc}
\sqrt{
\frac{
d_\beta+\log(8H/\xi)
}{
m_h
}
}
\end{equation}
with conditional probability at least \(1-\xi/(8H)\).

\paragraph*{Uniform local Hessian comparison.}
Since
\[
\lambda_{\min}
\left(
\mathcal J_{\beta,h}^{\rm obs}
\right)
=
\rho_{\beta,h}^{\rm obs}
>0
\]
and \(\mathcal J_h(\vartheta)\) is continuous in \(\vartheta\), the
neighborhood \(U_h\) may be chosen so that
\begin{equation}
\label{eq:beta-population-hessian-comparison}
\frac{3}{4}
\mathcal J_{\beta,h}^{\rm obs}
\preceq
\mathcal J_h(\vartheta)
\preceq
\frac{5}{4}
\mathcal J_{\beta,h}^{\rm obs},
\qquad
\vartheta\in U_h.
\end{equation}

For a fixed \(\vartheta\in U_h\), the matrix-valued summands
\[
\nabla_\vartheta^2
\ell_\beta
\left(
r_h^\tau,z_h^\tau;\vartheta
\right)
-
\mathcal J_h(\vartheta)
\]
are conditionally independent, centered, self-adjoint, and possess
uniform sub-exponential envelopes. A matrix Bernstein inequality,
combined with $\mathcal{N}_h$, a finite Euclidean net of \(U_h\), therefore gives
\[
\sup_{\vartheta\in\mathcal N_h}
\left\|
\left(
\mathcal J_{\beta,h}^{\rm obs}
\right)^{-1/2}
\left\{
\widehat{\mathcal J}_h(\vartheta)
-
\mathcal J_h(\vartheta)
\right\}
\left(
\mathcal J_{\beta,h}^{\rm obs}
\right)^{-1/2}
\right\|_{\rm op}
\le
\frac{1}{8}
\]
with conditional probability at least \(1-\xi/(8H)\), provided
\(m_h\ge m_{\rm BR}\).

The third-derivative envelope above implies that, on an event
of the same probability order,
\[
\left\|
\widehat{\mathcal J}_h(\vartheta)
-
\widehat{\mathcal J}_h(\vartheta')
\right\|_{\rm op}
\le
C_{\rm H}
\|\vartheta-\vartheta'\|_2,
\qquad
\vartheta,\vartheta'\in U_h.
\]
Choosing the mesh of \(\mathcal N_h\) sufficiently fine extends the
preceding matrix bound from \(\mathcal N_h\) to all of \(U_h\), yielding
\[
\sup_{\vartheta\in U_h}
\left\|
\left(
\mathcal J_{\beta,h}^{\rm obs}
\right)^{-1/2}
\left\{
\widehat{\mathcal J}_h(\vartheta)
-
\mathcal J_h(\vartheta)
\right\}
\left(
\mathcal J_{\beta,h}^{\rm obs}
\right)^{-1/2}
\right\|_{\rm op}
\le
\frac{1}{4}.
\]
Combining this bound with
\eqref{eq:beta-population-hessian-comparison} gives
\begin{equation}
\label{eq:beta-empirical-hessian-comparison}
\frac{1}{2}
\mathcal J_{\beta,h}^{\rm obs}
\preceq
\widehat{\mathcal J}_h(\vartheta)
\preceq
\frac{3}{2}
\mathcal J_{\beta,h}^{\rm obs},
\qquad
\vartheta\in U_h.
\end{equation}

\paragraph*{Global localization of the unregularized minimizer.}
Because \(\Psi_\beta\) is compact, \(\vartheta_h^*\) is identified, and
\(U_h\) is a neighborhood of \(\vartheta_h^*\), the population
identification gap
\[
\Delta_{{\rm id},h}
:=
\inf_{\vartheta\in\Psi_\beta\setminus U_h}
\left\{
\mathcal R_h(\vartheta)
-
\mathcal R_h(\vartheta_h^*)
\right\}
\]
is strictly positive.

The sub-exponential envelope for the loss and a finite net over
\(\Psi_\beta\) imply the uniform objective bound
\begin{equation}
\label{eq:beta-objective-concentration}
\sup_{\vartheta\in\Psi_\beta}
\left|
\widehat{\mathcal R}_h(\vartheta)
-
\mathcal R_h(\vartheta)
\right|
\le
\frac{\Delta_{{\rm id},h}}{4}
\end{equation}
with conditional probability at least \(1-\xi/(8H)\), provided
\(m_h\ge m_{\rm BR}\).

Since \(\Psi_\beta\) is compact and
\(\widehat{\mathcal R}_h\) is continuous, an empirical minimizer
\[
\widehat\vartheta_{\beta,h}
\in
\arg\min_{\vartheta\in\Psi_\beta}
\widehat{\mathcal R}_h(\vartheta)
\]
exists. For any \(\vartheta\in\Psi_\beta\setminus U_h\),
\[
\begin{aligned}
\widehat{\mathcal R}_h(\vartheta)
-
\widehat{\mathcal R}_h(\vartheta_h^*)
&\ge
\mathcal R_h(\vartheta)
-
\mathcal R_h(\vartheta_h^*)
\\
&\quad
-
2
\sup_{\vartheta'\in\Psi_\beta}
\left|
\widehat{\mathcal R}_h(\vartheta')
-
\mathcal R_h(\vartheta')
\right|
\\
&\ge
\Delta_{{\rm id},h}
-
\frac{\Delta_{{\rm id},h}}{2}
\\
&=
\frac{\Delta_{{\rm id},h}}{2}
>0.
\end{aligned}
\]
Hence every empirical minimizer satisfies
\[
\widehat\vartheta_{\beta,h}\in U_h.
\]
Because \(U_h\subset\operatorname{int}(\Psi_\beta)\), the minimizer is
interior and therefore obeys
\[
\nabla_\vartheta
\widehat{\mathcal R}_h
\left(
\widehat\vartheta_{\beta,h}
\right)
=0.
\]

\paragraph*{ Localization of the regularized minimizer.}
Let
\[
\widehat{\mathcal R}_{h,\lambda}(\vartheta)
:=
\widehat{\mathcal R}_h(\vartheta)
+
\frac{\lambda_{\beta,h}}{2}
\vartheta^\top P_\beta\vartheta.
\]
By the definition of \(B_\theta\),
\[
\sup_{\vartheta\in\Psi_\beta}
\left|
\frac{\lambda_{\beta,h}}{2}
\left\{
\vartheta^\top P_\beta\vartheta
-
(\vartheta_h^*)^\top
P_\beta\vartheta_h^*
\right\}
\right|
\le
\lambda_{\beta,h}B_\theta^2.
\]
The stated small-penalty condition, after choosing the localization
constant sufficiently small, ensures that
\[
\lambda_{\beta,h}B_\theta^2
\le
\frac{\Delta_{{\rm id},h}}{4}.
\]
Consequently, for every
\(\vartheta\in\Psi_\beta\setminus U_h\),
\[
\widehat{\mathcal R}_{h,\lambda}(\vartheta)
-
\widehat{\mathcal R}_{h,\lambda}(\vartheta_h^*)
\ge
\frac{\Delta_{{\rm id},h}}{4}
>0.
\]
Thus every regularized empirical minimizer
\[
\widehat\vartheta_{\beta,h}^{\rm reg}
\in
\arg\min_{\vartheta\in\Psi_\beta}
\widehat{\mathcal R}_{h,\lambda}(\vartheta)
\]
also belongs to \(U_h\subset\operatorname{int}(\Psi_\beta)\).

Finally, let \(\mathcal L_{\beta,h}\) denote the intersection of the score,
uniform-Hessian, objective-localization, and regularized-localization
events established above. By allocating failure probability among
the preceding steps,
\[
\mathbb P
\left(
\mathcal L_{\beta,h}^c
\,\middle|\,
\mathcal O_h
\right)
\le
\frac{\xi}{2H}
\]
whenever \(m_h\ge m_{\rm BR}\).

Since
\[
\mathcal N_\beta
=
\left\{
\min_{h\in[H]}n_h^{\rm obs}
\ge
m_{\rm BR}
\right\},
\qquad
\mathcal L_\beta
=
\bigcap_{h=1}^H
\mathcal L_{\beta,h},
\]
we have
\[
\mathcal L_\beta^c\cap\mathcal N_\beta
\subseteq
\bigcup_{h=1}^H
\left(
\mathcal L_{\beta,h}^c
\cap
\{n_h^{\rm obs}\ge m_{\rm BR}\}
\right).
\]
Hence,
\[
\begin{aligned}
\mathbb P
\left(
\mathcal L_\beta^c
\cap
\mathcal N_\beta
\right)
&\le
\sum_{h=1}^H
\mathbb E
\left[
\mathbf 1
\{n_h^{\rm obs}\ge m_{\rm BR}\}
\mathbb P
\left(
\mathcal L_{\beta,h}^c
\,\middle|\,
\mathcal O_h
\right)
\right]
\\
&\le
\sum_{h=1}^H
\frac{\xi}{2H}
=
\frac{\xi}{2}.
\end{aligned}
\]
This proves the claim.
\end{proof}

\begin{lemma}[Beta-generated value-class covering]
\label{lem:beta_value_cover}
For fixed nonnegative \(b_0\), let \(\mathcal V_{\beta,h}\) contain the
functions
\[
\begin{aligned}
V_h(x;\omega)
=\max_{a\in\Asc}\Big[{}&
m_\beta(z;\vartheta)+\phi_p(z)^\top b
-\{a_\beta(z;\vartheta)^\top M_\beta
a_\beta(z;\vartheta)\}^{1/2}-b_0\\
&-\{\phi_p(z)^\top M_p\phi_p(z)\}^{1/2}
\Big]_0^{H-h+1},
\qquad z=(x,a),
\end{aligned}
\]
where \([u]_0^M=\min\{M,\max(0,u)\}\). The class is indexed by
\(\omega=(\vartheta,b,M_\beta,M_p)\), with
\(\vartheta\in\Psi_\beta\), \(\|b\|_2\le B_p\), and
\(M_\beta,M_p\) are positive semidefinite with
\(\|M_\beta\|_{\rm F}\le R_\beta\) and
\(\|M_p\|_{\rm F}\le R_p\). Under the standing feature normalization,
\[
\log\mathcal N(\epsilon,\mathcal V_{\beta,h},\|\cdot\|_\infty)
\le
C(d_\beta+d_p)^2
\log\!\left\{
\frac{C'(1+B_p+R_\beta+R_p)}{\epsilon}
\right\}.
\]
Here \(C'\) may depend on the compact beta-model constants.
\end{lemma}

\begin{proof}
On \(\Psi_\beta\), both \(m_\beta(z;\vartheta)\) and
\(a_\beta(z;\vartheta)\) are uniformly Lipschitz in \(\vartheta\). The clipped
maximum is nonexpansive, and
\[
\left|
\sqrt{u^\top Mu}-\sqrt{u^\top M'u}
\right|
\le
\|M-M'\|_{\rm F}^{1/2}\|u\|_2.
\]
Euclidean nets for \(\vartheta\) and \(b\), together with Frobenius nets of
mesh order \(\epsilon^2\) for \(M_\beta\) and \(M_p\), therefore cover the
class. Lemma~\ref{lem:cover} gives vector exponents \(d_\beta,d_p\) and matrix
exponents \(d_\beta^2,d_p^2\), which proves the displayed order.
\end{proof}

\begin{theorem}[Unregularized beta regression with step-level reward observation]
\label{thm:beta_step_level_reward}
Under the linear transition model~\eqref{eq:grasp-model}, the standing feature
normalization, and the standing i.i.d. pooled-trajectory sampling conditions,
suppose Assumption~\ref{assump:beta_reward_regular} holds and the
reward-observation mechanism satisfies the missing-at-random, positivity, and
cross-trajectory sampling clauses of
Assumption~\ref{assump:step_level_reward_observation_transition_coverage}.
Suppose all \(n+N\) pooled
trajectories contribute a transition triple at every step. For
\(\xi\in(0,1)\), let
\[
\mathcal I_{\beta,\rm obs}
=
\min_{h\in[H]}
n_h^{\rm obs}\rho_{\beta,h}^{\rm obs}.
\]
Use the unregularized beta estimator
\(\widehat\vartheta_{\beta,h}\), corresponding to
\(\lambda_{\beta,h}=0\), and run
Algorithm~\ref{alg:grasp_step_level_reward_observation} with
\(m_\beta(z;\widehat\vartheta_{\beta,h})\) and
\(\widehat\Gamma_{\beta,r,h}\) in place of the GLM reward fit and radius,
while retaining all available transitions in the transition update. Set the
transition-ridge parameter \(\lambda=1\), and denote the resulting value
functions, transition coefficients, and policy by
\(\widehat V_h^{\rm BR}\), \(\widehat\beta_h^{\rm BR}\), and
\(\widehat\pi^{\rm BR}\), respectively. Set
\[
\alpha_p
=
c_p(d_p+d_\beta)H
\sqrt{\log\{2(d_p+d_\beta)H(n+N)/\xi\}},
\qquad
\widehat\Gamma_{\beta,h}(z)
=
\widehat\Gamma_{\beta,r,h}(z)+\widehat\Gamma_{p,h}(z).
\]
Then there is an event \(\mathcal E_\beta\) such that
\[
\P(\mathcal E_\beta^c\cap\mathcal N_\beta)\le\xi
\]
and, on \(\mathcal E_\beta\cap\mathcal N_\beta\),
\[
\operatorname{SubOpt}(\widehat\pi^{\rm BR};x)
\le
2\sum_{h=1}^H
\E_{\pi^*}\!\left[
\widehat\Gamma_{\beta,h}(z_h)
\mid x_1=x
\right],
\qquad z_h=(x_h,a_h).
\]
If additionally
\(\lambda_{\min}(\Lambda_h)\ge\rho_p>0\) for all \(h\), then
\[
\begin{aligned}
\operatorname{SubOpt}(\widehat\pi^{\rm BR};x)
&\le
\widetilde O\!\left(
\sqrt{\frac{d_\beta H^2}{\mathcal I_{\beta,\rm obs}}}
\right)
\\[-2pt]
&\quad+
\widetilde O\!\left(
\sqrt{
\frac{(d_p+d_\beta)^2H^4}
{(n+N)\rho_p}
}
\right).
\end{aligned}
\]
Let \(\mathcal C\) denote the pooled transition-coverage event defined in
Supplementary~\ref{sec:step_level_reward_observation}.
This bound holds unconditionally with probability at least
$1-\xi-\P(\mathcal N_{\beta}^c)-\P(\mathcal C^c)$.
Moreover,
positivity gives \(q_h\ge\bar p\); if
\(m_{\rm BR}\le n\bar p/2\), a Chernoff bound and a union bound yield
\[
\P(\mathcal N_{\beta}^c)
\le H\exp(-n\bar p/8).
\]
A matrix Bernstein inequality and a union bound also yield
\[
\mathbb P(\mathcal C^c)
\le
2Hd_p
\exp\left\{
-c(n+N)\rho_p^2
\right\},
\]
for some universal constant $c>0$.
Applying the preceding claims at confidence level \(\xi/3\) and taking
\(n\bar p\) and \(n+N\) sufficiently large so that both of the last two
upper bounds are at most \(\xi/3\) therefore
gives an unconditional probability of at least \(1-\xi\).
\end{theorem}

\begin{proof}
Work on \(\mathcal N_\beta\cap\mathcal L_\beta\), with
\(\mathcal L_\beta\) from Lemma~\ref{lem:beta_likelihood_event}. Fix \(h\), set
\[
g_{\beta,h}
=
\frac{1}{n_h^{\rm obs}}
\sum_{\tau\in\Isc_h^r}
\nabla_\vartheta
\ell_\beta(r_h^\tau,z_h^\tau;\vartheta_h^*),
\qquad
\Delta_h=\widehat\vartheta_{\beta,h}-\vartheta_h^*,
\]
and define the average path Hessian
\[
\overline{\mathcal J}_{\beta,h}
=
\int_0^1
\frac{1}{n_h^{\rm obs}}
\sum_{\tau\in\Isc_h^r}
\nabla_\vartheta^2\ell_\beta
(r_h^\tau,z_h^\tau;\vartheta_h^*+t\Delta_h)\,dt.
\]
Lemma~\ref{lem:beta_likelihood_event} places the estimator and the entire path
inside \(U_h\), so
\[
c_0\mathcal J_{\beta,h}^{\rm obs}
\preceq
\overline{\mathcal J}_{\beta,h}
\preceq
C_0\mathcal J_{\beta,h}^{\rm obs}.
\]

Because the localized minimizer is interior, its score equation and the
fundamental theorem of calculus give
\(0=g_{\beta,h}+\overline{\mathcal J}_{\beta,h}\Delta_h\). Hence
\[
\left\|
\widehat\vartheta_{\beta,h}-\vartheta_h^*
\right\|_{\mathcal J_{\beta,h}^{\rm obs}}
\le
C\sqrt{\frac{D_\beta}{n_h^{\rm obs}}}.
\]
At the endpoint, the same two-sided likelihood event gives
\[
c_0n_h^{\rm obs}\mathcal J_{\beta,h}^{\rm obs}
\preceq
\widehat{\mathcal J}_{\beta,h}
(\widehat\vartheta_{\beta,h})
\preceq
C_0n_h^{\rm obs}\mathcal J_{\beta,h}^{\rm obs}.
\]

Let \(\eta^*=\phi_r(z)^\top\theta_h^*\) and
\(\widehat\eta=\phi_r(z)^\top\widehat\theta_{\beta,h}\). The mean-value
theorem gives, for an intermediate \(\widetilde\eta\),
\[
m_\beta(z;\widehat\vartheta_{\beta,h})
-m_\beta(z;\vartheta_h^*)
=
\frac{\dot m(\widetilde\eta)}{\dot m(\widehat\eta)}
a_\beta(z;\widehat\vartheta_{\beta,h})^\top\Delta_h.
\]
Here $\dot m(u)=\operatorname{expit}(u)\{1-\operatorname{expit}(u)\}$.
The derivative ratio is uniformly bounded because all means in the local path
belong to \([\epsilon_\beta,1-\epsilon_\beta]\). The preceding two-sided norm
comparisons imply
\(\|\Delta_h\|_{\widehat{\mathcal J}_{\beta,h}
(\widehat\vartheta_{\beta,h})}\le C\sqrt{D_\beta}\). Cauchy--Schwarz therefore
gives
\[
\left|
m_\beta(z;\widehat\vartheta_{\beta,h})
-
m_\beta(z;\vartheta_h^*)
\right|
\le
\widehat\Gamma_{\beta,r,h}(z)
\]
simultaneously over \(z\) and \(h\), after increasing \(C_\beta\) by a fixed
factor. Since \(\|a_\beta(z;\vartheta)\|_2\le1/4\), it also gives
\[
\widehat\Gamma_{\beta,r,h}(z)
=
O\!\left[
\sqrt{
\frac{D_\beta}
{n_h^{\rm obs}\rho_{\beta,h}^{\rm obs}}
}
\right].
\]

For transition concentration, write
\[
M_{\beta,h}
=
\alpha_\beta^2
\widehat{\mathcal J}_{\beta,h}
(\widehat\vartheta_{\beta,h})^{-1},
\qquad
M_{p,h}
=
\alpha_p^2(\widehat\Lambda_h+\bI_{d_p})^{-1}.
\]
The endpoint information bound places \(M_{\beta,h}\) in a deterministic
Frobenius ball, while \(\|M_{p,h}\|_{\rm F}\le\sqrt{d_p}\alpha_p^2\).
Lemma~\ref{lem:norm} gives
\(\|\widehat\beta_h^{\rm BR}\|_2\le H\sqrt{n+N}\). Hence
Lemma~\ref{lem:beta_value_cover} contains every algorithm-generated value
function and gives
\[
\log\mathcal N_\beta(\epsilon)
\le
C(d_\beta+d_p)^2
\log\!\left\{
\frac{C'(n+N)H}{\epsilon\xi}
\right\}.
\]
Applying the finite-net and self-normalized transition argument from
Section~\ref{sec:shared_bellman_uncertainty}, with failure probability
\(\xi/2\) under the original unconditional trajectory law, therefore gives
\[
\left|
\left\langle
\phi_p(x,a),
\beta_h(\widehat V_{h+1}^{\rm BR})-\widehat\beta_h^{\rm BR}
\right\rangle
\right|
\le
\widehat\Gamma_{p,h}(x,a)
\]
simultaneously over \(h,x,a\).

On \(\mathcal N_\beta\cap\mathcal L_\beta\) and this transition event, the
reward and transition bounds make \(\widehat\Gamma_{\beta,h}\) an uncertainty
quantifier at every algorithm-generated continuation value. The clipping and
performance-difference argument in
\eqref{eq:partial-pessimism-residual} proves the first claim.  Finally,
summing the reward rate over \(h\), using
\(\mathcal I_{\beta,\rm obs}
=\min_h n_h^{\rm obs}\rho_{\beta,h}^{\rm obs}\), and applying transition
covariance concentration proves the displayed two-term rate. Taking
\(\mathcal E_\beta\) as the intersection of \(\mathcal L_\beta\) and the
transition event gives
\(\P(\mathcal E_\beta^c\cap\mathcal N_\beta)\le\xi\), because the likelihood
failure intersected with \(\mathcal N_\beta\) and the unconditional transition
failure are each at most \(\xi/2\).
\end{proof}

\begin{corollary}[Regularized beta regression with step-level reward observation]
\label{cor:beta_step_level_reward_regularized}
Under the assumptions and sample-size conditions of
Theorem~\ref{thm:beta_step_level_reward}, let
\(\lambda_{\beta,h}\ge0\) be deterministic and suppose
\[
\max_{h\in[H]}
\frac{\lambda_{\beta,h}B_\theta(1+B_\theta)}
{\rho_{\beta,h}^{\rm obs}}
\le c_{\rm loc}
\]
for the constant in Lemma~\ref{lem:beta_likelihood_event}. Run the same
recursion with \(m_\beta(z;\widehat\vartheta_{\beta,h}^{\rm reg})\),
\(\widehat\Gamma_{\beta,r,h}^{\rm reg}\), and
\[
\widehat\Gamma_{\beta,h}^{\rm reg}(z)
=
\widehat\Gamma_{\beta,r,h}^{\rm reg}(z)+\widehat\Gamma_{p,h}(z),
\]
and denote the returned value functions, transition coefficients, and policy by
\(\widehat V_h^{\rm BR,reg}\), \(\widehat\beta_h^{\rm BR,reg}\), and
\(\widehat\pi^{\rm BR,reg}\). Then there is an event
\(\mathcal E_\beta^{\rm reg}\) such that
\[
\P\{(\mathcal E_\beta^{\rm reg})^c\cap\mathcal N_\beta\}\le\xi,
\]
and, on \(\mathcal E_\beta^{\rm reg}\cap\mathcal N_\beta\),
\[
\operatorname{SubOpt}(\widehat\pi^{\rm BR,reg};x)
\le
2\sum_{h=1}^H
\E_{\pi^*}\!\left[
\widehat\Gamma_{\beta,h}^{\rm reg}(z_h)
\mid x_1=x
\right].
\]
If additionally \(\lambda_{\min}(\Lambda_h)\ge\rho_p>0\) for every \(h\),
then
\[
\begin{aligned}
\operatorname{SubOpt}(\widehat\pi^{\rm BR,reg};x)
&\le
\widetilde O\!\left(
\sqrt{\frac{d_\beta H^2}{\mathcal I_{\beta,\rm obs}}}
\right)
+
O\!\left(
H\max_{h\in[H]}
\frac{\lambda_{\beta,h}B_\theta}
{\rho_{\beta,h}^{\rm obs}}
\right)
\\[-2pt]
&\quad+
\widetilde O\!\left(
\sqrt{
\frac{(d_p+d_\beta)^2H^4}
{(n+N)\rho_p}
}
\right).
\end{aligned}
\]
This conclusion holds unconditionally with probability at least
\(1-\xi-\P(\mathcal N_\beta^c)-\P(\mathcal{C}^c)\). Under the observation-count condition and
confidence retuning stated in Theorem~\ref{thm:beta_step_level_reward}, this
probability is at least \(1-\xi\).
\end{corollary}

\begin{proof}
Work on \(\mathcal N_\beta\cap\mathcal L_\beta\). The localization part of
Lemma~\ref{lem:beta_likelihood_event} places the regularized estimator in
\(U_h\subset{\rm int}(\Psi_\beta)\) before the equality-form score equation is
used. Fix \(h\), let
\[
\Delta_h^{\rm reg}
=
\widehat\vartheta_{\beta,h}^{\rm reg}-\vartheta_h^*,
\qquad
\mathcal J_{\beta,h}^{\lambda}
=
\mathcal J_{\beta,h}^{\rm obs}+\lambda_{\beta,h}P_\beta,
\]
and use \(g_{\beta,h}\) from the proof of
Theorem~\ref{thm:beta_step_level_reward}. Define
\[
\overline{\mathcal J}_{\beta,h}^{\rm reg}
=
\int_0^1\left\{
\frac{1}{n_h^{\rm obs}}
\sum_{\tau\in\Isc_h^r}
\nabla_\vartheta^2\ell_\beta
(r_h^\tau,z_h^\tau;
\vartheta_h^*+t\Delta_h^{\rm reg})
+\lambda_{\beta,h}P_\beta
\right\}dt.
\]
Adding the same positive-semidefinite penalty to the two-sided Hessian bounds in
Lemma~\ref{lem:beta_likelihood_event} gives
\[
c\mathcal J_{\beta,h}^{\lambda}
\preceq
\overline{\mathcal J}_{\beta,h}^{\rm reg}
\preceq
C\mathcal J_{\beta,h}^{\lambda},
\qquad
c\mathcal J_{\beta,h}^{\lambda}
\preceq
\frac{1}{n_h^{\rm obs}}
\widehat{\mathcal J}_{\beta,h}^{\rm reg}
(\widehat\vartheta_{\beta,h}^{\rm reg})
\preceq
C\mathcal J_{\beta,h}^{\lambda}.
\]
The penalized score equation is therefore
\[
\Delta_h^{\rm reg}
=
-\left(\overline{\mathcal J}_{\beta,h}^{\rm reg}\right)^{-1}
\{g_{\beta,h}+\lambda_{\beta,h}P_\beta\vartheta_h^*\}.
\]
Because \(\mathcal J_{\beta,h}^{\lambda}\succeq
\mathcal J_{\beta,h}^{\rm obs}\),
\[
\|g_{\beta,h}\|_{(\mathcal J_{\beta,h}^{\lambda})^{-1}}
\le
\|g_{\beta,h}\|_{(\mathcal J_{\beta,h}^{\rm obs})^{-1}}
\le
C\sqrt{\frac{D_\beta}{n_h^{\rm obs}}}.
\]
The same mean-value derivative-ratio argument as in the unregularized proof,
followed by Cauchy--Schwarz in the penalized information norm, bounds the
stochastic part of the mean error by
\[
C\sqrt{D_\beta}
\left\|
a_\beta(z;\widehat\vartheta_{\beta,h}^{\rm reg})
\right\|_{
\widehat{\mathcal J}_{\beta,h}^{\rm reg}
(\widehat\vartheta_{\beta,h}^{\rm reg})^{-1}}.
\]
Since \(\|a_\beta(z;\vartheta)\|_2\le1/4\),
\(\|P_\beta\vartheta_h^*\|_2\le B_\theta\), and
\(\lambda_{\min}(\mathcal J_{\beta,h}^{\lambda})
\ge\rho_{\beta,h}^{\rm obs}\), the deterministic part is at most
\[
C\frac{\lambda_{\beta,h}B_\theta}
{\rho_{\beta,h}^{\rm obs}}.
\]
Increasing \(C_\beta\) by a fixed factor proves
\[
\left|
m_\beta(z;\widehat\vartheta_{\beta,h}^{\rm reg})
-m_\beta(z;\vartheta_h^*)
\right|
\le
\widehat\Gamma_{\beta,r,h}^{\rm reg}(z).
\]
The scalar \(b_{\beta,h}\) is fixed for a prespecified penalty, so it does not
alter the covering order in Lemma~\ref{lem:beta_value_cover}. Using
\(M_{\beta,h}=\alpha_\beta^2
\widehat{\mathcal J}_{\beta,h}^{\rm reg}
(\widehat\vartheta_{\beta,h}^{\rm reg})^{-1}\), that lemma and the same
unconditional transition event as in Theorem~\ref{thm:beta_step_level_reward}
give transition concentration for \(\widehat V_h^{\rm BR,reg}\). Componentwise
pessimism proves the first bound. Summing the stochastic reward, deterministic
penalty-bias, and transition terms gives the displayed rate. Taking
\(\mathcal E_\beta^{\rm reg}\) as the intersection of
\(\mathcal L_\beta\) and that transition event gives
\(\P\{(\mathcal E_\beta^{\rm reg})^c\cap\mathcal N_\beta\}\le\xi\).
\end{proof}

Theorem~\ref{thm:beta_step_level_reward} and
Corollary~\ref{cor:beta_step_level_reward_regularized} treat the reward penalty
as absent or prespecified, respectively, and both retain the observed beta
information condition \(\rho_{\beta,h}^{\rm obs}>0\). The no-\(\rho\) result in
Section~\ref{sec:relax} is a distinct complete-observation canonical-GLM
variant. Cross-validation selection of the reward penalty and pessimism
multiplier in the empirical sections is practical tuning and is not part of
these finite-sample claims.

If the complete-data beta Fisher information
\[
\mathcal J_{\beta,h}
=
\E_{\pi^b}\!\left[
\nabla_\vartheta^2
\ell_\beta(r_h,z_h;\vartheta_h^*)
\right]
\]
has minimum eigenvalue at least \(\rho_\beta>0\), the same calculation under missing at random as
in Lemma~\ref{lem:observed_sample_information} gives
\[
\rho_{\beta,h}^{\rm obs}
\ge
\frac{\bar p}{q_h}\rho_\beta.
\]
Together with \(n_h^{\rm obs}\asymp nq_h\), this implies
\(\mathcal I_{\beta,\rm obs}\gtrsim n\bar p\rho_\beta\).

\subsection{Relaxing the Reward-Curvature Assumption}
\label{sec:relax}
Given a regularization parameter $0<\lambda'=\lambda'_n\leq 1$  and a compact convex parameter space \(\Theta\subseteq\R^{d_r}\) whose interior contains the true and regularized population reward parameters, we define \[\mathcal{L}_{h,\lambda'}(\theta)\coloneqq\frac{1}{n}\sum_{\tau=1}^{n}\big(-r_h^\tau\langle\phi_r(x_h^\tau,a_h^\tau),\theta\rangle+G(\langle\phi_r(x_h^\tau,a_h^\tau),\theta\rangle)+\lambda'\|\theta\|_2^2\big),\]
and a new estimator for $\theta_h^*$ is defined as
\[
\widetilde \theta_{h,\lambda'}\coloneqq\argmin{\theta \in \Theta}\mathcal{L}_{h,\lambda'}(\theta).
\]
This is a reward-regularized estimator and is not the same object as the unregularized \(\widetilde\theta_h\) used in the main algorithm definition \eqref{est:theta}. The purpose of this variant is to replace the reward-curvature lower-eigenvalue condition by explicit regularization and obtain a valid coverage-adaptive, data-dependent reward radius without explicit dependence on \(\rho\). The radius is not guaranteed to vanish uniformly in reward-feature directions that remain unsupported; a vanishing uniform rate requires an additional coverage condition.
We also define
\[
   l_{h,\lambda'}(\theta)\coloneqq-r_h\langle\phi_r(x_h,a_h),\theta\rangle+G(\langle\phi_r(x_h,a_h),\theta\rangle)+\lambda'\|\theta\|_2^2,
   \] 
 \[
\theta^*_{h,\lambda'}\coloneqq\argmin{\theta \in \Theta}\ \E_{\pi^b} l_{h,\lambda'}(\theta),
\]
\[
\Sigma_{h,\lambda'}(\theta)\coloneqq\nabla^2[\E_{\pi^b} l_{h,\lambda'}(\theta)],
\] and \[
\widetilde \Sigma_{h,\lambda'}(\theta)\coloneqq n\nabla^2 \mathcal{L}_{h,\lambda'}(\theta)=\sum_{\tau=1}^n \dot{g}(\langle \phi_r(x_h^\tau,a_h^\tau),\theta\rangle)\phi_r(x_h^\tau,a_h^\tau)\phi_r(x_h^\tau,a_h^\tau)^\top+2n\lambda'\bI_{d_r}.
\]
For later use, define the population and empirical reward-feature Gram matrices
\[
\Lambda_{r,h}
=\E_{\pi^b}\!\left[\phi_r(x_h,a_h)\phi_r(x_h,a_h)^\top\right],
\qquad
\widetilde\Lambda_{r,h}
=\sum_{\tau=1}^n
\phi_r(x_h^\tau,a_h^\tau)\phi_r(x_h^\tau,a_h^\tau)^\top.
\]
The auxiliary reward-curvature lemmas are collected in Section~\ref{sec:technical_lemmas}.
Run the complete-observation recursion in Algorithm~\ref{alg:grasp_reward_observation} with $\widetilde\theta_{h,\lambda'}$ and the regularized reward radius below in place of their unregularized counterparts, and denote the resulting objects by $\widetilde Q_{h,\lambda'}$, $\widetilde V_{h,\lambda'}$, and $\widetilde\pi_{h,\lambda'}$. We define \[
\widetilde \Gamma_{h,\lambda'}(x,a)\coloneqq\widetilde \Gamma_{r,h,\lambda'}(x,a)+\widetilde \Gamma_{p,h}(x,a)
\]
with $\widetilde \Gamma_{r, h,\lambda'}$ equal to the right side of \eqref{eq:g_bound}.
For a continuation value \(V\), the corresponding empirical Bellman operator is
\[
\begin{aligned}
(\widetilde\B_{h,\lambda'}V)(x,a)
&:=g\!\left\{\langle\phi_r(x,a),
\widetilde\theta_{h,\lambda'}\rangle\right\}
+\langle\phi_p(x,a),\widetilde\beta_h(V)\rangle,\\
\widetilde\beta_h(V)
&:=(\widetilde\Lambda_h+\lambda\bI_{d_p})^{-1}
\sum_{\tau=1}^n
\phi_p(x_h^\tau,a_h^\tau)V(x_{h+1}^\tau).
\end{aligned}
\]
\begin{theorem}[Suboptimality for \texorpdfstring{GRASP under complete reward observation}{GRASP under complete reward observation} without Assumption \ref{assump:sigma}]
\label{thm:complete_reward_observation_relaxed_curvature}
Under Assumptions \ref{assump:noise} and \ref{assump:link}, suppose additionally that the conditional reward mean satisfies $0\le g(\langle\phi_r(x,a),\theta_h^*\rangle)\le1$ for every $h,x,a$. Set $\lambda=1$, $\alpha_p=c_p\left(d_p+d_r\right) H \sqrt{\zeta}$, where $\zeta=\log \left(4\left(d_r+d_p\right) H n / \xi\right)$, $c_p>0$ is a sufficiently large numerical constant and $\xi \in (0,1)$ is the confidence parameter; in the reward radius \eqref{eq:g_bound}, use confidence level $\xi/2$. Assume also that $n\lambda'\ge C\{d_r+\log(H/\xi)\}$ for a sufficiently large constant $C$. Then $\{\widetilde \Gamma_{h,\lambda'}\}_{h=1}^H$ defined above is a $\xi$-uncertainty quantifier for $\{\widetilde{\B}_{h,\lambda'}\widetilde V_{h+1,\lambda'}\}_{h=1}^H$. For any $x \in \Ssc$ and $n$ large enough, $\widetilde \pi_{\lambda'} = \{ \widetilde \pi_{h,\lambda'} \}_{h=1}^H$ satisfies
 $$\operatorname{SubOpt}\big(\widetilde \pi_{\lambda'} ;x \big) \leq 2 \sum_{h=1}^H\E_{\pi^{*}}\Bigl[ \widetilde \Gamma_{h,\lambda'}(x_h, a_h)\mid x_1=x\Bigr]$$
with probability at least $1-\xi$.
\end{theorem}
\begin{proof}
Invoke Lemma~\ref{lem:g1} at confidence level $\xi/2$ for the reward event. It
remains to check that transition concentration covers the value function
generated by the full regularized reward radius. Write
\[
a_{1,n}:=C_2\sqrt{d_r+\log(H/\xi)},
\quad
a_{2,n}:=C_2L\sqrt{\lambda'}.
\]
The regularized value class consists of two-sided clipped maxima of
\[
\begin{aligned}
g(\langle\phi_r,\theta\rangle)
&-\dot g(\langle\phi_r,\theta\rangle)
  \sqrt{\phi_r^\top M_1\phi_r}
-\sqrt{\phi_r^\top M_2\phi_r}
+\langle\phi_p,\beta\rangle
-\sqrt{\phi_p^\top M_p\phi_p},
\end{aligned}
\]
where $\theta\in\Theta$, $\|\beta\|_2\le B:=H\sqrt{n/\lambda}$, and the
three positive-semidefinite matrices range over balls containing
\[
M_1=a_{1,n}^2
\widetilde\Sigma_{h,\lambda'}(\widetilde\theta_{h,\lambda'})^{-1},
\quad
M_2=a_{2,n}^2
\left(2\lambda'\bI_{d_r}+n^{-1}\widetilde\Lambda_{r,h}\right)^{-1},
\quad
M_p=\alpha_p^2(\widetilde\Lambda_h+\lambda \bI_{d_p})^{-1}.
\]
Set
\[
R_{1,{\rm op}}=\frac{a_{1,n}^2}{2n\lambda'},
\quad
R_{2,{\rm op}}=\frac{a_{2,n}^2}{2\lambda'},
\quad
R_{p,{\rm op}}=\frac{\alpha_p^2}{\lambda},
\]
and take the corresponding Frobenius radii
\[
R_{1,F}=\sqrt{d_r}R_{1,{\rm op}},
\qquad
R_{2,F}=\sqrt{d_r}R_{2,{\rm op}},
\qquad
R_{p,F}=\sqrt{d_p}R_{p,{\rm op}}.
\]
These balls contain the three algorithmic matrices because
$\|A\|_{\rm F}\le\sqrt d\|A\|_{\rm op}$ for a $d\times d$ matrix.

Let $R_\Theta:=\sup_{\theta\in\Theta}\|\theta\|_2$. The maximum over actions
and the two-sided clipping map are one-Lipschitz. For two parameter tuples,
the mean-value theorem and $0<\dot g,|\ddot g|\le L$ give
\[
\begin{aligned}
\|V-V'\|_\infty
\le{}&
L(1+\sqrt{R_{1,{\rm op}}})\|\theta-\theta'\|_2
+\|\beta-\beta'\|_2\\
&+L\|M_1-M_1'\|_{\rm F}^{1/2}
+\|M_2-M_2'\|_{\rm F}^{1/2}
+\|M_p-M_p'\|_{\rm F}^{1/2}.
\end{aligned}
\]
Use Euclidean accuracies
$\epsilon/[5L(1+\sqrt{R_{1,{\rm op}}})]$ and $\epsilon/5$ for
$\theta$ and $\beta$, and Frobenius accuracies
$(\epsilon/(5L))^2,(\epsilon/5)^2,(\epsilon/5)^2$ for $M_1,M_2,M_p$.
Lemma~\ref{lem:cover} in the corresponding Euclidean matrix spaces gives
\[
\begin{aligned}
\log\Nsc_h^{\rm reg}(\epsilon)
&\le
d_r\log\!\left(1+
\frac{10L(1+\sqrt{R_{1,{\rm op}}})R_\Theta}{\epsilon}\right)
+d_p\log\!\left(1+\frac{10B}{\epsilon}\right)\\
&\quad+d_r^2\log\!\left(1+
\frac{50L^2R_{1,F}}{\epsilon^2}\right)
+d_r^2\log\!\left(1+
\frac{50R_{2,F}}{\epsilon^2}\right)
+d_p^2\log\!\left(1+
\frac{50R_{p,F}}{\epsilon^2}\right).
\end{aligned}
\]
With $\epsilon=H\sqrt{d_p/n}$, compactness of $\Theta$, and
$n\lambda'\ge C\{d_r+\log(H/\xi)\}$, substitution of the radii shows
\[
\log\Nsc_h^{\rm reg}(\epsilon)
\le C'(d_r+d_p)^2\{\log(c_p+1)+\zeta\}.
\]
The complete-observation finite-net argument, invoked with total failure
probability $\xi/2$, now gives transition confidence at
$\widetilde V_{h+1,\lambda'}$. Combining it with the reward event makes
$\widetilde\Gamma_{h,\lambda'}$ an uncertainty quantifier for
$\widetilde\B_{h,\lambda'}$ with probability at least $1-\xi$. The added
conditional-mean assumption ensures that the true Bellman backup belongs to
$[0,H-h+1]$, so the clipping proof in
\eqref{eq:partial-pessimism-residual} yields the result.
\end{proof}

\subsection{Online Regret Guarantee}
\label{sec:proof_regret_online}
\begin{proof}[Proof of Theorem~\ref{thm:regret_online}]
Let
\[
b_{h,s}(x,a)
:=
\gamma_r\|\phi_r(x,a)\|_{\Lambda_{r,h,s}^{-1}}
+\gamma_p\|\phi_p(x,a)\|_{\Lambda_{p,h,s}^{-1}}.
\]
On the simultaneous event in Lemmas~\ref{lem:theta_err_online} and
\ref{lem:beta_err_online}, the estimated, pre-bonus Bellman backup differs
from $(\B_h\bar V_{h+1,s})(x,a)$ by at most $b_{h,s}(x,a)$. Since the
algorithm adds the same bonus and clips to $[0,H-h+1]$, Lemma~\ref{lem:ucb}
and the clipping calculation give, simultaneously for every $h,s,x,a$,
\begin{equation}\label{eq:online_bellman_residual}
0\le
\bar Q_{h,s}(x,a)-(\B_h\bar V_{h+1,s})(x,a)
\le
c_{h,s}(x,a),
\qquad
c_{h,s}(x,a):=\min\{H-h+1,2b_{h,s}(x,a)\}.
\end{equation}
Indeed, before clipping the difference from the true backup is between zero
and $2b_{h,s}$, and the true backup belongs to the clipping interval.

The first policy is based on the deterministic initialization rather than a
data-based confidence set, so its suboptimality is bounded by $H$. Fix now
$t\ge2$ and write $\pi_t=\widehat\pi_{\cdot,t}$. Conditional on
$\Fsc_{t-1}$, this policy and $\bar V_{\cdot,t-1}$ are fixed. Define
\[
\delta_{h,t}(x):=\bar V_{h,t-1}(x)-V_h^{\pi_t}(x).
\]
Optimism implies
$V_1^*(x)-V_1^{\pi_t}(x)\le\delta_{1,t}(x)$. Because $\pi_t$ is greedy
with respect to $\bar Q_{\cdot,t-1}$, applying
\eqref{eq:online_bellman_residual} at the visited state-action pair gives
\[
\delta_{h,t}(x_{h,t})
\le
c_{h,t-1}(x_{h,t},a_{h,t})
+\E\{\delta_{h+1,t}(x_{h+1})
\mid\Fsc_{t-1},x_{h,t},a_{h,t}\}.
\]
Let
\[
m_{h,t}
:=
\E\{\delta_{h+1,t}(x_{h+1})
\mid\Fsc_{t-1},x_{h,t},a_{h,t}\}
-\delta_{h+1,t}(x_{h+1,t}).
\]
For the analysis, let
\[
\mathcal G_{h,t}
:=
\sigma\!\left(\Fsc_{t-1},
x_{1,t},a_{1,t},\ldots,x_{h,t},a_{h,t}\right),
\]
where the next state $x_{h+1,t}$ is revealed after $\mathcal G_{h,t}$. Then
$\E(m_{h,t}\mid\mathcal G_{h,t})=0$, so the $m_{h,t}$ form martingale differences
under the chronological, stepwise filtration. Moreover, optimism and clipping imply
$0\le\delta_{h+1,t}\le H-h$, and hence $|m_{h,t}|\le H-h$. Telescoping the
preceding recursion over $h$ and then summing over $t\ge2$ yields
\begin{equation}\label{eq:online_regret_decomposition}
\mathcal R_T^{\rm ps}(x)
\le
H+
\sum_{t=2}^T\sum_{h=1}^H
c_{h,t-1}(x_{h,t},a_{h,t})
+\sum_{t=2}^T\sum_{h=1}^H m_{h,t}.
\end{equation}
Because
$\sum_{t,h}(H-h)^2\le TH^3/3$, Azuma's inequality gives
\[
\sum_{t=2}^T\sum_{h=1}^H m_{h,t}
\le \sqrt{2TH^3\log(6/p_0)}
\]
with probability at least $1-p_0/3$; the displayed constant is a
conservative simplification of the direct bound.

Finally, $c_{h,t-1}\le2b_{h,t-1}$ and the elliptical-potential inequality
gives, for $j\in\{r,p\}$,
\[
\sum_{t=2}^T
\|\phi_j(x_{h,t},a_{h,t})\|_{\Lambda_{j,h,t-1}^{-1}}
\le
\sqrt{2T d_j\log(1+T/d_j)}.
\]
Substitution in \eqref{eq:online_regret_decomposition}, followed by summing
over $h$, proves the first display in Theorem~\ref{thm:regret_online}.
The reward- and transition-confidence events have joint probability at least
$1-2p_0/3$; combining them with the martingale event completes the stated
$1-p_0$ probability bound.
\end{proof}

\section{Technical Lemmas}
\label{sec:technical_lemmas}
We collect all supporting lemmas for the supplement in this section, ordered to follow their first use in the proofs, with short helper lemmas placed immediately before the lemmas that invoke them.

\begin{lemma}
\label{lem:sigma}
If $\lambda_{\min}\big(\Sigma_{h}(\theta_h^{*}) \big) \geq \rho >0$, then for $\xi \in (0,1)$, with sufficiently large  $n$, we have
$$\lambda_{\min} \big(\frac{1}{n}\widetilde \Sigma_{h}(\theta_h^{*}) \big) \geq \frac{3 \rho}{4}$$
with probability at least $1 - \xi$. 
\end{lemma}
\begin{proof}
By the matrix Bernstein inequality \citep{tropp2015introduction} and $\| \phi_r(x,a)\|_2 \leq 1$, we have 
\begin{equation}\|\frac{1}{n}\widetilde \Sigma_{h}(\theta_h^{*}) -\Sigma_{h}(\theta_h^{*}) \| \leq C \sqrt{\log(  d_r/\xi) / n}\label{eq:p_matrix_bernstein}\end{equation}
with probability at least $1- \xi/2$ for a constant $C > 0$ depending only on the fixed upper derivative bound $L$. Hence if $n$ is sufficiently large, we have
   $$\lambda_{\min} \big(\frac{1}{n}\widetilde \Sigma_{h}(\theta_h^{*}) \big) \geq \lambda_{\min} \big( \Sigma_{h}(\theta_h^{*}) \big) - \|\frac{1}{n}\widetilde \Sigma_{h}(\theta_h^{*}) -\Sigma_{h}(\theta_h^{*}) \| \geq \rho - \frac{ \rho}{4} = \frac{3 \rho}{4}.$$
 
\end{proof}
\begin{lemma}[Reward Score Concentration]
\label{lem:gradient_norm_bound}
Suppose that Assumptions~\ref{assump:noise} and
\ref{assump:link} hold, and that, for some constant
$C_{\mathrm{var}}>0$ independent of $n,d_r,H$, and $\xi$,
\begin{equation}\label{eq:cond for reward score concentartion}
\E_{\pi^b}\!\left[
\mathrm{Var} (r_h\mid x_h,a_h)
\phi_r(x_h,a_h)\phi_r(x_h,a_h)^\top
\right]
\preceq
C_{\mathrm{var}}\Sigma_h(\theta_h^*)
\qquad\text{for every }h\in[H].
\end{equation}
Suppose further that
\[
\lambda_{\min}\{\Sigma_h(\theta_h^*)\}\geq\rho
\qquad\text{for every }h\in[H].
\]
There exists a constant $C>0$, depending only on the fixed
reward-noise constant, $C_{\mathrm{var}}$, the parameter radius, and
the local upper bound of $\dot g$, such that, whenever
\[
n\rho
\geq
C\left\{
d_r+\log\frac{16Hd_r}{\xi}
\right\},
\]
with probability at least $1-\xi$, simultaneously for every
$h\in[H]$,
\[
\frac12\Sigma_h(\theta_h^*)
\preceq
\frac1n\widetilde\Sigma_h(\theta_h^*)
\preceq
\frac32\Sigma_h(\theta_h^*),
\]
and
\[
\left\|
\nabla\mathcal L_h(\theta_h^*)
\right\|_{
\left\{
n^{-1}\widetilde\Sigma_h(\theta_h^*)
\right\}^{-1}
}^2
\leq
C\frac{
d_r+\log(16H/\xi)
}{n}.
\]
\end{lemma}
\begin{remark}[Interpretation of \(C_{\mathrm{var}}\)]
The constant \(C_{\mathrm{var}}\) in \eqref{eq:cond for reward score concentartion}  is determined only by the behavior of
the reward distribution and \(\dot g\) at the true predictors
\(\langle\phi_r(x,a),\theta_h^*\rangle\). Under a uniform conditional
sub-Gaussian bound with variance proxy \(\sigma^2\), together with
\[
\inf_{h,x,a}
\dot g\!\left(
\langle\phi_r(x,a),\theta_h^*\rangle
\right)
\geq \underline g_*>0,
\]
one may take \(C_{\mathrm{var}}=\sigma^2/\underline g_*\). This is only
a generic sufficient choice. When the conditional reward distribution
has additional structure, \(C_{\mathrm{var}}\) may be chosen more
sharply; for example, if its conditional variance is bounded by
\(c_0\dot g(\langle\phi_r(x,a),\theta_h^*\rangle)\), then one may take
\(C_{\mathrm{var}}=c_0\). Thus, no global lower bound on \(\dot g\) over
the entire parameter space is required.
\end{remark}
\begin{proof}
Fix $h\in[H]$ and define
\[
A_{h,\tau}
:=
\Sigma_h(\theta_h^*)^{-1/2}
\dot g\!\left(
\left\langle
\phi_{r,h}^\tau,\theta_h^*
\right\rangle
\right)
\phi_{r,h}^\tau(\phi_{r,h}^\tau)^\top
\Sigma_h(\theta_h^*)^{-1/2}.
\]
Then $A_{h,\tau}\succeq0$ and $\E A_{h,\tau}=\bI_{d_r}$. Bounded
features and the local upper bound of $\dot g$ imply
\[
\|A_{h,\tau}\|_{\rm op}
\leq
\frac{C}{\rho}.
\]
Since a positive semidefinite matrix satisfying
$A_{h,\tau}\preceq C\rho^{-1}\bI_{d_r}$ also satisfies
\[
A_{h,\tau}^2
\preceq
\frac{C}{\rho}A_{h,\tau},
\]
we have
\[
\left\|
\sum_{\tau=1}^n
\E\left[
\left(
A_{h,\tau}-\bI_{d_r}
\right)^2
\right]
\right\|_{\rm op}
\leq
\frac{Cn}{\rho}.
\]
Matrix Bernstein's inequality and a union bound over $h\in[H]$ give
\[
\left\|
\Sigma_h(\theta_h^*)^{-1/2}
\left\{
\frac1n\widetilde\Sigma_h(\theta_h^*)
-
\Sigma_h(\theta_h^*)
\right\}
\Sigma_h(\theta_h^*)^{-1/2}
\right\|_{\rm op}
\leq\frac12
\]
simultaneously for every $h\in[H]$, proving the asserted Loewner
comparison.

Next,
\[
\nabla\mathcal L_h(\theta_h^*)
=
\frac1n\sum_{\tau=1}^n
\left\{
g\!\left(
\left\langle
\phi_{r,h}^\tau,\theta_h^*
\right\rangle
\right)
-r_h^\tau
\right\}
\phi_{r,h}^\tau.
\]
The summands are centered. For every $v\in\mathbb S^{d_r-1}$, the
score--information comparison condition gives
\[
\mathrm{Var} \!\left[
\left\{
g\!\left(
\left\langle
\phi_{r,h},\theta_h^*
\right\rangle
\right)-r_h
\right\}
v^\top\Sigma_h(\theta_h^*)^{-1/2}\phi_{r,h}
\right]
\leq C_{\mathrm{var}}.
\]
Moreover, Assumption~\ref{assump:noise}, bounded features, and
$\Sigma_h(\theta_h^*)\succeq\rho\bI_{d_r}$ imply that the corresponding
sub-exponential norm is at most $C/\sqrt{\rho}$. Bernstein's inequality
therefore yields
\[
\begin{aligned}
&
\left|
v^\top
\Sigma_h(\theta_h^*)^{-1/2}
\nabla\mathcal L_h(\theta_h^*)
\right|
\\
&\qquad\leq
C\left\{
\sqrt{\frac{d_r+\log(16H/\xi)}{n}}
+
\frac{d_r+\log(16H/\xi)}{n\sqrt{\rho}}
\right\}
\end{aligned}
\]
simultaneously over a $1/2$-net of $\mathbb S^{d_r-1}$ and over
$h\in[H]$. The standard net argument and
$n\rho\geq C\{d_r+\log(16H/\xi)\}$ consequently give
\[
\left\|
\nabla\mathcal L_h(\theta_h^*)
\right\|_{\Sigma_h(\theta_h^*)^{-1}}
\leq
C\sqrt{\frac{d_r+\log(16H/\xi)}{n}}.
\]
Finally,
\[
\frac1n\widetilde\Sigma_h(\theta_h^*)
\succeq
\frac12\Sigma_h(\theta_h^*)
\]
implies
\[
\left\{
\frac1n\widetilde\Sigma_h(\theta_h^*)
\right\}^{-1}
\preceq
2\Sigma_h(\theta_h^*)^{-1},
\]
which proves the score bound.
\end{proof}
\begin{lemma}[Reward-Estimator Error]
\label{lem:init}
Suppose that the conditions of Lemma~\ref{lem:gradient_norm_bound} hold. Then
there exists a constant $C>0$, independent of
$n,d_r,H$, and $\xi$, such that, whenever
\[
n\rho
\geq
C\left\{
d_r+\log\frac{16H}{\xi}
\right\},
\]
with probability at least $1-\xi$, simultaneously for every
$h\in[H]$,
\[
\left\|
\widetilde\theta_h-\theta_h^*
\right\|_{
n^{-1}\widetilde\Sigma_h(\theta_h^*)
}^2
\leq
C\frac{d_r+\log(16H/\xi)}{n},
\]
\[
\left\|
\widetilde\theta_h-\theta_h^*
\right\|_{
n^{-1}\widetilde\Sigma_h(\widetilde\theta_h)
}^2
\leq
C\frac{d_r+\log(16H/\xi)}{n},
\]
and
\[
\left\|
\widetilde\theta_h-\theta_h^*
\right\|_2^2
\leq
C\frac{d_r+\log(16H/\xi)}{n\rho}.
\]
\end{lemma}

\begin{proof}
Work on the simultaneous event in Lemma~\ref{lem:gradient_norm_bound}, and let
$C_0>0$ satisfy
\[
\left\|
\nabla\mathcal L_h(\theta_h^*)
\right\|_{
\left\{
n^{-1}\widetilde\Sigma_h(\theta_h^*)
\right\}^{-1}
}
\leq
C_0
\sqrt{\frac{d_r+\log(16H/\xi)}{n}}
\]
for every $h\in[H]$.

Consider any $\Delta\in \R^{d_r}$ satisfying
\[
\|\Delta\|_{
n^{-1}\widetilde\Sigma_h(\theta_h^*)
}
=
8C_0
\sqrt{\frac{d_r+\log(16H/\xi)}{n}}.
\]
For sufficiently large \(C\) in the sample-size condition,
\[
\sup_{\substack{
\theta\in\mathbb R^{d_r}:
\|\theta-\theta_h^*\|_2
\leq
8\sqrt{2}C_0\sqrt{
\frac{d_r+\log(16H/\xi)}{n\rho}
}\\
(x,a)\in\mathcal S\times\mathcal A
}}
\left|
\frac{
\ddot g\!\left(
\langle\phi_r(x,a),\theta\rangle
\right)
}{
\dot g\!\left(
\langle\phi_r(x,a),\theta\rangle
\right)
}
\right|
\,
8\sqrt{2}C_0\sqrt{
\frac{d_r+\log(16H/\xi)}{n\rho}
}
\leq \log 2.
\]
Indeed, every \(\Delta\) on the information-norm boundary considered
above satisfies
\[
\|\Delta\|_2
\leq
8\sqrt{2}C_0\sqrt{
\frac{d_r+\log(16H/\xi)}{n\rho}
},
\]
so that
\(\theta_h^*+t\Delta\) belongs to the preceding Euclidean ball for every
\(t\in[0,1]\). Therefore,
\[
\begin{aligned}
&
\left|
\log\dot g\!\left(
\left\langle
\phi_{r,h}^\tau,\theta_h^*+t\Delta
\right\rangle
\right)
-
\log\dot g\!\left(
\left\langle
\phi_{r,h}^\tau,\theta_h^*
\right\rangle
\right)
\right|
\\
&\qquad\leq
\sup_{\substack{
\theta\in\mathbb R^{d_r}:
\|\theta-\theta_h^*\|_2
\leq
8\sqrt{2}C_0\sqrt{
\frac{d_r+\log(16H/\xi)}{n\rho}
}\\
(x,a)\in\mathcal S\times\mathcal A
}}
\left|
\frac{
\ddot g\!\left(
\langle\phi_r(x,a),\theta\rangle
\right)
}{
\dot g\!\left(
\langle\phi_r(x,a),\theta\rangle
\right)
}
\right|
\|\Delta\|_2
\leq \log 2
\end{aligned}
\] 
The preceding logarithmic bound implies that, for every
$t\in[0,1]$ and $\tau\in[n]$,

\[
\frac12
\dot g\!\left(
\left\langle
\phi_{r,h}^\tau,\theta_h^*
\right\rangle
\right)
\leq
\dot g\!\left(
\left\langle
\phi_{r,h}^\tau,\theta_h^*+t\Delta
\right\rangle
\right)
\leq
2
\dot g\!\left(
\left\langle
\phi_{r,h}^\tau,\theta_h^*
\right\rangle
\right).
\]
Multiplying by
$\phi_{r,h}^\tau(\phi_{r,h}^\tau)^\top$ and summing over
$\tau\in[n]$ yields
\begin{equation}\label{eq:matrix comparison bound}
\frac12
\frac1n\widetilde\Sigma_h(\theta_h^*)
\preceq
\frac1n\widetilde\Sigma_h(\theta_h^*+t\Delta)
\preceq
2
\frac1n\widetilde\Sigma_h(\theta_h^*),
\qquad t\in[0,1].
\end{equation}
Consequently,
\[
\begin{aligned}
&\int_0^1
(1-t)
\Delta^\top
\left\{
\frac1n
\widetilde\Sigma_h(\theta_h^*+t\Delta)
\right\}
\Delta\,dt
\\
&\qquad\geq
\frac12
\int_0^1(1-t)\,dt\,
\|\Delta\|_{
n^{-1}\widetilde\Sigma_h(\theta_h^*)
}^2
=
\frac14
\|\Delta\|_{
n^{-1}\widetilde\Sigma_h(\theta_h^*)
}^2.
\end{aligned}
\]
Taylor's integral formula gives
\[
\begin{aligned}
&
\mathcal L_h(\theta_h^*+\Delta)
-
\mathcal L_h(\theta_h^*)
\\
&=
\left\langle
\nabla\mathcal L_h(\theta_h^*),\Delta
\right\rangle
+
\int_0^1
(1-t)
\Delta^\top
\left\{
\frac1n
\widetilde\Sigma_h(\theta_h^*+t\Delta)
\right\}
\Delta\,dt
\\
&\geq
-
\left\|
\nabla\mathcal L_h(\theta_h^*)
\right\|_{
\left\{
n^{-1}\widetilde\Sigma_h(\theta_h^*)
\right\}^{-1}
}
\|\Delta\|_{
n^{-1}\widetilde\Sigma_h(\theta_h^*)
}
+
\frac14
\|\Delta\|_{
n^{-1}\widetilde\Sigma_h(\theta_h^*)
}^2
\\
&\geq
8C_0^2
\frac{d_r+\log(16H/\xi)}{n}
>0.
\end{aligned}
\]
The preceding strict boundary inequality
implies
\[
\left\|
\widetilde\theta_h-\theta_h^*
\right\|_{
n^{-1}\widetilde\Sigma_h(\theta_h^*)
}
\leq
8C_0
\sqrt{\frac{d_r+\log(16H/\xi)}{n}}.
\]

Therefore, combining \eqref{eq:matrix comparison bound}, we obtain
\begin{equation}\label{eq:sigma}
\frac12
\frac1n\widetilde\Sigma_h(\theta_h^*)
\preceq
\frac1n\widetilde\Sigma_h
\left(
\widetilde\theta_h
\right)
\preceq
2
\frac1n\widetilde\Sigma_h(\theta_h^*).
\end{equation}
Hence\[
\left\|
\widetilde\theta_h-\theta_h^*
\right\|_{
n^{-1}\widetilde\Sigma_h(\widetilde\theta_h)
}^2
\leq
2\left\|
\widetilde\theta_h-\theta_h^*
\right\|_{
n^{-1}\widetilde\Sigma_h(\theta_h^*)
}^2,
\]
Finally,
\[
\frac1n\widetilde\Sigma_h(\theta_h^*)
\succeq
\frac{\rho}{2}\bI_{d_r}
\]
gives
\[
\left\|
\widetilde\theta_h-\theta_h^*
\right\|_2^2
\leq
\frac{2}{\rho}
\left\|
\widetilde\theta_h-\theta_h^*
\right\|_{
n^{-1}\widetilde\Sigma_h(\theta_h^*)
}^2\le 128C_0^2
\frac{d_r+\log(16H/\xi)}{n\rho}
\]

completing the proof.
\end{proof}

\begin{lemma}[Reward prediction bound]
\label{lem:g}
Suppose that the conditions of Lemma~\ref{lem:init} hold. There exists
a constant $C>0$, independent of $n,d_r,H$, and $\xi$, such that, with
probability at least $1-\xi$, simultaneously for every $h\in[H]$ and
$(x,a)\in\Ssc\times\Asc$,
\[
\begin{aligned}
&
\left|
g\!\left(
\left\langle
\phi_r(x,a),\widetilde\theta_h
\right\rangle
\right)
-
g\!\left(
\left\langle
\phi_r(x,a),\theta_h^*
\right\rangle
\right)
\right|
\\
&\qquad\leq
C
\sqrt{
d_r+\log\frac{16H}{\xi}
}
\,
\dot g\!\left(
\left\langle
\phi_r(x,a),\widetilde\theta_h
\right\rangle
\right)
\left\|
\phi_r(x,a)
\right\|_{
\widetilde\Sigma_h(\widetilde\theta_h)^{-1}
}.
\end{aligned}
\]
\end{lemma}

\begin{proof}
Fix $h\in[H]$ and $(x,a)\in\Ssc\times\Asc$. By the mean-value theorem,
there exists $s\in[0,1]$ such that
\[
\begin{aligned}
&
\left|
g\!\left(
\left\langle
\phi_r(x,a),\widetilde\theta_h
\right\rangle
\right)
-
g\!\left(
\left\langle
\phi_r(x,a),\theta_h^*
\right\rangle
\right)
\right|
\\
&=
\dot g\!\left(
\left\langle
\phi_r(x,a),
\theta_h^*
+s(\widetilde\theta_h-\theta_h^*)
\right\rangle
\right)
\left|
\left\langle
\phi_r(x,a),
\widetilde\theta_h-\theta_h^*
\right\rangle
\right|.
\end{aligned}
\]
The Euclidean localization in Lemma~\ref{lem:init} and the local
Lipschitz property of $\log\dot g$ imply
\[
\dot g\!\left(
\left\langle
\phi_r(x,a),
\theta_h^*
+s(\widetilde\theta_h-\theta_h^*)
\right\rangle
\right)
\leq
2
\dot g\!\left(
\left\langle
\phi_r(x,a),\widetilde\theta_h
\right\rangle
\right).
\]
Cauchy--Schwarz in the empirical information norm and
Lemma~\ref{lem:init} therefore give
\[
\begin{aligned}
&
\left|
g\!\left(
\left\langle
\phi_r(x,a),\widetilde\theta_h
\right\rangle
\right)
-
g\!\left(
\left\langle
\phi_r(x,a),\theta_h^*
\right\rangle
\right)
\right|
\\
&\leq
C
\dot g\!\left(
\left\langle
\phi_r(x,a),\widetilde\theta_h
\right\rangle
\right)
\left\|
\phi_r(x,a)
\right\|_{
\left\{
n^{-1}\widetilde\Sigma_h(\widetilde\theta_h)
\right\}^{-1}
}
\left\|
\widetilde\theta_h-\theta_h^*
\right\|_{
n^{-1}\widetilde\Sigma_h(\widetilde\theta_h)
}
\\
&\leq
C
\dot g\!\left(
\left\langle
\phi_r(x,a),\widetilde\theta_h
\right\rangle
\right)
\sqrt{\frac{d_r+\log(16H/\xi)}{n}}
\left\|
\phi_r(x,a)
\right\|_{
\left\{
n^{-1}\widetilde\Sigma_h(\widetilde\theta_h)
\right\}^{-1}
}.
\end{aligned}
\]
Finally,
\[
\left\|
\phi_r(x,a)
\right\|_{
\left\{
n^{-1}\widetilde\Sigma_h(\widetilde\theta_h)
\right\}^{-1}
}
=
\sqrt n
\left\|
\phi_r(x,a)
\right\|_{
\widetilde\Sigma_h(\widetilde\theta_h)^{-1}
},
\]
which proves the claim.
\end{proof}

\begin{lemma}[Bounded Coefficients] For any function $V:\Ssc\rightarrow[0,V_{\max}]$, where $V_{\max}>0$ is a finite bound (and may equal the horizon-dependent bound $H$), the vector $\beta_h=\int_{\Ssc}V(x')\,\mu_h(dx')$ satisfies $\|\beta_h\|_2\leq V_{\max}\sqrt{d_p}$. Moreover,
$$ \|\widetilde{\beta}_{h}\|_2 \leq H \sqrt{n / \lambda}. $$
\label{lem:norm}
\end{lemma}
\begin{proof}
The normalization on the transition measure gives
\[
\|\beta_h\|_2
\leq V_{\max}\int_{\Ssc}\|\mu_h(dx')\|_2
\leq V_{\max}\sqrt{d_p}.
\]
For the empirical coefficient, let $\Phi_h$ have rows
$\phi_p(x_h^\tau,a_h^\tau)^\top$ and define
\[
v_h=\bigl(\widetilde V_{h+1}(x_{h+1}^1),\ldots,
\widetilde V_{h+1}(x_{h+1}^n)\bigr)^\top,
\qquad
\widetilde\Lambda_h=\Phi_h^\top\Phi_h.
\]
Then
$\|v_h\|_2\leq H\sqrt n$, and
\[
\begin{aligned}
\|\widetilde\beta_h\|_2
&\leq\lambda^{-1/2}
 \|\widetilde\beta_h\|_{\widetilde\Lambda_h+\lambda\bI_{d_p}}
 =\lambda^{-1/2}
 \|(\widetilde\Lambda_h+\lambda\bI_{d_p})^{-1/2}\Phi_h^\top v_h\|_2\\
&\leq\lambda^{-1/2}\|v_h\|_2
\leq H\sqrt{n/\lambda},
\end{aligned}
\]
where the penultimate inequality uses
$\|(\widetilde\Lambda_h+\lambda\bI_{d_p})^{-1/2}\Phi_h^\top\|_{\mathrm{op}}\leq1$.
\end{proof}

\begin{lemma}[Concentration of Self-Normalized Processes]
Let $\{\Fsc_t \}^\infty_{t=0}$ be a filtration and $\{\epsilon_t\}^\infty_{t=1}$ be an $\R$-valued stochastic process such that $\epsilon_t$ is $\Fsc_{t} $-measurable for all $t\geq 1$.
Moreover, suppose that conditioning on $\Fsc_{t-1}$, 
 $\epsilon_t $ is a  zero-mean and $\sigma$-sub-Gaussian random variable for all $t\geq 1$, that is,  
 $$
  \E[\epsilon_t \mid \Fsc_{t-1}]=0,\qquad \E\bigl[ \exp(\lambda \epsilon_t) \mid \Fsc_{t-1}\bigr]\leq \exp(\lambda^2\sigma^2/2) , \qquad \forall \lambda \in \R. 
$$
 Meanwhile, let $\{\phi_t\}_{t=1}^\infty$ be an $\R^d$-valued stochastic process such that  $\phi_t $  is $\Fsc_{t -1}$-measurable for all $ t\geq 1$. 
Also, let  $\bM_0 \in \R^{d\times d}$ be a  deterministic positive-definite matrix and 
$$
\bM_t = \bM_0 + \sum_{s=1}^t \phi_s\phi_s^\top
$$ for all $t\geq 1$. For all $\delta\in (0,1)$, it holds that
\begin{equation*}
\Big\| \sum_{s=1}^t \phi_s \epsilon_s \Big\|_{ \bM_t ^{-1}}^2 \leq 2\sigma^2\cdot  \log \Bigl( \frac{\det(\bM_t)^{1/2} \det(\bM_0)^{- 1/2}}{\delta} \Bigr)
\end{equation*}
for all $t\ge1$ with probability at least $1-\delta$.
\label{lem:concen_self_normalized}
\end{lemma}
\begin{proof}
See Theorem 1 of \cite{NIPS2011_e1d5be1c} for a detailed proof. 
\end{proof}

\begin{lemma}[Concentration of Self-Normalized Processes]  \label{lem:self_normalized_value_concentration}
Let $V: \mathcal{S} \rightarrow[0, H-1]$ be any fixed function. For any fixed $h \in[H]$ and any $\delta \in(0,1)$, we have
$$
\P_{\Dsc}\left( \Big \| \sum_{\tau=1}^{n}  \phi_p(x_h^{\tau},a_h^{\tau})  \epsilon_{h}^{\tau}  (V) \Big \|_{\Omega_{h}} > H\sqrt{2 \log (1 / \delta)+d_p \log (1+ n/ \lambda)}\right) \leq \delta.
$$
\end{lemma}
\begin{proof}
For the fixed $h \in[H]$, we define the $\sigma$-algebra
$$
\mathcal{F}_{h}^{\tau} =\sigma\big(\big\{\big(x_{h}^{i}, a_{h}^{i}\big)\big\}_{i=1}^{n} \cup\big\{ x_{h+1}^{i} \big\}_{i=1}^{\tau}\big),
$$
where $\sigma(\cdot)$ denotes the $\sigma$-algebra generated by a set of random variables. For all $\tau \in[n]$, we have $\phi_p(x_{h}^{\tau}, a_{h}^{\tau}) \in \mathcal{F}_{h}^{\tau-1}$, as $(x_{h}^{\tau}, a_{h}^{\tau})$ is $\mathcal{F}_{h}^{\tau-1}$-measurable.  Also,
for the fixed function $V$ and all $\tau \in [n]$, we have 
$$
\epsilon_{h}^{\tau}(V) = V\big(x_{h+1}^{\tau}\big)- \E \big[V (x_{h+1}) \mid x_h = x_{h}^{\tau}, a_h = a_{h}^{\tau} \big]  \in \mathcal{F}_{h}^{\tau}
$$
as $x_{h+1}^{\tau}$ is $\mathcal{F}_{h}^{\tau}$-measurable. Hence, $\big\{\epsilon_{h}^{\tau}(V)\big\}_{\tau=1}^{n}$ is a stochastic process adapted to the filtration $\left\{\mathcal{F}_{h}^{\tau}\right\}_{\tau=0}^{n}$. We have
$$
\begin{aligned}
 & \E_{\Dsc}[\epsilon_{h}^{\tau}(V) \mid \mathcal{F}_{h}^{\tau-1}] = \E[\epsilon_{h}^{\tau}(V) \mid \mathcal{F}_{h}^{\tau-1}] = 0.
\end{aligned}
$$
As a result, $\epsilon_h^{\tau}(V)$ is mean-zero and $H$-sub-Gaussian conditioning on $\mathcal{F}_{h}^{\tau-1}$. 

We invoke Lemma \ref{lem:concen_self_normalized} with $M_{0}=\lambda \bI_{d_p}$ and $M_{n} = (\Omega_{h})^{-1} = \widetilde \Lambda_{h} + \lambda \bI_{d_p}$. For the fixed function $V$ and fixed $h \in[H]$, we have
$$
\P_{\Dsc}\left( \Big \| \sum_{\tau=1}^{n}  \phi_p(x_h^{\tau},a_h^{\tau})  \epsilon_{h}^{\tau}  (V) \Big \|_{\Omega_{h}}^2 >  2 H^{2} \cdot \log \big(\frac{\operatorname{det}\big(   \widetilde \Lambda_{h} + \lambda \bI_{d_p} \big)^{1 / 2}}{\delta \cdot \operatorname{det}(\lambda  \bI_{d_p})^{1 / 2}}\big)  \right) \leq \delta.
$$
for all $\delta \in(0,1)$. Note that $\|\phi_p(x, a)\| \leq 1$ for all $(x, a) \in \mathcal{S} \times \Asc$. We have
$\|   \widetilde \Lambda_{h} + \lambda \bI_{d_p} \|_{\rm op} \leq \lambda + n$. Hence, $\operatorname{det}\big(  \widetilde \Lambda_{h} + \lambda \bI_{d_p} \big)  \leq(\lambda+ n)^{d_p}$ and $\operatorname{det}(\lambda \bI_{d_p})=\lambda^{d_p}$, which implies
$$
\P_{\Dsc}\left( \Big \| \sum_{\tau=1}^{n}  \phi_p(x_h^{\tau},a_h^{\tau})  \epsilon_{h}^{\tau}  (V) \Big \|_{\Omega_{h}} > H\sqrt{2 \log (1 / \delta)+d_p \log (1+ n/ \lambda)}\right) \leq \delta.
$$
Therefore, we finish the proof.
\end{proof}

\begin{lemma}[Covering Number of Euclidean Ball] For any $\epsilon >0$, the $\epsilon$-covering number of the Euclidean ball in $\mathbb{R}^d$ with radius $R>0$ is upper bounded by $(1+2 R / \epsilon)^d$.
\label{lem:cover}
\end{lemma}
\begin{proof}
	See Lemma 5.2 in \cite{vershynin2010introduction}.
\end{proof}

\begin{lemma}
\label{lem:cover0}
Let \(\Vsc_h(R,B,J_r,J_p,\underline{\rho},\lambda)\) be the class of functions
indexed by
\(\omega=(\theta,\beta,\Sigma,\Lambda,\gamma_r,\gamma_p)\), with
\[
\begin{aligned}
V_h(x;\omega)
=\max_{a\in\Asc}\Bigg\{\min\Bigg[{}&
g(\langle\phi_r(x,a),\theta\rangle)
-\gamma_r\dot g(\langle\phi_r(x,a),\theta\rangle)
\|\phi_r(x,a)\|_{\Sigma^{-1}}\\
&+\langle\phi_p(x,a),\beta\rangle
-\gamma_p\|\phi_p(x,a)\|_{\Lambda^{-1}},
\ H-h+1\Bigg]^+\Bigg\}.
\end{aligned}
\]
where
\[
\|\theta\|_2\le R,\quad
\|\beta\|_2\le B,\quad
\gamma_r\in[0,J_r],\quad
\gamma_p\in[0,J_p],\quad
\Sigma\succeq\underline{\rho}\bI_{d_r},\quad
\Lambda\succeq\lambda\bI_{d_p}.
\]
Assume \(\max\{\|\phi_r(x,a)\|_2,\|\phi_p(x,a)\|_2\}\le1\) for all
\((x,a)\in\Ssc\times\Asc\). Let
\(\Nsc_h(\epsilon;R,B,J_r,J_p,\underline{\rho},\lambda)\) be its \(\epsilon\)-covering
number under \({\rm dist}(V,V')=\sup_{x\in\Ssc}|V(x)-V'(x)|\). Then
\[
\begin{aligned}
\log\Nsc_h(\epsilon;R,B,J_r,J_p,\underline{\rho},\lambda)
&\le d_r\log\!\left(1+\frac{8L(1+J_r/\sqrt{\underline{\rho}})R}{\epsilon}\right)
+d_p\log(1+8B/\epsilon)\\
&\quad+d_r^2\log\!\left(1+\frac{32L^2d_r^{1/2}J_r^2}{\underline{\rho}\epsilon^2}\right)
+d_p^2\log\!\left(1+\frac{32d_p^{1/2}J_p^2}{\lambda\epsilon^2}\right).
\end{aligned}
\]
\end{lemma}
\begin{proof}
Set \(M_r=\gamma_r^2\Sigma^{-1}\) and
\(M_p=\gamma_p^2\Lambda^{-1}\). Then
\[
\|M_r\|_{\rm op}\le \frac{J_r^2}{\underline{\rho}},
\qquad
\|M_p\|_{\rm op}\le \frac{J_p^2}{\lambda}.
\]
Thus the admissible \(M_r\)'s and \(M_p\)'s are contained in
Frobenius balls of radii \(d_r^{1/2}J_r^2/\underline{\rho}\) and
\(d_p^{1/2}J_p^2/\lambda\), respectively. This enlargement of the two
matrix parameter spaces can only increase their covering numbers.

For fixed \(x,a\), let \(F_{\theta,\beta,M_r,M_p}(x,a)\) denote the
unclipped expression inside the value-function maximum. The maps
\(z\mapsto\min\{z,H-h+1\}^+\) and
\((q_a)_{a\in\Asc}\mapsto\max_{a\in\Asc}q_a\) are nonexpansive in the
supremum norm. Hence it is enough to cover \(F\). For two parameter
quadruples, Assumption~\ref{assump:link}, the feature normalization, and
\[
|\sqrt{x}-\sqrt{y}|\le\sqrt{|x-y|},
\qquad x,y\ge0,
\]
give
\[
\begin{aligned}
{\rm dist}(V,V')
&\le
L\left(1+\frac{J_r}{\sqrt{\underline{\rho}}}\right)\|\theta-\theta'\|_2
+\|\beta-\beta'\|_2\\
&\quad+L\|M_r-M_r'\|_{\rm F}^{1/2}
+\|M_p-M_p'\|_{\rm F}^{1/2}.
\end{aligned}
\]
Indeed, the first two terms control the changes in the reward mean and the
linear transition component. The remaining terms follow by writing the
reward penalty as
\(\dot g(\langle\phi_r,\theta\rangle)
\sqrt{\phi_r^\top M_r\phi_r}\) and using
\[
\sqrt{\phi_r^\top M_r\phi_r}
\le \frac{J_r}{\sqrt{\underline{\rho}}},
\qquad
\left|
\sqrt{\phi^\top M\phi}
-\sqrt{\phi^\top M'\phi}
\right|
\le \|M-M'\|_{\rm F}^{1/2}.
\]

Take Euclidean covers of the \(\theta\)- and \(\beta\)-balls at radii
\[
\delta_\theta
=\frac{\epsilon}{4L(1+J_r/\sqrt{\underline{\rho}})},
\qquad
\delta_\beta=\frac{\epsilon}{4},
\]
and Frobenius-norm covers of the \(M_r\)- and \(M_p\)-balls at radii
\[
\delta_{M_r}=\frac{\epsilon^2}{16L^2},
\qquad
\delta_{M_p}=\frac{\epsilon^2}{16}.
\]
Each of the four terms in the preceding distance bound then contributes at
most \(\epsilon/4\). By Lemma~\ref{lem:cover}, the four covering numbers are
bounded by
\[
\left(1+\frac{8L(1+J_r/\sqrt{\underline{\rho}})R}{\epsilon}\right)^{d_r},
\qquad
\left(1+\frac{8B}{\epsilon}\right)^{d_p},
\]
and
\[
\left(1+\frac{32L^2d_r^{1/2}J_r^2}
{\underline{\rho}\epsilon^2}\right)^{d_r^2},
\qquad
\left(1+\frac{32d_p^{1/2}J_p^2}
{\lambda\epsilon^2}\right)^{d_p^2},
\]
respectively. Multiplying these bounds and taking logarithms proves the
claim. The dimensions \(d_r^2\) and \(d_p^2\) deliberately cover the ambient
matrix spaces; exploiting symmetry would only improve the bound.
\end{proof}

\begin{lemma}[Observed-sample information under missing at random and positivity]\label{lem:observed_sample_information}
Suppose the missing-at-random, positivity, and cross-trajectory sampling
clauses of
Assumption~\ref{assump:step_level_reward_observation_transition_coverage}
hold. Using the main-text definitions of \(q_h\),
\(\Sigma_h^{\rm obs}(\theta_h^*)\), and \(\rho_h^{\rm obs}\),
Assumption~\ref{assump:sigma} implies
\[
\rho_h^{\rm obs}
\ge
\frac{\bar p}{q_h}\rho
>0.
\]
In particular, if \(p_h^{\rm obs}(x,a)\equiv q_h\), then
\(\Sigma_h^{\rm obs}(\theta_h^*)=\Sigma_h(\theta_h^*)\) and
\(\rho_h^{\rm obs}\ge\rho\).

Moreover, conditional on \(\{O_h^\tau\}_{\tau=1}^n\), the observed-sample
weighted design matrix \(\widetilde\Sigma_h^{\rm obs}\) defined in
Supplementary~\ref{sec:step_level_reward_observation} satisfies
\[
\lambda_{\min}\!\left\{
\frac{1}{n_h^{\rm obs}}
\widetilde\Sigma_h^{\rm obs}(\theta_h^*)
\right\}
\ge
\frac{3\rho_h^{\rm obs}}{4}
\]
with probability at least \(1-\xi/(4H)\), provided
\(n_h^{\rm obs}\) is large enough that
\(C\sqrt{\log(Hd_r/\xi)/n_h^{\rm obs}}\le\rho_h^{\rm obs}/4\).
Here \(C>0\) depends only on the fixed upper derivative bound \(L\).
\end{lemma}

\begin{proof}
Under the missing at random condition and by the tower property, the conditional information matrix is
\[
\Sigma_h^{\rm obs}(\theta_h^*)
=
\frac{1}{q_h}
\E_{\pi^b}\!\left[
\dot g\{\langle\phi_r(x_h,a_h),\theta_h^*\rangle\}
p_h^{\rm obs}(x_h,a_h)
\phi_r(x_h,a_h)\phi_r(x_h,a_h)^\top
\right].
\]
Since \(p_h^{\rm obs}(x,a)\ge\bar p\) on \(\mathcal Z_h^b\), Assumption~\ref{assump:sigma}
implies
\[
\Sigma_h^{\rm obs}(\theta_h^*)
\succeq
\frac{\bar p}{q_h}\Sigma_h(\theta_h^*)
\succeq
\frac{\bar p\rho}{q_h}\bI_{d_r}.
\]
If \(p_h^{\rm obs}(x,a)\equiv q_h\), the first display reduces to
\(\Sigma_h^{\rm obs}(\theta_h^*)=\Sigma_h(\theta_h^*)\).

Conditional on the observation indicators, the selected state-action pairs are
independent draws from their conditional law given \(O_h=1\). Matrix Bernstein
therefore gives
\[
\left\|
\frac{1}{n_h^{\rm obs}}
\widetilde\Sigma_h^{\rm obs}(\theta_h^*)
-
\Sigma_h^{\rm obs}(\theta_h^*)
\right\|_{\rm op}
\le
C\sqrt{\frac{\log(Hd_r/\xi)}{n_h^{\rm obs}}}
\]
with probability at least \(1-\xi/(4H)\). Weyl's inequality and the stated
sample-size condition yield the empirical lower bound.
\end{proof}

\begin{lemma}\label{lem:rough_bound}
  Under Assumptions \ref{assump:noise} and \ref{assump:link}, we have
  \[
  \|\theta^*_{h,\lambda'}-\theta_h^*\|_2\leq \|\theta_h^*\|_2
  \]
  for any $h\in[H]$.
\end{lemma}
\begin{proof}
 By the definitions of $\theta_h^*$ and $\theta_{h,\lambda'}^*$, we have
\[
0=\E_{\pi^b}\big[\big(g(\langle\phi_r,\theta^*_{h,\lambda'}\rangle)-g(\langle\phi_r,\theta_h^*\rangle)\big)\phi_r\big]+2\lambda'\theta^*_{h,\lambda'}
\]
Using Taylor's expansion for $F_v(\theta)\coloneqq\E_{\pi^b}\big[g(\langle\phi_r,\theta\rangle)\langle\phi_r,v \rangle\big]$, where $v\in \mathbb{R}^{d_r}$, there exists $t\in [0,1]$ and $\bar \theta=t\theta_h^*+(1-t)\theta_{h,\lambda'}^*$ such that
\begin{equation}\label{eq:taylor}
-2\lambda'\langle v,\theta^*_{h,\lambda'}\rangle=F_v(\theta^*_{h,\lambda'})-F_v(\theta_h^*)=\langle \E_{\pi^b}\big[\dot{g}(\langle\phi_r,\bar \theta\rangle)\phi_r\phi_r^\top\big]v,\theta^*_{h,\lambda'}-\theta_h^*\rangle
\end{equation}
We take $v=\theta_h^*-\theta^*_{h,\lambda'}$, then
\begin{align*}
   2\lambda'\|v\|_2\|\theta_h^*\|_2
   &\geq 2\lambda'\langle v,\theta_h^*\rangle \\
   &=\left\langle \big(2\lambda' \bI_{d_r}+\E_{\pi^b}\big[\dot{g}(\langle\phi_r,\bar \theta\rangle)\phi_r\phi_r^\top\big]\big)v,v\right\rangle \\
   &\geq 2\lambda'\|v\|_2^2.
\end{align*}
Hence $\|\theta^*_{h,\lambda'}-\theta_h^*\|_2\leq \|\theta_h^*\|_2$.
\end{proof}

\begin{lemma}[Regularized Reward-Estimator Error without Curvature]
\label{lem:init1}
Suppose that Assumptions~\ref{assump:noise} and
\ref{assump:link} hold, and that
$0<\lambda'\leq 1$. 
Suppose further that there exists a constant
$C_{\mathrm{score}}>0$, independent of
$n,d_r,H,\xi$, and $\lambda'$, such that

\begin{align}
&
\E_{\pi^b}\!\left[
\left(
\left\{
g\!\left(
\left\langle
\phi_r(x_h,a_h),\theta_{h,\lambda'}^*
\right\rangle
\right)
-r_h
\right\}
\phi_r(x_h,a_h)
+
2\lambda'\theta_{h,\lambda'}^*
\right)
\right.\notag
\\
&\hspace{28mm}\left.
\times
\left(
\left\{
g\!\left(
\left\langle
\phi_r(x_h,a_h),\theta_{h,\lambda'}^*
\right\rangle
\right)
-r_h
\right\}
\phi_r(x_h,a_h)
+
2\lambda'\theta_{h,\lambda'}^*
\right)^\top
\right]\notag
\\
&\hspace{20mm}\preceq
C_{\mathrm{score}}\,
\Sigma_{h,\lambda'}
\!\left(
\theta_{h,\lambda'}^*
\right).\label{eq:ass score}
\end{align}

There exists a constant $C>0$, independent of
$n,d_r,H,\xi$, and $\lambda'$, such that, whenever
\[
n\lambda'
\geq
C\left\{
d_r+\log\frac{16Hd_r}{\xi}
\right\},
\]
with probability at least $1-\xi$, simultaneously for every
$h\in[H]$,
\[
\frac12
\Sigma_{h,\lambda'}
\!\left(
\theta_{h,\lambda'}^*
\right)
\preceq
\frac1n
\widetilde\Sigma_{h,\lambda'}
\!\left(
\theta_{h,\lambda'}^*
\right)
\preceq
\frac32
\Sigma_{h,\lambda'}
\!\left(
\theta_{h,\lambda'}^*
\right),
\]
\[
\left\|
\nabla\mathcal L_{h,\lambda'}
\!\left(
\theta_{h,\lambda'}^*
\right)
\right\|_{
\left\{
n^{-1}
\widetilde\Sigma_{h,\lambda'}
\left(
\theta_{h,\lambda'}^*
\right)
\right\}^{-1}
}^2
\leq
C\frac{
d_r+\log(16H/\xi)
}{n},
\]
\[
\left\|
\widetilde\theta_{h,\lambda'}
-
\theta_{h,\lambda'}^*
\right\|_{
n^{-1}
\widetilde\Sigma_{h,\lambda'}
\left(
\theta_{h,\lambda'}^*
\right)
}^2
\leq
C\frac{
d_r+\log(16H/\xi)
}{n},
\]
\[
\left\|
\widetilde\theta_{h,\lambda'}
-
\theta_{h,\lambda'}^*
\right\|_{
n^{-1}
\widetilde\Sigma_{h,\lambda'}
\left(
\widetilde\theta_{h,\lambda'}
\right)
}^2
\leq
C\frac{
d_r+\log(16H/\xi)
}{n},
\]
and
\[
\left\|
\widetilde\theta_{h,\lambda'}
-
\theta_{h,\lambda'}^*
\right\|_2^2
\leq
C\frac{
d_r+\log(16H/\xi)
}{
n\lambda'
}.
\]
\end{lemma}
\begin{remark}[Interpretation of $C_{\mathrm{score}}$]
Under our standing assumptions and Lemma~\ref{lem:rough_bound}, the
rewards are uniformly sub-Gaussian,
\[
\dot g\!\left(
\langle\phi_r(x_h,a_h),\theta_{h,\lambda'}^*\rangle
\right)
\]
is uniformly bounded away from zero, and the regularization energy
\(\lambda'\|\theta_{h,\lambda'}^*\|_2^2\) is uniformly bounded.
Consequently, the constant \(C_{\mathrm{score}}\) in
\eqref{eq:ass score} is a universal constant independent of
\(n,d_r,H,\xi\), and \(\lambda'\). Indeed, by the elementary inequality
\[
(a+b)(a+b)^\top\preceq
2aa^\top+2bb^\top,
\]
the score covariance is controlled by the conditional residual second
moment together with the ridge term, both of which are uniformly
controlled under the above assumptions. Hence,
\eqref{eq:ass score} does not impose any additional scaling requirement
beyond the standard assumptions already used throughout the paper.
\end{remark}
\begin{proof}
Fix $h\in[H]$ and define
\[
\begin{aligned}
A_{h,\tau}
:={}&
\Sigma_{h,\lambda'}
\!\left(
\theta_{h,\lambda'}^*
\right)^{-1/2}
\dot g\!\left(
\left\langle
\phi_{r,h}^\tau,\theta_{h,\lambda'}^*
\right\rangle
\right)
\phi_{r,h}^\tau
(\phi_{r,h}^\tau)^\top
\\
&\qquad\times
\Sigma_{h,\lambda'}
\!\left(
\theta_{h,\lambda'}^*
\right)^{-1/2}.
\end{aligned}
\]
Since
\[
\Sigma_{h,\lambda'}
\!\left(
\theta_{h,\lambda'}^*
\right)
\succeq
2\lambda'\bI_{d_r},
\]
bounded features and boundedness of $\dot g$ on the relevant compact
predictor range imply
\[
0\preceq A_{h,\tau}
\preceq
\frac{C}{\lambda'}\bI_{d_r},
\qquad
\E A_{h,\tau}\preceq\bI_{d_r}.
\]
Consequently,
\[
A_{h,\tau}^2
\preceq
\frac{C}{\lambda'}A_{h,\tau},
\]
and therefore
\[
\begin{aligned}
&
\left\|
\sum_{\tau=1}^n
\E\left[
\left(
A_{h,\tau}-\E A_{h,\tau}
\right)^2
\right]
\right\|_{\mathrm{op}}
\\
&\qquad\leq
\left\|
\sum_{\tau=1}^n
\E\left[
A_{h,\tau}^2
\right]
\right\|_{\mathrm{op}}
\leq
\frac{Cn}{\lambda'}.
\end{aligned}
\]
Matrix Bernstein's inequality and a union bound over $h\in[H]$ yield
\[
\begin{aligned}
&
\left\|
\Sigma_{h,\lambda'}
\!\left(
\theta_{h,\lambda'}^*
\right)^{-1/2}
\left\{
\frac1n
\widetilde\Sigma_{h,\lambda'}
\!\left(
\theta_{h,\lambda'}^*
\right)
-
\Sigma_{h,\lambda'}
\!\left(
\theta_{h,\lambda'}^*
\right)
\right\}
\right.
\\
&\hspace{30mm}\left.
\times
\Sigma_{h,\lambda'}
\!\left(
\theta_{h,\lambda'}^*
\right)^{-1/2}
\right\|_{\mathrm{op}}
\leq
\frac12
\end{aligned}
\]
simultaneously for every $h\in[H]$, provided
\[
n\lambda'
\geq
C\log\frac{16Hd_r}{\xi}.
\]
This proves the relative Loewner comparison.

For every $\tau\in[n]$, let
\[
\begin{aligned}
Z_{h,\tau}
:={}&
\left\{
g\!\left(
\left\langle
\phi_{r,h}^\tau,\theta_{h,\lambda'}^*
\right\rangle
\right)
-r_h^\tau
\right\}
\phi_{r,h}^\tau
+
2\lambda'\theta_{h,\lambda'}^*.
\end{aligned}
\]
The population first-order condition gives
$\E Z_{h,\tau}=0$, and
\[
\nabla\mathcal L_{h,\lambda'}
\!\left(
\theta_{h,\lambda'}^*
\right)
=
\frac1n\sum_{\tau=1}^n Z_{h,\tau}.
\]
The score--information comparison condition gives
\[
\E\left[
Z_{h,\tau}Z_{h,\tau}^\top
\right]
\preceq
C_{\mathrm{score}}\,
\Sigma_{h,\lambda'}
\!\left(
\theta_{h,\lambda'}^*
\right).
\]
Moreover, Assumption~\ref{assump:noise}, bounded features,
Lemma~\ref{lem:rough_bound}, and
\[
\Sigma_{h,\lambda'}
\!\left(
\theta_{h,\lambda'}^*
\right)
\succeq
2\lambda'\bI_{d_r}
\]
imply that, for every $v\in\mathbb S^{d_r-1}$,
\[
\left\|
v^\top
\Sigma_{h,\lambda'}
\!\left(
\theta_{h,\lambda'}^*
\right)^{-1/2}
Z_{h,\tau}
\right\|_{\psi_1}
\leq
\frac{C}{\sqrt{\lambda'}}.
\]
Bernstein's inequality on a $1/2$-net of
$\mathbb S^{d_r-1}$, followed by a union bound over $h\in[H]$,
therefore yields
\[
\begin{aligned}
&
\left\|
\nabla\mathcal L_{h,\lambda'}
\!\left(
\theta_{h,\lambda'}^*
\right)
\right\|_{
\Sigma_{h,\lambda'}
\left(
\theta_{h,\lambda'}^*
\right)^{-1}
}
\\
&\qquad\leq
C\left\{
\sqrt{
\frac{
d_r+\log(16H/\xi)
}{n}
}
+
\frac{
d_r+\log(16H/\xi)
}{
n\sqrt{\lambda'}
}
\right\}.
\end{aligned}
\]
The sample-size condition implies
\[
\frac{
d_r+\log(16H/\xi)
}{
n\sqrt{\lambda'}
}
\leq
C
\sqrt{
\frac{
d_r+\log(16H/\xi)
}{n}
},
\]
and hence
\[
\left\|
\nabla\mathcal L_{h,\lambda'}
\!\left(
\theta_{h,\lambda'}^*
\right)
\right\|_{
\Sigma_{h,\lambda'}
\left(
\theta_{h,\lambda'}^*
\right)^{-1}
}
\leq
C
\sqrt{
\frac{
d_r+\log(16H/\xi)
}{n}
}.
\]
The relative Loewner comparison consequently gives
\[
\left\|
\nabla\mathcal L_{h,\lambda'}
\!\left(
\theta_{h,\lambda'}^*
\right)
\right\|_{
\left\{
n^{-1}
\widetilde\Sigma_{h,\lambda'}
\left(
\theta_{h,\lambda'}^*
\right)
\right\}^{-1}
}
\leq
C
\sqrt{
\frac{
d_r+\log(16H/\xi)
}{n}
}.
\]

Let $C_{\mathrm{sc}}>0$ denote the constant in the preceding score
bound, and consider
\[
\left\{
\left\|
\theta-\theta_{h,\lambda'}^*
\right\|_{
n^{-1}
\widetilde\Sigma_{h,\lambda'}
\left(
\theta_{h,\lambda'}^*
\right)
}
\leq
8C_{\mathrm{sc}}
\sqrt{
\frac{
d_r+\log(16H/\xi)
}{n}
}
\right\}.
\]
Because
\[
\frac1n
\widetilde\Sigma_{h,\lambda'}
\!\left(
\theta_{h,\lambda'}^*
\right)
\succeq
2\lambda'\bI_{d_r},
\]
every $\theta$ in this set satisfies
\[
\left\|
\theta-\theta_{h,\lambda'}^*
\right\|_2
\leq
\frac{8C_{\mathrm{sc}}}{\sqrt{2\lambda'}}
\sqrt{
\frac{
d_r+\log(16H/\xi)
}{n}
}.
\]

By Lemma~\ref{lem:rough_bound}, the population targets
$\theta_{h,\lambda'}^*$ belong to a fixed bounded set. Let
$\mathcal I$ be a fixed compact interval containing all predictors
generated by parameters within Euclidean distance one of these
population targets. Since $\dot g$ is strictly positive and
$\ddot g$ is continuous on $\mathcal I$, the function
$\log\dot g$ is Lipschitz on $\mathcal I$. Thus, there exists a
constant $C>0$, independent of $n,d_r,H,\xi$, and $\lambda'$, such
that
\[
\left|
\log\dot g(u)-\log\dot g(v)
\right|
\leq
C|u-v|,
\qquad
u,v\in\mathcal I.
\]
By increasing the constant in the sample-size condition, every
parameter in the preceding information ball is within Euclidean
distance one of $\theta_{h,\lambda'}^*$ and satisfies
\[
C
\left\|
\theta-\theta_{h,\lambda'}^*
\right\|_2
\leq
\log2.
\]
Consequently, for every such $\theta$, every $t\in[0,1]$, and every
$\tau\in[n]$,
\[
\begin{aligned}
&
\left|
\log
\frac{
\dot g\!\left(
\left\langle
\phi_{r,h}^\tau,
\theta_{h,\lambda'}^*
+t(\theta-\theta_{h,\lambda'}^*)
\right\rangle
\right)
}{
\dot g\!\left(
\left\langle
\phi_{r,h}^\tau,\theta_{h,\lambda'}^*
\right\rangle
\right)
}
\right|
\\
&\qquad\leq
C\left\|
\theta-\theta_{h,\lambda'}^*
\right\|_2
\leq
\log2.
\end{aligned}
\]
It follows that
\[
\begin{aligned}
&
\frac12
\dot g\!\left(
\left\langle
\phi_{r,h}^\tau,\theta_{h,\lambda'}^*
\right\rangle
\right)
\\
&\qquad\leq
\dot g\!\left(
\left\langle
\phi_{r,h}^\tau,
\theta_{h,\lambda'}^*
+t(\theta-\theta_{h,\lambda'}^*)
\right\rangle
\right)
\\
&\qquad\leq
2
\dot g\!\left(
\left\langle
\phi_{r,h}^\tau,\theta_{h,\lambda'}^*
\right\rangle
\right).
\end{aligned}
\]
Multiplying by
$\phi_{r,h}^\tau(\phi_{r,h}^\tau)^\top$, summing over $\tau$, and
adding the ridge term gives
\[
\begin{aligned}
&
\frac12
\frac1n
\widetilde\Sigma_{h,\lambda'}
\!\left(
\theta_{h,\lambda'}^*
\right)
\\
&\qquad\preceq
\frac1n
\widetilde\Sigma_{h,\lambda'}
\!\left(
\theta_{h,\lambda'}^*
+t(\theta-\theta_{h,\lambda'}^*)
\right)
\\
&\qquad\preceq
2
\frac1n
\widetilde\Sigma_{h,\lambda'}
\!\left(
\theta_{h,\lambda'}^*
\right).
\end{aligned}
\]

For every feasible $\Delta$ satisfying
\[
\|\Delta\|_{
n^{-1}
\widetilde\Sigma_{h,\lambda'}
\left(
\theta_{h,\lambda'}^*
\right)
}
=
8C_{\mathrm{sc}}
\sqrt{
\frac{
d_r+\log(16H/\xi)
}{n}
},
\]
Taylor's integral formula gives
\[
\begin{aligned}
&
\mathcal L_{h,\lambda'}
\!\left(
\theta_{h,\lambda'}^*+\Delta
\right)
-
\mathcal L_{h,\lambda'}
\!\left(
\theta_{h,\lambda'}^*
\right)
\\
&=
\left\langle
\nabla\mathcal L_{h,\lambda'}
\!\left(
\theta_{h,\lambda'}^*
\right),
\Delta
\right\rangle
\\
&\quad+
\int_0^1
(1-t)
\Delta^\top
\left\{
\frac1n
\widetilde\Sigma_{h,\lambda'}
\!\left(
\theta_{h,\lambda'}^*+t\Delta
\right)
\right\}
\Delta\,dt
\\
&\geq
-
C_{\mathrm{sc}}
\sqrt{
\frac{
d_r+\log(16H/\xi)
}{n}
}
\|\Delta\|_{
n^{-1}
\widetilde\Sigma_{h,\lambda'}
\left(
\theta_{h,\lambda'}^*
\right)
}
\\
&\quad+
\frac14
\|\Delta\|_{
n^{-1}
\widetilde\Sigma_{h,\lambda'}
\left(
\theta_{h,\lambda'}^*
\right)
}^2
\\
&=
8C_{\mathrm{sc}}^2
\frac{
d_r+\log(16H/\xi)
}{n}
>0.
\end{aligned}
\]
Convexity of $\Theta$ and optimality of
$\widetilde\theta_{h,\lambda'}$ imply
\[
\left\|
\widetilde\theta_{h,\lambda'}
-
\theta_{h,\lambda'}^*
\right\|_{
n^{-1}
\widetilde\Sigma_{h,\lambda'}
\left(
\theta_{h,\lambda'}^*
\right)
}
\leq
8C_{\mathrm{sc}}
\sqrt{
\frac{
d_r+\log(16H/\xi)
}{n}
}.
\]
Taking $t=1$ in the Hessian comparison proves the endpoint
information-norm bound. Finally,
\[
\frac1n
\widetilde\Sigma_{h,\lambda'}
\!\left(
\theta_{h,\lambda'}^*
\right)
\succeq
2\lambda'\bI_{d_r}
\]
gives
\[
\left\|
\widetilde\theta_{h,\lambda'}
-
\theta_{h,\lambda'}^*
\right\|_2^2
\leq
\frac{1}{2\lambda'}
\left\|
\widetilde\theta_{h,\lambda'}
-
\theta_{h,\lambda'}^*
\right\|_{
n^{-1}
\widetilde\Sigma_{h,\lambda'}
\left(
\theta_{h,\lambda'}^*
\right)
}^2.
\]
A union bound over the preceding events completes the proof.
\end{proof}
\begin{lemma}[Regularized Reward Prediction Bound without Curvature]
\label{lem:g1}
Suppose that the conditions of Lemmas~\ref{lem:rough_bound} and
\ref{lem:init1} hold and that $g$ is $L$-Lipschitz. There exists a
constant $C>0$, independent of $n,d_r,H,\xi$, and $\lambda'$, such
that, with probability at least $1-\xi$, simultaneously for every
$h\in[H]$ and $(x,a)\in\Ssc\times\Asc$,
\begin{align}
&
\left|
g\!\left(
\left\langle
\phi_r(x,a),\widetilde\theta_{h,\lambda'}
\right\rangle
\right)
-
g\!\left(
\left\langle
\phi_r(x,a),\theta_h^*
\right\rangle
\right)
\right| \notag
\\
&\leq
C
\sqrt{
d_r+\log\frac{16H}{\xi}
}
\,
\dot g\!\left(
\left\langle
\phi_r(x,a),\widetilde\theta_{h,\lambda'}
\right\rangle
\right)
\left\|
\phi_r(x,a)
\right\|_{
\widetilde\Sigma_{h,\lambda'}
\left(
\widetilde\theta_{h,\lambda'}
\right)^{-1}
}\label{eq:g_bound}
\\
&+CL\sqrt{\lambda'}
\left\|
\phi_r(x,a)
\right\|_{
(2\lambda'\bI_{d_r}
+n^{-1}\widetilde\Lambda_{r,h})^{-1}
}.\notag
\end{align}
\end{lemma}

\begin{proof}
By the triangle inequality,
\[
\begin{aligned}
&
\left|
g\!\left(
\left\langle
\phi_r(x,a),\widetilde\theta_{h,\lambda'}
\right\rangle
\right)
-
g\!\left(
\left\langle
\phi_r(x,a),\theta_h^*
\right\rangle
\right)
\right|
\\
&\leq
\left|
g\!\left(
\left\langle
\phi_r(x,a),\widetilde\theta_{h,\lambda'}
\right\rangle
\right)
-
g\!\left(
\left\langle
\phi_r(x,a),\theta_{h,\lambda'}^*
\right\rangle
\right)
\right|
\\
&\quad+
\left|
g\!\left(
\left\langle
\phi_r(x,a),\theta_{h,\lambda'}^*
\right\rangle
\right)
-
g\!\left(
\left\langle
\phi_r(x,a),\theta_h^*
\right\rangle
\right)
\right|.
\end{aligned}
\]

By Lemma~\ref{lem:init1},
\[
\left\|
\widetilde\theta_{h,\lambda'}
-
\theta_{h,\lambda'}^*
\right\|_2
\leq
C
\sqrt{
\frac{
d_r+\log(16H/\xi)
}{
n\lambda'
}
}.
\]
The compactness argument in the proof of Lemma~\ref{lem:init1}
implies that $\log\dot g$ is Lipschitz on a fixed compact interval
containing all predictors along the segment joining
$\theta_{h,\lambda'}^*$ and $\widetilde\theta_{h,\lambda'}$.
Consequently, under the sample-size condition of
Lemma~\ref{lem:init1}, the mean-value theorem gives
\[
\begin{aligned}
&
\left|
g\!\left(
\left\langle
\phi_r(x,a),\widetilde\theta_{h,\lambda'}
\right\rangle
\right)
-
g\!\left(
\left\langle
\phi_r(x,a),\theta_{h,\lambda'}^*
\right\rangle
\right)
\right|
\\
&\leq
2
\dot g\!\left(
\left\langle
\phi_r(x,a),\widetilde\theta_{h,\lambda'}
\right\rangle
\right)
\left|
\left\langle
\phi_r(x,a),
\widetilde\theta_{h,\lambda'}
-
\theta_{h,\lambda'}^*
\right\rangle
\right|.
\end{aligned}
\]
Cauchy--Schwarz and Lemma~\ref{lem:init1} therefore yield
\[
\begin{aligned}
&
\left|
g\!\left(
\left\langle
\phi_r(x,a),\widetilde\theta_{h,\lambda'}
\right\rangle
\right)
-
g\!\left(
\left\langle
\phi_r(x,a),\theta_{h,\lambda'}^*
\right\rangle
\right)
\right|
\\
&\leq
C
\dot g\!\left(
\left\langle
\phi_r(x,a),\widetilde\theta_{h,\lambda'}
\right\rangle
\right)
\sqrt{
\frac{
d_r+\log(16H/\xi)
}{n}
}
\\
&\qquad\times
\left\|
\phi_r(x,a)
\right\|_{
\left\{
n^{-1}
\widetilde\Sigma_{h,\lambda'}
\left(
\widetilde\theta_{h,\lambda'}
\right)
\right\}^{-1}
}
\\
&=
C
\sqrt{
d_r+\log\frac{16H}{\xi}
}
\,
\dot g\!\left(
\left\langle
\phi_r(x,a),\widetilde\theta_{h,\lambda'}
\right\rangle
\right)
\left\|
\phi_r(x,a)
\right\|_{
\widetilde\Sigma_{h,\lambda'}
\left(
\widetilde\theta_{h,\lambda'}
\right)^{-1}
}.
\end{aligned}
\]

The population first-order conditions at $\theta_h^*$ and
$\theta_{h,\lambda'}^*$, together with Taylor's integral formula, imply
\[
\begin{aligned}
&
\left\{
2\lambda'\bI_{d_r}
+
\int_0^1
\E_{\pi^b}\!\left[
\dot g\!\left(
\left\langle
\phi_r(x_h,a_h),
\theta_h^*
+t\left(
\theta_{h,\lambda'}^*-\theta_h^*
\right)
\right\rangle
\right)
\phi_r(x_h,a_h)\phi_r(x_h,a_h)^\top
\right]dt
\right\}
\\
&\qquad\times
\left(
\theta_{h,\lambda'}^*-\theta_h^*
\right)
=
-2\lambda'\theta_h^*.
\end{aligned}
\]
Moreover, suppose $c_1=\inf_{\|x\|_2\le 2\max_h\|\theta^*_h\|_2}\dot g(x)$, then
\[
\begin{aligned}
&
\int_0^1
\E_{\pi^b}\!\left[
\dot g\!\left(
\left\langle
\phi_r(x_h,a_h),
\theta_h^*
+t\left(
\theta_{h,\lambda'}^*-\theta_h^*
\right)
\right\rangle
\right)
\phi_r(x_h,a_h)\phi_r(x_h,a_h)^\top
\right]dt
\\
&\qquad\succeq c_1\Lambda_{r,h}.
\end{aligned}
\]
It follows that, for some constant $c_2>0$,
\[
\begin{aligned}
\left\|
\theta_{h,\lambda'}^*-\theta_h^*
\right\|_{
2\lambda'\bI_{d_r}+c_1\Lambda_{r,h}
}
&\leq
\left\|
\theta_{h,\lambda'}^*-\theta_h^*
\right\|_{
2\lambda'\bI_{d_r}
+
\displaystyle
\int_0^1
\E_{\pi^b}\!\left[
\dot g\!\left(
\left\langle
\phi_r(x_h,a_h),
\theta_h^*
+t\left(
\theta_{h,\lambda'}^*-\theta_h^*
\right)
\right\rangle
\right)
\phi_r(x_h,a_h)\phi_r(x_h,a_h)^\top
\right]dt
}
\\
&=
2\lambda'
\left\|
\theta_h^*
\right\|_{
\left\{
2\lambda'\bI_{d_r}
+
\displaystyle
\int_0^1
\E_{\pi^b}\!\left[
\dot g\!\left(
\left\langle
\phi_r(x_h,a_h),
\theta_h^*
+t\left(
\theta_{h,\lambda'}^*-\theta_h^*
\right)
\right\rangle
\right)
\phi_r(x_h,a_h)\phi_r(x_h,a_h)^\top
\right]dt
\right\}^{-1}
}
\\
&\leq
\sqrt{c_2\lambda'}.
\end{aligned}
\]
Since $g$ is $L$-Lipschitz, Cauchy--Schwarz gives
\[
\begin{aligned}
&
\left|
g\!\left(
\left\langle
\phi_r(x,a),\theta_{h,\lambda'}^*
\right\rangle
\right)
-
g\!\left(
\left\langle
\phi_r(x,a),\theta_h^*
\right\rangle
\right)
\right|
\\
&\leq
L
\left\|
\phi_r(x,a)
\right\|_{
(2\lambda'\bI_{d_r}+c_1\Lambda_{r,h})^{-1}
}
\left\|
\theta_{h,\lambda'}^*-\theta_h^*
\right\|_{
2\lambda'\bI_{d_r}+c_1\Lambda_{r,h}
}
\\
&\leq
L\sqrt{\frac{c_2\lambda'}{c_1}}\,
\left\|
\phi_r(x,a)
\right\|_{
(2\lambda'\bI_{d_r}+\Lambda_{r,h})^{-1}
}.
\end{aligned}
\]

Moreover, a regularized relative matrix Bernstein inequality and a
union bound over $h\in[H]$ imply that, with probability at least
$1-\xi$,
\[
\begin{aligned}
\max_{h\in[H]}
\Bigl\|
&
(2\lambda'\bI_{d_r}+\Lambda_{r,h})^{-1/2}
\left(
n^{-1}\widetilde\Lambda_{r,h}-\Lambda_{r,h}
\right)
\\
&\qquad\qquad\times
(2\lambda'\bI_{d_r}+\Lambda_{r,h})^{-1/2}
\Bigr\|_{\mathrm{op}}
\\
&\leq
c_3
\left\{
\sqrt{
\frac{\log(Hd_r/\xi)}
{n\lambda'}
}
+
\frac{\log(Hd_r/\xi)}
{n\lambda'}
\right\}.
\end{aligned}
\]
Consequently,
\[
\begin{aligned}
n^{-1}\widetilde\Lambda_{r,h}-\Lambda_{r,h}
\preceq
c_3
\left\{
\sqrt{
\frac{\log(Hd_r/\xi)}
{n\lambda'}
}
+
\frac{\log(Hd_r/\xi)}
{n\lambda'}
\right\}
(2\lambda'\bI_{d_r}+\Lambda_{r,h}),
\end{aligned}
\]
and hence
\[
\begin{aligned}
2\lambda'\bI_{d_r}
+n^{-1}\widetilde\Lambda_{r,h}
\preceq
\left[
1+
c_3
\left\{
\sqrt{
\frac{\log(Hd_r/\xi)}
{n\lambda'}
}
+
\frac{\log(Hd_r/\xi)}
{n\lambda'}
\right\}
\right]
(2\lambda'\bI_{d_r}+\Lambda_{r,h}).
\end{aligned}
\]
By the order-reversing property of matrix inversion,
\[
\begin{aligned}
(2\lambda'\bI_{d_r}+\Lambda_{r,h})^{-1}
\preceq
\left[
1+
c_3
\left\{
\sqrt{
\frac{\log(Hd_r/\xi)}
{n\lambda'}
}
+
\frac{\log(Hd_r/\xi)}
{n\lambda'}
\right\}
\right]
(2\lambda'\bI_{d_r}
+n^{-1}\widetilde\Lambda_{r,h})^{-1}.
\end{aligned}
\]
Therefore,
\[
\begin{aligned}
&
\left|
g\!\left(
\left\langle
\phi_r(x,a),\theta_{h,\lambda'}^*
\right\rangle
\right)
-
g\!\left(
\left\langle
\phi_r(x,a),\theta_h^*
\right\rangle
\right)
\right|
\\
&\leq
L\sqrt{\frac{c_2\lambda'}{c_1}}\,
\left[
1+
c_3
\left\{
\sqrt{
\frac{\log(Hd_r/\xi)}
{n\lambda'}
}
+
\frac{\log(Hd_r/\xi)}
{n\lambda'}
\right\}
\right]^{1/2}
\\
&\qquad\qquad\times
\left\|
\phi_r(x,a)
\right\|_{
(2\lambda'\bI_{d_r}
+n^{-1}\widetilde\Lambda_{r,h})^{-1}
}.
\end{aligned}
\]
In particular, whenever
\[
c_3
\left\{
\sqrt{
\frac{\log(Hd_r/\xi)}
{n\lambda'}
}
+
\frac{\log(Hd_r/\xi)}
{n\lambda'}
\right\}
\leq \frac12,
\]
we obtain
\[
\begin{aligned}
&
\left|
g\!\left(
\left\langle
\phi_r(x,a),\theta_{h,\lambda'}^*
\right\rangle
\right)
-
g\!\left(
\left\langle
\phi_r(x,a),\theta_h^*
\right\rangle
\right)
\right|
\\
&\leq
L\sqrt{\frac{3c_2\lambda'}{2c_1}}\,
\left\|
\phi_r(x,a)
\right\|_{
(2\lambda'\bI_{d_r}
+n^{-1}\widetilde\Lambda_{r,h})^{-1}
}.
\end{aligned}
\]
\end{proof}
\begin{lemma}\label{lem:ucb}
    Under the assumptions of Theorem \ref{thm:regret_online}, with probability at least $1-2p_0/3$, \[\bar Q_{h,t}(x,a)\geq Q^*_h(x,a)\] holds for all $h\in[H]$, $t\in[T]$, and $(x,a)\in\Ssc\times\Asc$.
    \end{lemma}
    \begin{proof}
    Write $\bar V_{h+1,t}(x')=\max_{a'\in\Asc}\bar Q_{h+1,t}(x',a')$. We argue by backward induction, starting from $\bar V_{H+1,t}=V^*_{H+1}=0$. Suppose $\bar V_{h+1,t}\geq V^*_{h+1}$ pointwise. On the simultaneous reward- and transition-confidence event from Lemmas~\ref{lem:theta_err_online} and~\ref{lem:beta_err_online}, the optimistic bonuses in Algorithm~\ref{alg:online} give
    \[
    \begin{aligned}
    \bar Q_{h,t}(x,a)
    &\geq \E\{r_h+\bar V_{h+1,t}(x_{h+1})\mid x_h=x,a_h=a\}\\
    &\geq \E\{r_h+V^*_{h+1}(x_{h+1})\mid x_h=x,a_h=a\}\\
    &=Q_h^*(x,a).
    \end{aligned}
    \]
    Before clipping, the optimistic update is at least the first conditional
    expectation in the display. This expectation lies in $[0,H-h+1]$ because
    rewards lie in $[0,1]$ and $0\le\bar V_{h+1,t}\le H-h$. Therefore the
    two-sided clipping in Algorithm~\ref{alg:online} preserves the inequality.
    Taking the maximum over $a$ completes the induction. The simultaneous event
    has probability at least $1-2p_0/3$.
    \end{proof}
    
\begin{lemma}[Covering the online optimistic value class]
\label{lem:online_value_cover}
For $R,B,J_r,J_p>0$, let $\Vsc_h^{\rm on}(R,B,J_r,J_p)$ contain the functions
\[
\begin{aligned}
V(x)=\max_{a\in\Asc}\max\!\bigg\{0,\min\!\bigg\{H-h+1,
&g(\phi_r(x,a)^\top\theta)+\phi_p(x,a)^\top\beta\\
&+\sqrt{\phi_r(x,a)^\top M_r\phi_r(x,a)}
+\sqrt{\phi_p(x,a)^\top M_p\phi_p(x,a)}\bigg\}\bigg\}.
\end{aligned}
\]
where $\|\theta\|_2\le R$, $\|\beta\|_2\le B$, and $M_r,M_p$ are positive
semidefinite with $\|M_r\|_{\rm op}\le J_r^2$ and
$\|M_p\|_{\rm op}\le J_p^2$. Under the feature normalization, its
$\epsilon$-covering number in supremum norm satisfies
\[
\begin{aligned}
\log\Nsc_h^{\rm on}(\epsilon)
&\le d_r\log(1+8KR/\epsilon)
+d_p\log(1+8B/\epsilon)\\
&\quad+d_r^2\log\!\left(1+
\frac{32d_r^{1/2}J_r^2}{\epsilon^2}\right)
+d_p^2\log\!\left(1+
\frac{32d_p^{1/2}J_p^2}{\epsilon^2}\right).
\end{aligned}
\]
Moreover, all $\bar V_{h,t}$ with $t\in [T]$ generated by Algorithm~\ref{alg:online} belong
to $\Vsc_h^{\rm on}(M,H\sqrt{d_pT},\gamma_r,\gamma_p)$.
\end{lemma}
\begin{proof}
The maximum over actions and the two-sided clipping operator are
one-Lipschitz in the supremum norm. For two parameter tuples, the difference
between their unclipped functions is at most
\[
K\|\theta-\theta'\|_2+\|\beta-\beta'\|_2
+\|M_r-M_r'\|_{\rm F}^{1/2}
+\|M_p-M_p'\|_{\rm F}^{1/2}.
\]
Euclidean covers for $\theta,\beta$ and Frobenius covers for $M_r,M_p$, as in
Lemma~\ref{lem:cover0}, give the displayed entropy bound. The constraint in
\eqref{eq:glm_est} gives $\|\widehat\theta_{h,t}\|_2\le M$. Because the
two-sided clipping makes $0\le\bar V_{h+1,t}\le H$, the ridge-regression
definition and the same singular-value calculation as Lemma~\ref{lem:norm}
give $\|\widehat\beta_{h,t}\|_2\le H\sqrt{d_pt}\le H\sqrt{d_pT}$.
Finally, take
$M_r=\gamma_r^2\Lambda_{r,h,t}^{-1}$ and
$M_p=\gamma_p^2\Lambda_{p,h,t}^{-1}$. Since both covariance matrices dominate
the identity, their operator-norm constraints hold, proving membership.
\end{proof}
\begin{lemma}\label{lem:theta_err_online}
       Under Assumptions \ref{assump:theta_online} and \ref{assump:g_online}, for any $p_1 \in (0,1)$, with probability at least $1-p_1$, simultaneously for all $h\in[H]$ and $t\in[T]$,
       \[
       \Big|g(\phi_r(x,a)^\top\widehat\theta_{h,t})-g(\phi_r(x,a)^\top\theta_{h}^*)\Big|
       \leq K\sqrt{4M^2+\frac{4+8\sqrt{A_t(p_1)}}{k}+\frac{16A_t(p_1)}{k^2}}
       \|\phi_r(x,a)\|_{\Lambda_{r,h,t}^{-1}}.
       \]
    \end{lemma}
\begin{proof}
        Denote $\Delta_{h,t}=\widehat \theta_{h,t}-\theta_h^*$. By the definition of $\widehat\theta_{h,t}$, we have\[
        \sum_{\tau=1}^tr_{h,\tau}\langle \phi_r(x_{h,\tau},a_{h,\tau}),-\Delta_{h,t}\rangle\leq \sum_{\tau=1}^t\int_{\langle \phi_r(x_{h,\tau},a_{h,\tau}),\widehat \theta_{h,t}\rangle}^{\langle \phi_r(x_{h,\tau},a_{h,\tau}),\theta_h^*\rangle}g(u)du,
        \]
        Together with \eqref{eq:grasp-model}, we have\begin{equation}\label{eq:online_theta}\begin{split}
      &\sum_{\tau=1}^t(r_{h,\tau}-g(\langle \phi_r(x_{h,\tau},a_{h,\tau}), \theta_{h}^*\rangle))\langle \phi_r(x_{h,\tau},a_{h,\tau}),\Delta_{h,t}\rangle\\&\geq \sum_{\tau=1}^t\int_{\langle \phi_r(x_{h,\tau},a_{h,\tau}),\theta_h^*\rangle}^{\langle \phi_r(x_{h,\tau},a_{h,\tau}),\widehat \theta_{h,t}\rangle}\big(g(u)-g(\langle \phi_r(x_{h,\tau},a_{h,\tau}), \theta_{h}^*\rangle)\big)du,
      \end{split}
        \end{equation}
        The right side is larger than \[
        \frac{1}{2}\sum_{\tau=1}^tk\langle \phi_r(x_{h,\tau},a_{h,\tau}),\Delta_{h,t}\rangle^2:=\frac{1}{2}kV_{h,t}.
        \]
        where we use the uniform lower bound on $\dot{g}$. Under the alternative conditions in Remark~\ref{rmk:self_concordance}, the same inequality holds with $k_{\mathrm{eff}}$ in place of $k$.
        We bound the left-hand side of \eqref{eq:online_theta} using Lemma 9 of \citet{wang2019optimism}: for any $\delta \in (0,1)$, with probability at least $1-\delta$,
        \[
        \begin{aligned}
        &\sum_{\tau=1}^t
        \bigl(r_{h,\tau}-g(\langle \phi_r(x_{h,\tau},a_{h,\tau}),
        \theta_{h}^*\rangle)\bigr)
        \langle \phi_r(x_{h,\tau},a_{h,\tau}),\Delta_{h,t}\rangle\\
        &\qquad\leq
        1+2\Big(1+\sqrt{V_{h,t}}\Big)
        \sqrt{d_r\log(4Mt)+\log(1/\delta)}.
        \end{aligned}
        \] 
        Here the covering radius in Lemma 9 of \citet{wang2019optimism} is
        $2M$ because $\|\Delta_{h,t}\|_2\le2M$. Thus let
        $B_t(\delta)=d_r\log(4Mt)+\log(1/\delta)$. Combining the preceding upper and lower bounds gives
        \[
        kV_{h,t}\leq 2+4\sqrt{B_t(\delta)}+4\sqrt{B_t(\delta)V_{h,t}}.
        \]
        Applying $4\sqrt{B_t(\delta)V_{h,t}}\leq (k/2)V_{h,t}+8B_t(\delta)/k$ and rearranging yields
        \[
        V_{h,t}\leq \frac{4+8\sqrt{B_t(\delta)}}{k}+\frac{16B_t(\delta)}{k^2}.
        \]
        Taking $\delta=p_1/(TH)$ gives $B_t(\delta)=A_t(p_1)$; a union bound proves the simultaneous claim.
        By the mean-value theorem and Assumption \ref{assump:g_online}, we have
        \[
\Big|g(\phi_r(x,a)^\top\widehat\theta_{h,t})-g(\phi_r(x,a)^\top\theta_{h}^*)\Big|
\leq K|\langle \phi_r(x,a),\Delta_{h,t}\rangle|
\leq K\|\phi_r(x,a)\|_{\Lambda_{r,h,t}^{-1}}\|\Delta_{h,t}\|_{\Lambda_{r,h,t}}.
        \]
        Since $\Lambda_{r,h,t}=\bI_{d_r}+\sum_{\tau=1}^t\phi_r(x_{h,\tau},a_{h,\tau})\phi_r(x_{h,\tau},a_{h,\tau})^\top$, we have
        \[
        \|\Delta_{h,t}\|_{\Lambda_{r,h,t}}^2=\|\Delta_{h,t}\|_2^2+\sum_{\tau=1}^t\langle \phi_r(x_{h,\tau},a_{h,\tau}),\Delta_{h,t}\rangle^2\leq 4M^2+V_{h,t},
        \]
        where we used $\|\widehat\theta_{h,t}\|_2,\|\theta_h^*\|_2\leq M$. Combining this with the bound on $V_{h,t}$ proves the claim.
    \end{proof}

\begin{lemma}[Uniform concentration for the current online value]
\label{lem:online_current_value_concentration}
Under the assumptions of Theorem~\ref{thm:regret_online}, with probability at
least $1-p_0/3$, simultaneously for $h\in[H]$ and $t\in[T]$,
\[
\left\|
\sum_{\tau=1}^t
\phi_p(x_{h,\tau},a_{h,\tau})
\left[
\bar V_{h+1,t}(x_{h+1,\tau})
-\E\{\bar V_{h+1,t}(x_{h+1})\mid x_{h,\tau},a_{h,\tau}\}
\right]
\right\|_{\Lambda_{p,h,t}^{-1}}
\le C H(d_r+d_p)\sqrt{\zeta_{\rm on}(p_0)}
\]
for a universal constant $C>0$.
\end{lemma}
\begin{proof}
Fix $h$ and a deterministic $V:\Ssc\to[0,H]$. Across episodes, the feature
$\phi_p(x_{h,\tau},a_{h,\tau})$ is predictable before the next state is
drawn, while
\[
V(x_{h+1,\tau})-\E\{V(x_{h+1})\mid x_{h,\tau},a_{h,\tau}\}
\]
is a conditionally mean-zero, $H$-sub-Gaussian martingale difference.
Lemma~\ref{lem:concen_self_normalized} therefore applies simultaneously over
$t$ to each fixed $V$.

Take an $\epsilon$-net of the class in
Lemma~\ref{lem:online_value_cover} with $\epsilon=H/T$, and apply the preceding
self-normalized inequality to every net element and every $h$, with a union
bound. The determinant bound
$\det(\Lambda_{p,h,t})\le(1+t/d_p)^{d_p}$ gives, simultaneously over the net,
\[
\left\|\sum_{\tau=1}^t\phi_{p,h,\tau}
\{V(x_{h+1,\tau})-\P_hV(x_{h,\tau},a_{h,\tau})\}
\right\|_{\Lambda_{p,h,t}^{-1}}
\le
H\sqrt{2\log\!\left(\frac{3H\Nsc_{h+1}^{\rm on}(\epsilon)}{p_0}\right)
+d_p\log(1+T/d_p)}.
\]
For a class member and its nearest net element, the difference of the two
centered targets is at most $2\epsilon$ pointwise. Hence
\[
\left\|\sum_{\tau=1}^t\phi_{p,h,\tau}\delta_\tau
\right\|_{\Lambda_{p,h,t}^{-1}}
\le\|\delta\|_2
\le2\epsilon\sqrt t\le2H.
\]
Substituting $R=M$, $B=H\sqrt{d_pT}$, $J_r=\gamma_r$, and
$J_p=\gamma_p$ in Lemma~\ref{lem:online_value_cover} gives
\[
\log\Nsc_{h+1}^{\rm on}(H/T)
\le C'(d_r+d_p)^2\zeta_{\rm on}(p_0).
\]
Combining the last three displays and using the class-membership conclusion of
Lemma~\ref{lem:online_value_cover} proves the claim for the data-dependent
current function $\bar V_{h+1,t}$.
\end{proof}

\begin{lemma}\label{lem:beta_err_online}
Under the assumptions of Theorem~\ref{thm:regret_online}, with probability at
least $1-p_0/3$, simultaneously for every $h\in[H]$, $t\in[T]$, and
$(x,a)\in\Ssc\times\Asc$,
\[
\left|
\langle\phi_p(x,a),\widehat\beta_{h,t}\rangle
-\E\{\bar V_{h+1,t}(x_{h+1})\mid x_h=x,a_h=a\}
\right|
\le
\gamma_p\|\phi_p(x,a)\|_{\Lambda_{p,h,t}^{-1}}.
\]
\end{lemma}
\begin{proof}
Define
\[
\beta_{h,t}
:=
\int_{\Ssc}\bar V_{h+1,t}(x')\,\mu_h(dx').
\]
The linear-transition representation and Assumption~\ref{assump:theta_online}
give
\[
\E\{\bar V_{h+1,t}(x_{h+1})\mid x_h=x,a_h=a\}
=\langle\phi_p(x,a),\beta_{h,t}\rangle,
\qquad
\|\beta_{h,t}\|_2\le H\sqrt{d_p}.
\]
The ridge normal equation yields
\[
\Lambda_{p,h,t}(\widehat\beta_{h,t}-\beta_{h,t})
=
-\beta_{h,t}
+\sum_{\tau=1}^t\phi_p(x_{h,\tau},a_{h,\tau})
\left[
\bar V_{h+1,t}(x_{h+1,\tau})
-\E\{\bar V_{h+1,t}(x_{h+1})\mid x_{h,\tau},a_{h,\tau}\}
\right].
\]
On the event in Lemma~\ref{lem:online_current_value_concentration},
Cauchy--Schwarz therefore gives
\begin{align*}
&\left|
\langle\phi_p(x,a),\widehat\beta_{h,t}-\beta_{h,t}\rangle
\right|\\
&\quad\le
\left[
H\sqrt{d_p}+CH(d_r+d_p)\sqrt{\zeta_{\rm on}(p_0)}
\right]
\|\phi_p(x,a)\|_{\Lambda_{p,h,t}^{-1}}\\
&\quad\le
\gamma_p\|\phi_p(x,a)\|_{\Lambda_{p,h,t}^{-1}},
\end{align*}
where the last inequality follows from the definition of $\gamma_p$ after
choosing the numerical constant $c_p$ sufficiently large.
\end{proof}

\begin{lemma}[Empirical transition-covariance concentration]\label{lem:concen}
For sufficiently large \(n\), with probability at least \(1-\xi\),
\[
\max_{h\in[H]}
\left\|n^{-1}\widetilde\Lambda_h-\Lambda_h\right\|_{\rm op}
\le
C\sqrt{\frac{\log(2d_pH/\xi)}{n}},
\]
where \(C>0\) is an absolute constant.
\end{lemma}
\begin{proof}
For each \(h\in[H]\), define
\[
Z_h^\tau
=
\phi_p(x_h^\tau,a_h^\tau)\phi_p(x_h^\tau,a_h^\tau)^\top
-\Lambda_h,
\qquad \tau\in[n].
\]
By the sampling convention, the \(Z_h^\tau\) are independent and mean zero
across \(\tau\), and the feature normalization gives
\(\|Z_h^\tau\|_{\rm op}\le2\). Matrix Bernstein's inequality therefore yields
\[
\left\|n^{-1}\sum_{\tau=1}^n Z_h^\tau\right\|_{\rm op}
\le
C\sqrt{\frac{\log(2d_pH/\xi)}{n}}
\]
with failure probability at most \(\xi/H\), after enlarging \(C\) if needed.
A union bound over \(h\in[H]\), together with
\(\widetilde\Lambda_h=\sum_{\tau=1}^n
\phi_p(x_h^\tau,a_h^\tau)\phi_p(x_h^\tau,a_h^\tau)^\top\),
proves the claim.
\end{proof}

\end{document}